\documentclass{llncs}%
\usepackage{graphicx}
\usepackage{amsmath}
\usepackage{amsfonts}
\usepackage{amssymb}
\usepackage{amsbsy}
\usepackage[draft]{changes} 
\usepackage{xcolor}
\usepackage{todonotes}
\usepackage{bm}
\usepackage{times}
\usepackage{todonotes}
\usepackage[toc,page]{appendix}
\usepackage{subfig}

\hyphenchar\font=\string"7F
\usepackage{mymacros}

\begin{document}
\title{Evaluating Conformance Measures in Process Mining using Conformance Propositions}
\subtitle{(Extended Version)}
%
%

\author{Anja F. Syring\inst{1} \and Niek Tax \inst{2} \and Wil M.P. van der Aalst \inst{1,2,3}}
\authorrunning{Anja F. Syring, Niek Tax,  Wil M.P. van der Aalst}
%
\institute{ Process and Data Science (Informatik 9), \\ RWTH Aachen University, D-52056 Aachen, Germany \and
			Architecture of Information Systems,\\ Eindhoven University of Technology, Eindhoven, The Netherlands \and
Fraunhofer Institute for Applied Information Technology FIT,\\ Sankt Augustin, Germany}

\maketitle              

\begin{abstract}
Process mining sheds new light on the relationship between process models and real-life processes.
Process discovery can be used to learn process models from event logs. Conformance checking is concerned with quantifying the \emph{quality} of a business process model in relation to event data that was logged during the execution of the business process. There exist different categories of conformance measures. \emph{Recall}, also called fitness, is concerned with quantifying how much of the behavior that was observed in the event log fits the process model. \emph{Precision} is concerned with quantifying how much behavior a process model allows for that was never observed in the event log. \emph{Generalization} is concerned with quantifying how well a process model generalizes to behavior that is possible in the business process but was never observed in the event log. Many recall, precision, and generalization measures have been developed throughout the years, but they are often defined in an ad-hoc manner without formally defining the desired properties up front.
To address these problems, we formulate \emph{21 conformance propositions} and we use these propositions to evaluate current and existing conformance measures.
The goal is to trigger a discussion by clearly formulating the challenges and requirements
(rather than proposing new measures). Additionally, this paper serves as an overview of the conformance checking measures that are available in the process mining area.

\end{abstract}
\section{Introduction}
\label{sec:intro}
Process mining~\cite{process-mining-book-2016} is a fast growing discipline that focuses on the analysis of event data that is logged during the execution of a business process. Events in such an event log contain information on what was done, by whom, for whom, where, when, etc. Such event data are often readily available from information systems that support the execution of the business process, such as ERP, CRM, or BPM systems. \emph{Process discovery}, the task of automatically generating a process model that accurately describes a business process based on such event data, plays a prominent role in process mining. Throughout the years, many process discovery algorithms have been developed, producing process models in various forms, such as Petri nets, process trees, and BPMN.

Event logs are often incomplete, i.e., they only contain a sample of all possible behavior in the business process. This not only makes process discovery challenging; it is also difficult to assess the quality of the process model in relation to the log.
Process discovery algorithms take an event log as input and aim to output a process model that satisfies certain properties, which are often referred to as the four quality dimensions \cite{process-mining-book-2016} of process mining:
(1) \emph{recall}: the discovered model should allow for the behavior seen in the event log (avoiding ``non-fitting'' behavior),
(2) \emph{precision}: the discovered model should not allow for behavior completely unrelated to what was seen in the event log (avoiding ``underfitting''),
(3) \emph{generalization}: the discovered model should generalize the example behavior seen in the event log (avoiding ``overfitting''), and
(4) \emph{simplicity}: the discovered model should not be unnecessarily complex.
The simplicity dimension refers to Occam's Razor: ``one should not increase, beyond what is necessary, the number of entities required to explain anything''.
In the context of process mining, this is often operationalized by quantifying the complexity of the model (number of nodes, number of arcs, understandability, etc.).
We do \emph{not} consider the simplicity dimension in this paper, since we focus on \emph{behavior} and abstract from the actual model representation.
Recall is often referred to as \emph{fitness} in process mining literature. Sometimes fitness refers to a combination of the four quality dimensions.
To avoid later confusion, we use the term recall which is commonly used in pattern recognition, information retrieval, and (binary) classification. Many conformance measures have been proposed throughout the years, e.g.,~\cite{process-mining-book-2016,wires-replay,arya-cost-fitness-edoc11,conf-check-book-2018,anti-align-bpm2016,alignments-Caise17,conf-check-SE2018,felix-computing-conf-2016,freshconf-bpm2010,anne_confcheck_is,jochen-de-weert-is2012,weerdt-F-measure_ssci_2011}.

So far it remains an open question whether existing measures for recall, precision, and generalization measure what they are aiming to measure. This motivates the need for a formal framework for conformance measures.
Users of existing conformance measures should be aware of seemingly obvious quality issues of existing approaches and researchers and developers that aim to create new measures should be clear on what conformance characteristics they aim to support. To address this open question, this paper evaluates state-of-the-art conformance measures based on 21 propositions introduced in~\cite{propositions-2018}.

The remainder is organized as follows.
Section~\ref{sec:relwork} discusses related work.
Section~\ref{sec:prelim} introduces basic concepts and notations.
The rest of the paper is split into two parts where the first one discusses the topics of recall and precision (Section~\ref{sec:recprec}) and the second part is dedicated to generalization (Section~\ref{sec:gen}).
In both parts, we introduce the corresponding conformance propositions and provide an overview of existing conformance measures. Furthermore, we discuss our findings of validating existing these measures on the propositions. Additionally, Section~\ref{sec:recprec} demonstrates the importance of the propositions on several baseline conformance measures, while Section~\ref{sec:gen} includes a discussion about the different points of view on generalization.
Section~\ref{sec:concl} concludes the paper.

\section{Related work} \label{sec:relwork}
In early years, when process mining started to gain in popularity and the community around it grew, many process discovery algorithms were developed. But at that time there was no standard method to evaluate the results of these algorithms and to compare them to the performance of other algorithms. Based on this, Rozinat et~al.\@~\cite{anne-BPI-BPM-ws-2007} called on the process mining community to develop a standard framework to evaluate process discovery algorithms.
This led to a variety of fitness/recall, precision, generalization and simplicity notions \cite{process-mining-book-2016}.
These notions can be quantified in different ways and there are often trade-offs between the different quality dimensions. As shown using generic algorithms assigning weights to the different quality dimensions \cite{genetic-mining-four-dim-coopis2012}, one quickly gets degenerate models when leaving out one or two dimensions. For example, it is very easy to create a simple model with perfect recall (i.e., all observed behavior fits perfectly) that has poor precision and provides no insights.

Throughout the years, several conformance measures have been developed for each quality dimension. However, it is unclear whether these measures actually measure what they are supposed to.
An initial step to address the need for a framework to evaluate conformance measures was made in~\cite{Niek-IPL2018-imprecision}. Five so-called \emph{axioms} for precision measures were defined that characterize the desired properties of such measures. Additionally, \cite{Niek-IPL2018-imprecision} showed that none of the existing precision measures satisfied all of the formulated axioms.
In comparison to \cite{Niek-IPL2018-imprecision} Janssenswillen et~al.\@~\cite{is-janssenswillen} did not rely on qualitative criteria, but quantitatively compared existing recall, precision and generalization measures under the aspect of feasibility, validity and sensitivity. The results showed that all recall and precision measures tend to behave in a similar way, while generalization measures seemed to differ greatly from each other.
In~\cite{propositions-2018} van der Aalst made a follow-up step to~\cite{Niek-IPL2018-imprecision} by formalizing recall and generalization in addition to precision and by extending the precision requirements, resulting in a list of 21 conformance propositions. Furthermore, \cite{propositions-2018} showed the importance of probabilistic conformance measures that also take into account trace probabilities in process models.
Beyond that, \cite{Niek-IPL2018-imprecision} and~\cite{propositions-2018} motivated the process mining community to develop new precision measures, taking the axioms and propositions as a design criterion, resulting in the measures among others the measures that are proposed in~\cite{polyvyanyy-conf} and in~\cite{augusto-precision}.
Using the 21 propositions of ~\cite{propositions-2018} we evaluate state-of-the-art recall (e.g.~\cite{aal_min_TKDE,polyvyanyy-conf,wires-replay,stijn-JMLR2009,sander-scalable-procmin-SOSYM,anne_confcheck_is,tonbeta166}), precision (e.g.~\cite{wires-replay,stijn-JMLR2009,GrecoTKDE2006,sander-scalable-procmin-SOSYM,anti-align-bpm2016,polyvyanyy-conf,anne_confcheck_is,weighted-n-events}) and generalization (e.g.~\cite{wires-replay,anti-align-bpm2016,stijn-JMLR2009}) measures.

This paper uses the mainstream view that there are at least four quality dimensions: fitness/recall, precision, generalization, and simplicity \cite{process-mining-book-2016}. We deliberately do not consider simplicity, since we focus on behavior only (i.e., not the model representation). Moreover, we treat generalization separately.
In a controlled experiment one can assume the existence of a so-called ``system model''. This model can be simulated to create a synthetic event log used for discovery. In this setting, conformance checking can be reduced to measuring the similarity between the discovered model and the system model \cite{thesis_buijs,BPM-jans}. In terms of the well-known confusion matrix, one can then reason about true positives, false positives, true negatives, and false negatives. However, without a system model and just an event log, it is not possible to find false positives (traces possible in the model but not in reality). Hence, precision cannot be determined in the traditional way.
Janssenswillen and Depaire \cite{Janssenswillen2018} conclude in their evaluation of state-of-the-art conformance measures that none of the existing approaches reliably measures this similarity. However, in this paper, we follow the traditional view on the quality dimensions and exclude the concept of the system from our work.

Whereas there are many fitness/recall and precision measures there are fewer generalization measures.
Generalization deals with future cases that were not yet observed. There is no consensus on how to define generalization and in \cite{is-janssenswillen} it was shown that there is no agreement between existing generalization metrics. Therefore, we cover generalization in a separate section (Section~\ref{sec:gen}).
However, as discussed in \cite{process-mining-book-2016} and demonstrated through experimentation  \cite{genetic-mining-four-dim-coopis2012}, one cannot leave out the generalization dimension. The model that simply enumerates all the traces in the log has perfect fitness/recall and precision. However, event logs cannot be assumed to be complete, thus proving that a generalization dimension is needed.

\section{Preliminaries}
\label{sec:prelim}
A \emph{multiset} over a set $X$ is a function $B:X\rightarrow\mathbb{N}$ which we write as $[a_1^{w_1},a_2^{w_2},\dots,a_n^{w_n}]$ where for all $i\in[1,n]$ we have $a_i\in X$ and $w_i\in\mathbb{N}^*$. $\mathbb{B}(X)$ denotes the set of all multisets over set $X$. For example, $[a^3,b,c^2]$ is a multiset over set $X=\{a,b,c\}$ that contains three $a$ elements, one $b$ element and two $c$ elements. $\card{B}$ is the number of elements in multiset $B$ and $B(x)$ denotes the number of $x$ elements in $B$.
$B_1 \uplus B_2$ is the sum of two multisets: $(B_1 \uplus B_2)(x) = B_1(x) + B_2(x).$
$B_1 \setminus B_2$ is the difference containing all elements from $B_1$ that do not occur in $B_2$. Thus, $(B_1 \setminus B_2)(x) = \max\{ B_1(x)-B_2(x),0\}$.
$B_1 \cap B_2$ is the intersection of two multisets. Hence, $(B_1 \cap B_2)(x) = \min\{ B_1(x),B_2(x)\}$.
$[x \in B \mid b(x)]$ is the multiset of all elements in $B$ that satisfy some condition $b$.
$B_1 \subseteq B_2$ denotes that $B_1$ is contained in $B_2$, e.g., $[a^2,b] \subseteq [a^2,b^2,c]$, but $[a^2,b^3] \not\subseteq [a^2,b^2,c^2]$ and $[a^2,b^2,c] \not\subseteq [a^3,b^3]$.

Process mining techniques focus on the relationship between observed behavior and modeled behavior.
Therefore, we first formalize event logs (i.e., observed behavior) and process models (i.e., modeled behavior).
To do this, we consider a very simple setting where we only focus on the control-flow, i.e., sequences of activities.

\subsection{Event Logs}
\label{sec:evl}

The starting point for process mining is an event log.
Each \emph{event} in such a log refers to an \emph{activity}
possibly executed by a \emph{resource} at a particular \emph{time} and for a particular \emph{case}.
An event may have many more attributes, e.g., transactional information, costs, customer, location, and unit.
Here, we focus on control-flow. Therefore, we only consider activity labels and the ordering of events within cases.

\begin{definition}[Traces] 
$\uacts$ is the universe of \emph{activities}.
A \emph{trace} $t \in \uacts^*$ is a sequence of activities.
$\utraces = \uacts^*$ is the universe of traces.
\end{definition}

Trace $t = \la a,b,c,d,a \ra$ refers to 5 events belonging to the same case (i.e., $\card{t}=5$).
An event log is a collection of cases each represented by a trace.

\begin{definition}[Event Log] 
$\ulogs = \ubagtraces$ is the universe of event logs.
An \emph{event log} $l \in \ulogs$ is a finite multiset of observed traces.
$\traces{l} = \{t \in l\}\subseteq \utraces$ is the set of traces appearing in $l \in \ulogs$.
$\compltr{l} = \utraces\setminus\traces{l}$ is the complement of the set of non-observed traces.
\end{definition}

Event log $l = [ \la a,b,c \ra^5,\la b,a,d \ra^3,\la a,b,d \ra^2 ]$
refers to 10 cases (i.e., $\card{l}=10$). Five cases are represented by the trace $\la a,b,c \ra$, three cases are represented by the trace $\la b,a,d \ra$, and two cases are represented by the trace $\la a,b,d \ra$. Hence, $l(\la a,b,d \ra) = 2$.

\subsection{Process Models}
\label{sec:pm}

The behavior of a process model $m$ is simply the set of traces allowed by $m$. In our definition, we will abstract from the actual representation (e.g. Petri nets or BPMN).

\begin{definition}[Process Model] 
$\umodels$ is the set of process models. A process model $m \in \umodels$ allows for a set of traces $\traces{m} \subseteq \utraces$. $\compltr{m} = \mathcal{T}\setminus\traces{m}$ is the complement of the set of traces allowed by model $m \in \umodels$.
\end{definition}

A process model $m\in\mathcal{M}$ may abstract from the real process and leave out unlikely behavior.
Furthermore, this abstraction can result in $\traces{m}$ allowing for traces that cannot happen (e.g., particular interleavings or loops).

\begin{figure}[t]
	{
		\centering
		\includegraphics[width=0.99\textwidth]{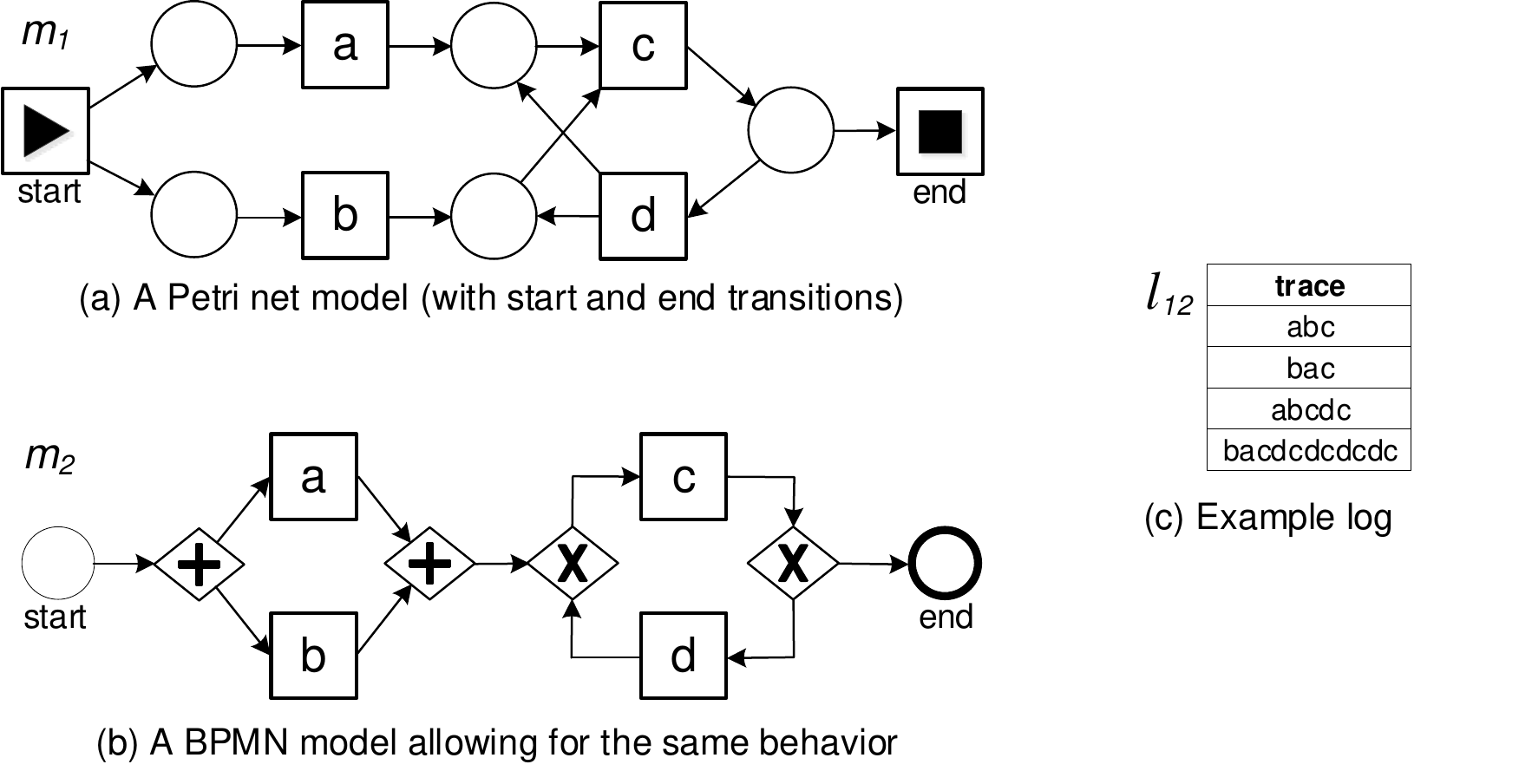}
		\caption{Two process models $m_1$ and $m_2$ allowing for the same set of traces ($\traces{m_1} = \traces{m_2}$) with an example log $l_{12}$ (c).
		}\label{f-procmod-loop}
	}
\end{figure}

We distinguish between \emph{representation} and \emph{behavior} of a model.
Process model $m \in \umodels$ can be represented using a plethora of modeling languages, e.g., Petri nets, BPMN models, UML activity diagrams, automata, and process trees.
Here, we abstract from the actual representation and focus on behavioral characteristics
$\traces{m} \subseteq \utraces$.

Figure~\ref{f-procmod-loop} (a) and (b) show two process models that have the same behavior: \allowbreak $\traces{m_1} = \traces{m_2} =\{\la a,b,c\ra,\la a,c,b\ra,\la a,b,c,d,c\ra,\la b,a,c,d,c\ra,\dots\}$.
Figure~\ref{f-procmod-loop}(c) shows a possible event log generated by one of these models $l_{12} = [ \la a,b,c \ra^3, \allowbreak \la b,a,c \ra^5, \break \la a,b,c,d,c \ra^2, \allowbreak \la b,a,c,d,c,d,c,d,c,d,c \ra^2 ]$ .

The behavior $\tau(m)$ of a process model $m\in\mathcal{M}$ can be of infinite size. We use Figure~\ref{f-procmod-loop} to illustrate this. There is a ``race'' between $a$ and $b$. After $a$ and $b$, activity $c$ will occur. Then there is a probability that the process ends or $d$ can occur. Let $t_{a,k} = \la a,b \ra \cdot (\la c,d \ra)^k \cdot \la c \ra$ be the trace that starts with $a$ and where $d$ is executed $k$ times. $t_{b,k} = \la b,a \ra \cdot (\la c,d \ra)^k \cdot \la c \ra$ is the trace that starts with $b$ and where $d$ is executed $k$ times.
$\traces{m_1} = \traces{m_2} = \bigcup_{k \geq 0} \{t_{a,k},t_{b,k}\}$.
Some examples are given in Figure~\ref{f-procmod-loop}(c).

Since any log contains only a finite number of traces, one can never observe all traces possible
in $m_1$ or $m_2$.

\subsection{Process Discovery}
\label{sec:pd}

\begin{figure}[t]
	{
		\centering
		\includegraphics[width=0.7\textwidth]{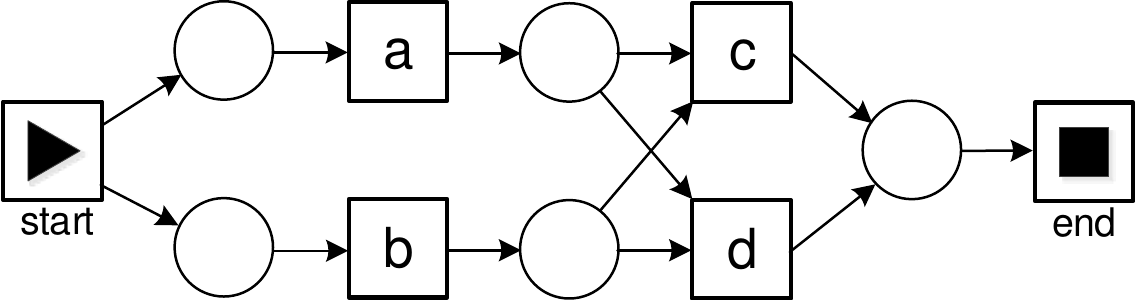}
		\caption{A process model $m_3$ discovered based on log $l_3 = [ \la a,b,c \ra^5,\la b,a,d \ra^3,\la a,b,d \ra^2 ]$.
		}\label{f-mod2}
	}
\end{figure}

A discovery algorithm takes an event log as input and returns a process model.
For example, the model $m_3$ in Figure~\ref{f-mod2} could have been discovered based on event log $l_3 = [ \la a,b,c \ra^5, \allowbreak \la b,a,d \ra^3, \allowbreak \la a,b,d \ra^2 ]$.
Ideally, the process model should capture the (dominant) behavior observed but it should also generalize without becoming too imprecise. For example, the model allows for trace $t = \la b,a,c \ra$ although this was never observed.

\begin{definition}[Discovery Algorithm]
A discovery algorithm can be described as a function $\mi{disc} \in \ulogs \rightarrow \umodels$ mapping event logs onto process models.
\end{definition}

We abstract from concrete discovery algorithms. Over 100 discovery algorithms have been proposed in literature \cite{process-mining-book-2016}.
Merely as a reference to explain basic notions, we define three simple, but extreme, algorithms: $\mi{disc}_{\mi{ofit}}$, $\mi{disc}_{\mi{ufit}}$, and $\mi{disc}_{\mi{nfit}}$.
Let $l \in \ulogs$ be a log.
$\mi{disc}_{\mi{ofit}}(l) = m_o$ such that $\traces{m_o} = \traces{l}$ produces an overfitting model that allows only for the behavior seen in the log.
$\mi{disc}_{\mi{ufit}}(l) = m_u$ such that $\traces{m_u} = \utraces$ produces an underfitting model that allows for any behavior.
$\mi{disc}_{\mi{nfit}}(l) = m_n$ such that $\traces{m_n} = \compltr{l}$ produces a non-fitting model that allows for all behavior \emph{not} seen in the log.




\section{Recall and Precision}\label{sec:recprec}

Many recall measures have been proposed in literature \cite{process-mining-book-2016,wires-replay,arya-cost-fitness-edoc11,conf-check-book-2018,anti-align-bpm2016,alignments-Caise17,conf-check-SE2018,felix-computing-conf-2016,freshconf-bpm2010,anne_confcheck_is,jochen-de-weert-is2012,weerdt-F-measure_ssci_2011}.
In recent years, also several precision measures have been proposed \cite{align-ETC2013,Niek-IPL2018-imprecision}.
Only few generalization measures have been proposed \cite{wires-replay}.
The goal of this paper is to evaluate these quality measures. To achieve this, in the following the propositions introduced in \cite{propositions-2018} are applied to existing conformance measures.

The notion of recall and precision are well established in the process mining community. Definitions are in place and there is an agreement on what these two measures are supposed to measure. However, this is not the case for generalization. There exist different points of view on what generalization is supposed to measure. Depending on these, existing generalization measures might greatly differ from each other.

To account for the different levels of maturity in recall, precision and generalization and to address the controversy in the generalization area, the following section will solely handle recall and precision while Section~\ref{sec:gen} focuses on generalization. Both sections establish baseline measures, introduce the corresponding propositions of \cite{propositions-2018}, present existing conformance measures and evaluate them using the propositions.

\subsection{Baseline Recall and Precision measures}\label{sec:base_recprec}
We assume the existence of two functions: $\mi{rec}()$ and $\mi{prec}()$ respectively denoting recall and precision. Both take a log and model as input and return a value between 0 and 1. The higher the value, the better.

\begin{definition}[Recall]\label{defr}
A \emph{recall measure} $\mi{rec} \in \ulogs \times \umodels \rightarrow [0,1]$ aims to quantify the fraction of observed behavior that is allowed by the model.
\end{definition}

\begin{definition}[Precision]
A \emph{precision measure} $\mi{prec} \in \ulogs \times \umodels \rightarrow [0,1]$ aims to quantify the fraction of behavior allowed by the model that was actually observed.
\end{definition}

If we ignore frequencies of traces, we can simply count fractions of traces yielding the following two simple measures.

\begin{definition}[Trace-Based L2M Precision and Recall]\label{l2mtrprecision} 
Let $l \in \ulogs$ and $m \in \umodels$ be an event log and a process model. Trace-based L2M precision and recall are defined as follows:
\begin{eqnarray}
\mi{rec}_{\mi{TB}}(l,m) = \frac{\card{\traces{l}\cap\traces{m}}}{\card{\traces{l}}} & ~~ &
\mi{prec}_{\mi{TB}}(l,m) = \frac{\card{\traces{l}\cap\traces{m}}}{\card{\traces{m}}}
\end{eqnarray}
\end{definition}

Since $\card{\traces{l}}$ is bounded by the size of the log, $\mi{rec}_{\mi{TB}}(l,m)$ is well-defined.
However, $\mi{prec}_{\mi{TB}}(l,m)$ is undefined when $\traces{m}$ is unbounded (e.g., in case of loops).

One can argue, that the frequency of traces should be taken into account when evaluating conformance which yields the following measure.
Note that it is not possible to define frequency-based precision based on a process model that does not define the probability of its traces.
Since probabilities are specifically excluded from the scope of this paper, the following approach only defines frequency-based recall.


\begin{definition}[Frequency-Based L2M Recall]\label{l2mestprecision} 
Let $l \in \ulogs$ and $m \in \umodels$ be an event log and a process model.
Frequency-based L2M recall is defined as follows:
\begin{eqnarray}
\mi{rec}_{\mi{FB}}(l,m) & = &
\frac{\card{[t \in l \mid t \in \traces{m}]}}{\card{l}}
\end{eqnarray}
\end{definition}


\subsection{A Collection of Conformance Propositions}
\label{sec:confprop}
In \cite{propositions-2018}, 21 \emph{conformance propositions} covering the different conformance dimensions (except simplicity) were given.
In this section, we focus on the general, recall and precision propositions introduced in \cite{propositions-2018}.
We discuss the generalization propositions separately, because they reason about unseen cases not yet recorded in the event log.
Most of the conformance propositions have broad support from the community, i.e., there is broad consensus that these propositions should hold. These are marked with a ``$+$''. More controversial propositions are marked with a ``$0$'' (rather than a ``$+$'').

\subsubsection{General Propositions}
\label{sec:algprop}
The first two propositions are commonly accepted; the computation of a quality measure should be deterministic (\textbf{DetPro${}^{+}$})
and only depend on behavioral aspects (\textbf{BehPro${}^{+}$}). The latter is a design choice. We deliberately exclude simplicity notions.

  \begin{proposition}[\textbf{DetPro${}^{+}$}] \label{DetPro}
    $\mi{rec}()$, $\mi{prec}()$, $\mi{gen}()$ are deterministic functions, i.e., the measures $\mi{rec}(l,m)$, $\mi{prec}(l,m)$, $\mi{gen}(l,m)$ are fully determined by $l \in \ulogs$ and $m \in \umodels$.
  \end{proposition}

   \begin{proposition}[\textbf{BehPro${}^{+}$}] \label{BehPro}
   For any $l \in \ulogs$ and $m_1,m_2 \in \umodels$ such that $\traces{m_1} = \traces{m_2}$: $\mi{rec}(l,m_1)=\mi{rec}(l,m_2)$, $\mi{prec}(l,m_1)=\mi{prec}(l,m_2)$, and $\mi{gen}(l,m_1) = \linebreak \mi{gen}(l,m_2)$,
   i.e., the measures are fully determined by the behavior observed and the behavior described by the model (representation does not matter).
  \end{proposition}

\subsubsection{Recall Propositions}
\label{sec:recprop}

In this subsection, we consider a few  \emph{recall propositions}.
$\mi{rec} \in \ulogs \times \umodels \rightarrow [0,1]$ aims to quantify the fraction of observed behavior that is allowed by the model. Proposition \textbf{RecPro1${}^{+}$} states that extending the model to allow for more behavior can never result in a lower recall. From the definition follows, that this proposition implies \textbf{BehPro${}^{+}$}. Recall measures violating \textbf{BehPro${}^{+}$} also violate \textbf{RecPro1${}^{+}$} which is demonstrated as follows:

For two models $m_1,m_2$ with $\traces{m_1}=\traces{m_2}$ it follows from \textbf{RecPro1${}^{+}$} that $\mi{rec}(l,m_1) \leq \mi{rec}(l,m_2)$ because $\traces{m_1} \subseteq \traces{m_2}$. From \textbf{RecPro1${}^{+}$} follows that \linebreak $\mi{rec}(l,m_2) \leq \mi{rec}(l,m_1)$ because $\traces{m_2} \subseteq \traces{m_1}$. Combined, $\mi{rec}(l,m_2) \leq \mi{rec}(l,m_1)$ and $\mi{rec}(l,m_1) \leq \mi{rec}(l,m_2)$ gives $\mi{rec}(l,m_2) = \mi{rec}(l,m_1)$, thus, recall measures that fulfill \textbf{RecPro1${}^{+}$} are fully determined by the behavior observed and the behavior described by the model, i.e., representation does not matter.

 \begin{proposition}[\textbf{RecPro1${}^{+}$}] \label{RecPro1}
    For any $l \in \ulogs$ and $m_1,m_2 \in \umodels$ such that $\traces{m_1} \subseteq \traces{m_2}$: $\mi{rec}(l,m_1) \leq \mi{rec}(l,m_2)$.
  \end{proposition}

Similarly to \textbf{RecPro1${}^{+}$}, it cannot be the case that adding fitting behavior to the event logs, lowers recall (\textbf{RecPro2${}^{+}$}).

  \begin{proposition}[\textbf{RecPro2${}^{+}$}] \label{RecPro2}
    For any $l_1,l_2,l_3 \in \ulogs$ and $m \in \umodels$ such that
    $l_2 = l_1 \uplus l_3$ and $\traces{l_3} \subseteq \traces{m}$: $\mi{rec}(l_1,m) \leq \mi{rec}(l_2,m)$.
  \end{proposition}

Similarly to \textbf{RecPro2${}^{+}$}, one can argue that adding non-fitting behavior to event logs should not be able to increase recall (\textbf{RecPro3${}^{0}$}). However, one could also argue that recall should not be measured on a trace-level, but should instead distinguish between non-fitting traces by measuring the \emph{degree} in which a non-fitting trace is still fitting. Therefore, unlike the previous propositions, this requirement is debatable as is indicated by the ``$0$'' tag.
    \begin{proposition}[\textbf{RecPro3${}^{0}$}] \label{RecPro3}
    For any $l_1,l_2,l_3 \in \ulogs$ and $m \in \umodels$ such that
    $l_2 = l_1 \uplus l_3$ and $\traces{l_3} \subseteq \compltr{m}$: $\mi{rec}(l_1,m) \geq \mi{rec}(l_2,m)$.
  \end{proposition}

For any $k \in\mathbb{N}$: $l^k(t) = k \cdot l(t)$, e.g., if $l = [ \la a,b \ra^{3},\la c \ra^{2} ]$, then $l^{4} = [ \la a,b \ra^{12},\la c \ra^{8} ]$.
We use this notation to enlarge event logs without changing the original distribution. One could argue that this should not influence recall ($\textbf{RecPro4${}^{0}$}$),
e.g., $\mi{rec}([ \la a,b \ra^{3},\allowbreak \la c \ra^{2} ],m) = \mi{rec}([ \la a,b \ra^{12},\la c \ra^{8} ],m)$. On the other hand, larger logs can provide more confidence that the log is indeed a representative sample of the possible behavior. Therefore, it is debatable whether the size of the event log should have influence on recall as indicated by the ``$0$'' tag.

  \begin{proposition}[\textbf{RecPro4${}^{0}$}] \label{RecPro4}
    For any $l \in \ulogs$, $m \in \umodels$, and $k\geq 1$: $\mi{rec}(l^k,m) = \mi{rec}(l,m)$.
  \end{proposition}

Finally, we provide a proposition stating that recall should be 1 if all traces in the log fit the model (\textbf{RecPro5${}^{+}$}). As a result, the empty log has recall 1 for any model. Based on this proposition, $\mi{rec}(l,\mi{disc}_{\mi{ofit}}(l)) =\allowbreak  \mi{rec}(l,\mi{disc}_{\mi{ufit}}(l)) =\allowbreak 1$ for any log $l$.

      \begin{proposition}[\textbf{RecPro5${}^{+}$}] \label{RecPro5}
    For any $l \in \ulogs$ and $m \in \umodels$ such that
    $\traces{l} \subseteq \traces{m}$: $\mi{rec}(l,m)=1$.
  \end{proposition}

\subsubsection{Precision Propositions}
\label{sec:precprop}
Precision ($\mi{prec} \in \ulogs \times \umodels \rightarrow [0,1]$) aims to quantify the fraction of behavior allowed by the model that was actually observed. Initial work in the area of checking requirements of conformance checking measures started with~\cite{Niek-IPL2018-imprecision}, where five axioms for precision measures were introduced. The precision propositions that we state below partly overlap with these axioms, but some have been added and some have been strengthened. Axiom 1 of~\cite{Niek-IPL2018-imprecision} specifies \textbf{DetPro${}^{+}$} for the case of precision, while we have generalized it to the recall and generalization dimension. Furthermore, \textbf{BehPro${}^{+}$} generalizes axiom 4 of~\cite{Niek-IPL2018-imprecision} from its initial focus on precision to also cover recall and generalization.
\textbf{PrecPro1${}^{+}$} states that removing behavior from a model that does not happen in the event log cannot lead to a lower precision. From the definition follows, that this proposition implies \textbf{BehPro${}^{+}$}. Precision measures violating \textbf{BehPro${}^{+}$} also violate \textbf{PrecPro1${}^{+}$}.
Adding fitting traces to the event log can also not lower precision (\textbf{PrecPro2${}^{+}$}).
However, adding non-fitting traces to the event log should not change precision (\textbf{PrecPro3${}^{0}$}).

  \begin{proposition}[\textbf{PrecPro1${}^{+}$}] \label{PrecPro1}
    For any $l \in \ulogs$ and $m_1,m_2 \in \umodels$ such that
    $\traces{m_1} \subseteq \traces{m_2}$ and $\traces{l} \cap (\traces{m_2} \setminus \traces{m_1}) = \emptyset$: $\mi{prec}(l,m_1) \geq \mi{prec}(l,m_2)$.
  \end{proposition}

This proposition captures the same idea as axiom 2 in~\cite{Niek-IPL2018-imprecision}, but it is more general. Axiom 2 only put this requirement on precision when $\traces{l}\subseteq\traces{m_1}$, while \textbf{PrecPro1${}^{+}$} also concerns the situation where this does not hold.

  \begin{proposition}[\textbf{PrecPro2${}^{+}$}] \label{PrecPro2}
    For any $l_1,l_2,l_3 \in \ulogs$ and $m \in \umodels$ such that
    $l_2 = l_1 \uplus l_3$ and $\traces{l_3} \subseteq \traces{m}$: $\mi{prec}(l_1,m) \leq \mi{prec}(l_2,m)$.
  \end{proposition}

This proposition is identical to axiom 5 in~\cite{Niek-IPL2018-imprecision}.

    \begin{proposition}[\textbf{PrecPro3${}^{0}$}] \label{PrecPro3}
    For any $l_1,l_2,l_3 \in \ulogs$ and $m \in \umodels$ such that
    $l_2 = l_1 \uplus l_3$ and $\traces{l_3} \subseteq \compltr{m}$: $\mi{prec}(l_1,m) = \mi{prec}(l_2,m)$.
  \end{proposition}

One could also argue that duplicating the event log should not influence precision because the distribution remains the same (\textbf{PrecPro4${}^{0}$}),
e.g., $\mi{prec}([ \la a,b \ra^{20},\la c \ra^{20} ],m) = \mi{prec}([ \la a,b \ra^{40},\la c \ra^{40} ],m)$. Similar to (\textbf{RecPro3${}^{0}$}) and (\textbf{RecPro4${}^{0}$}), the equivalents on the precision side are tagged with ``0".

  \begin{proposition}[\textbf{PrecPro4${}^{0}$}] \label{PrecPro4}
    For any $l \in \ulogs$, $m \in \umodels$, and $k\geq 1$: $\mi{prec}(l^k,m) = \mi{prec}(l,m)$.
  \end{proposition}

If the model allows for the behavior observed and nothing more,  precision should be maximal (\textbf{PrecPro5${}^{+}$}).
One could also argue that if all modeled behavior was observed, precision should also be 1 (\textbf{PrecPro6${}^{0}$}).
The latter proposition is debatable because it implies that the non-fitting behavior cannot influence perfect precision, as indicated by the ``0" tag.
Consider for example extreme cases where the model covers just a small fraction of all observed behavior (or even more extreme situations like $\traces{m} = \emptyset$). According to \textbf{PrecPro5${}^{+}$}and \textbf{PrecPro6${}^{0}$}, $\mi{rec}(l,\mi{disc}_{\mi{ofit}}(l)) =\allowbreak  1$ for any log $l$.

  \begin{proposition}[\textbf{PrecPro5${}^{+}$}] \label{PrecPro5}
    For any $l \in \ulogs$ and $m \in \umodels$ such that
    $\traces{m} = \traces{l}$: $\mi{prec}(l,m)=1$.
  \end{proposition}

  \begin{proposition}[\textbf{PrecPro6${}^{0}$}] \label{PrecPro6}
    For any $l \in \ulogs$ and $m \in \umodels$ such that
    $\traces{m} \subseteq \traces{l}$: $\mi{prec}(l,m)=1$.
  \end{proposition}

\subsection{Evaluation of Baseline Conformance Measures}

To illustrate the presented propositions and justify their formulation, we evaluate the conformance measures defined as baselines in Section~\ref{sec:base_recprec}. Note that these 3 baseline measures were introduced to provide simple examples that can be used to discuss the propositions. We conduct this evaluation under the assumption that $l \neq [~]$, $\traces{m} \neq \emptyset$ and $\la\ra\not\in\traces{m}$. Table~\ref{tabeval} in Appendix~\ref{appendixtab} summarizes the evaluation.

\subsubsection{General Propositions.}
Based on the definition of $\mi{rec}_{\mi{TB}}$ and $\mi{rec}_{\mi{FB}}$ it is clear that all measures can be fully determined by the log and the model. Consequently, \textbf{DetPro${}^{+}$} hold for these two baseline conformance measures. However, $\mi{prec}_{\mi{TB}}$ is undefined when $\traces{m}$ is unbound and, therefore, non-deterministic.

The behavior of the model is defined as sets of traces $\traces{m}$, which abstracts from the representation of the process model itself. Therefore, all recall and precision baseline conformance measures fulfill \textbf{BehPro${}^{+}$}.

\subsubsection{Recall Propositions.}

Considering measure $\mi{rec}_{\mi{TB}}$, it is obvious that \textbf{RecPro1${}^{+}$} holds if
$\traces{m_1} \subseteq \traces{m_2}$, because the intersection between $\traces{m_2}$ and $\traces{l}$ will always be equal or bigger to the intersection of $\traces{m_1}$ and $\traces{l}$.
The \textbf{RecPro2${}^{+}$} proposition holds for $\mi{rec}_{\mi{TB}}$, if $l_2 = l_1 \uplus l_3$ and $\traces{l_3} \subseteq \traces{m}$, because the additional fitting behavior is added to the nominator as well as the denominator of the formula: $\card{\traces{l_1}\cap\traces{m}} + \card{\traces{l_3}}) /(\card{\traces{l_1}} + \card{\traces{l_3}}$. This can never decrease recall.
Furthermore, \textbf{RecPro3${}^{0}$} propositions holds for $\mi{rec}_{\mi{TB}}$ since adding unfitting behavior cannot increase the intersection between traces of the model and the log if $l_2 = l_1 \uplus l_3$ and $\traces{l_3} \subseteq \compltr{m}$.  
Consequently, only the denominator of the formula grows, which decreases recall.
Similarly, we can show that these two proposition hold for $\mi{rec}_{\mi{FB}}$.

Duplication of the event log cannot affect $\mi{rec}_{\mi{TB}}$, since it is defined based on the set of traces and not the multiset. The proposition also holds for $\mi{rec}_{\mi{FB}}$ since nominator and denominator of the formula will grow in proportion. Hence, \textbf{RecPro4${}^{0}$} holds for both baseline measures.
Considering $\mi{rec}_{\mi{TB}}$, \textbf{RecPro5${}^{+}$} holds, since  $\traces{l}\cap\traces{m}= \traces{l}$ if $\traces{l} \subseteq \traces{m}$ and consequently $\card{\traces{l}\cap\traces{m}}/{\card{\traces{l}}} = \card{\traces{l}}/{\card{\traces{l}}} = 1$.
The same conclusions can be drawn for $\mi{rec}_{\mi{FB}}$.

\subsubsection{Precision Propositions.}

Consider proposition \textbf{PrecPro1${}^{+}$} together with $\mi{prec}_{\mi{TB}}$. The proposition holds, since removing behavior from the model that does not happen in the event log will not affect the intersection between the traces of the model and the log: $\traces{l}\cap\traces{m_2} = \traces{l}\cap\traces{m_1}$ if
$\traces{m_1} \subseteq \traces{m_2}$ and $\traces{l} \cap (\traces{m_2} \setminus \traces{m_1}) = \emptyset$. At the same time the denominator of the formula decreases, which can never decrease precision itself.
\textbf{PrecPro2${}^{+}$} also holds for $\mi{prec}_{\mi{TB}}$, since the fitting behavior increases the intersection between traces of the model and the log, while the denominator of the formula stays the same.
Furthermore, \textbf{PrecPro3${}^{0}$} holds for $\mi{prec}_{\mi{TB}}$, since unfitting behavior cannot affect the intersection between traces of the model and the log.

Duplication of the event log cannot affect $\mi{prec}_{\mi{TB}}$, since it is defined based on the set of traces and not the multiset, i.e. \textbf{PrecPro4${}^{0}$} holds.

Considering $\mi{prec}_{\mi{TB}}$, \textbf{PrecPro5${}^{+}$} holds, since  $\traces{l}\cap\traces{m}= \traces{m}$ if $\traces{m} = \traces{l}$ and consequently $\card{\traces{l}\cap\traces{m}}  /{\card{\traces{m}}}  = \card{\traces{m}}/{\card{\traces{m}}} = 1$. Similarly, \textbf{PrecPro6${}^{0}$} holds for $\mi{prec}_{\mi{TB}}$. 

\subsection{Existing Recall Measures}\label{subsec:stateofart}
The previous evaluation of the simple baseline measures shows that the recall measures fulfill all propositions and the baseline precision measure only violates one proposition. However, the work presented in \cite{Niek-IPL2018-imprecision} demonstrated for precision, that most of the existing approaches violate seemingly obvious requirements. This is surprising compared to the results of our simple baseline measure. Inspired by \cite{Niek-IPL2018-imprecision}, this paper takes a broad look at existing conformance measures with respect to the previously presented propositions. In the following section, existing recall and precision measures are introduced, before they will be evaluated in Section~\ref{eval}.

\subsubsection{Causal footprint recall ($rec_A$).}
Van der Aalst et~al.\@ \cite{aal_min_TKDE} introduce the concept of the footprint matrix, which captures the relations between the different activities in the log. The technique relies on the principle that if activity $a$ is followed by $b$ but $b$ is never followed by $a$, then there is a causal dependency between $a$ and $b$. The log can be described using four different relations types.
In \cite{process-mining-book-2016} it is stated that a footprint matrix can also be derived for a process model by generating a complete event log from it. Recall can be measured by counting the mismatches between both matrices. Note that this approach assumes an event log which is complete with respect to the directly follows relations.

\subsubsection{Token replay recall ($rec_B$).}
Token replay measures recall by replaying the log on the model and counting mismatches in the form of missing and remaining tokens. This approach was proposed by Rozinat and van der Aalst \cite{anne_confcheck_is}. During replay, four types of tokens are distinguished: $p$ the number of \textit{produced} tokens, $c$ the number of \textit{consumed} tokens, $m$ the number of \textit{missing} tokens that had to be added because a transition was not enabled during replay and $r$ the number of \textit{remaining} tokens that are left in the model after replay. In the beginning, a token is produced in the initial place. Similarly, the approach ends by consuming a token from the final place. The more missing and remaining tokens are counted during replay the lower recall: $rec_B = \frac{1}{2}(1 - \frac{m}{c}) + \frac{1}{2}(1 - \frac{r}{p}) $ Note that the approach assumes a relaxed sound workflow net, but it allows for duplicate and silent transitions.


\subsubsection{Alignment recall ($rec_C$).}
Another approach to determine recall was proposed by van der Aalst et~al.\@ \cite{wires-replay}. It calculates recall based on alignments, which detect process deviations by mapping the steps taken in the event log to the ones of the process model. This map can contain three types of steps (so-called moves): \textit{synchronous} moves when event log and model agree, \textit{log} moves if the event was recorded in the event log but should not have happened according to the model and \textit{model} moves if the event should have happened according to the model but did not in the event log. The approach uses a function that assigns costs to log moves and model moves. This function is used to compute the optimal alignment for each trace in the log (i.e. the alignment with the least cost associated).

To compute recall, the total alignment cost of the log is normalized with respect to the cost of the worst-case scenario where there are only moves in the log and in the model but never together. Note, that the approach assumes an accepting Petri net with an initial and final state. However, it allows for duplicate and silent transitions in the process model.


\subsubsection{Behavioral recall ($rec_D$).}
Goedertier et~al.\@ \cite{stijn-JMLR2009} define recall according to its definition in the data mining field using true positive (TP) and false negative (FN) counters.
$TP(l,m)$ denotes the number of true positives, i.e., the number of events in the log that can be parsed correctly in model $m$ by firing a corresponding enabled transition. $FN(l,m)$ denotes the number of false negatives, i.e., the number of events in the log for which the corresponding transition that was needed to mimic the event was not enabled and needed to be force-fired. The recall measure is defined as follows: $rec_D(l,m)=\frac{TP(l,m)}{TP(l,m)+FN(l,m)}$.

\subsubsection{Projected recall ($rec_E$).}
Leemans et~al.\@ \cite{sander-scalable-procmin-SOSYM} developed a conformance checking approach that is also able to handle big event logs. This is achieved by projecting the event log as well as the model on all possible subsets of activities of size $k$. The behavior of a projected log and projected model is translated into the minimal deterministic finite automata (DFA)\footnote{Every regular language has a unique minimal DFA according to the Myhill--Nerode theorem.}. 
Recall is calculated by checking the fraction of the behavior that is allowed for by the minimal log-automaton that is also allowed for by the minimal model-automaton for each projection and by averaging the recall over each projection.

\subsubsection{Continued parsing measure ($rec_F$).}
This continued parsing measure was developed in the context of the heuristic miner by Weijters et~al.\@ \cite{tonbeta166}. It abstracts from the representation of the process model by translating the Petri net into a causal matrix. This matrix defines input and output expressions for each activity, which describe possible in- and output behavior. When replaying the event log on the causal matrix, one has to check whether the corresponding input and output expressions are activated and therefore enable the execution of the activity. To calculate the continued parsing measure the number of events $e$ in the event log, as well as the number of missing activated input expressions $m$ and remaining activated output expressions $r$ are counted. Note, that the approach allows for silent transitions in the process model but excludes duplicate transitions.


\subsubsection{Eigenvalue recall ($rec_G$).}
Polyvyanyy et~al.\@ \cite{polyvyanyy-conf} introduce a framework for the definition of language quotients that guarantee several properties similar to the propositions introduced in \cite{propositions-2018}. To illustrate this framework, they apply it in the process mining context and define a recall measure. Hereby they rely on the relation between the language of a deterministic finite automaton (DFA) that describes the behavior of the model and the language of the log. In principle, recall is defined as in Definition~\ref{l2mtrprecision}. However, the measure is undefined if the language of the model or the log are infinite. Therefore, instead of using the cardinality of the languages and their intersection, the measure computes their corresponding eigenvalues and sets them in relation. To compute these eigenvalues, the languages have to be irreducible. Since this is not the case for the language of event logs, Polyvyanyy et~al.\@ \cite{polyvyanyy-conf} introduce a short-circuit measure over languages and proved that it is a deterministic measure over any arbitrary regular language.

\subsection{Existing Precision Measures}

\subsubsection{Soundness ($prec_H$).}
The notion of soundness as defined by Greco et~al.\@ \cite{GrecoTKDE2006} states that a model is precise if all possible enactments of the process have been observed in the event log. Therefore, it divides the number of unique traces in the log compliant with the process model by the number of unique traces through the model. Note, that this approach assumes the process model in the shape of a workflow net. Furthermore, it is equivalent to the baseline precision measure $prec_TB$.


\subsubsection{Simple behavioral appropriateness ($prec_I$).}
Rozinat and van der Aalst \cite{anne_confcheck_is} introduce simple behavioral appropriateness to measure the precision of process models. The approach assumes that imprecise models enable a lot of transitions during replay. Therefore, the approach computes the mean number of enabled transitions $x_i$ for each unique trace $i$ and puts it in relation to the visible transitions $T_V$ in the process model. Note, that the approach assumes a sound workflow net. However, it allows for duplicate and silent transitions in the process model.



\subsubsection{Advanced behavioral appropriateness ($prec_J$).}
In the same paper, Rozinat and van der Aalst \cite{anne_confcheck_is} define advanced behavioral appropriateness. This approach abstracts from the process model by describing the relation between activities of both the log and model with respect to whether these activities \textit{follow} and/or \textit{precede} each other. Hereby they differentiate between \textit{never}, \textit{sometimes} and \textit{always} precede/follow relations. To calculate precision the set of sometimes followed relations of the log $S^{l}_F$ and the model $S^{m}_F$ are considered, as well as their sometimes precedes relations $S^{l}_P$ and $S^{m}_P$. The fraction of sometimes follows/precedes relations of the model which are also observed by the event log defines precision. Note, that the approach assumes a sound workflow net. However, it allows for duplicate and silent transitions in the process model.


\subsubsection{ETC-one/ETC-rep ($prec_K$) and ETC-all ($prec_L$).}
Munoz-Gama and Carmona \cite{freshconf-bpm2010} introduced a precision measure which constructs an automaton that reflects the states of the model which are visited by the event log. For each state, it is evaluated whether there are activities which were allowed by the process model but not observed by the event log. These activities are added to the automaton as so-called escaping edges. Since this approach is not able to handle unfitting behavior, \cite{align-ETC2013} and \cite{wires-replay} extended the approach with a preprocessing step that aligned the log to the model before the construction of the automaton. Since it is possible that traces result in multiple optimal alignments, there are three variations of the precision measure. One can randomly pick one alignment and construct the alignment automaton based on it (ETC-one), select a representative set of multiple alignments (ETC-rep) or use all optimal alignments (ETC-all). For each variation, \cite{wires-replay} defines an approach that assigns appropriate weights to the edges of the automaton. Precision is then computed by comparing for each state of the automaton, the weighted number of non-escaping edges to the total number of edges.


\subsubsection{Behavioral specificity ($prec_M$) and Behavioral precision ($prec_N$).}
Goedertier et~al.\@ \cite{stijn-JMLR2009} introduced a precision measure based on the concept of negative events that is defined based on the concept of a confusion matrix as used in the data mining field. In this confusion matrix, the induced negative events are considered to be the ground truth and the process model is considered to be a prediction machine that predicts whether an event can or cannot occur. A negative event expresses that at a certain position in a trace, a particular event cannot occur. To induce the negative events into an event log, the traces are split in subsequences of length $k$. For each event $e$ in the trace, it is checked whether another event $e_n$ could be a negative event. Therefore the approach searches whether the set of subsequences contains a similar sequence to the one preceding $e$. If no matching sequence is found that contains $e_n$ at the current position of $e$, $e_n$ is recorded as a negative event of $e$. To check conformance the log, that was induced with negative events, is replayed on the model.

For both measures, the log that was induced with negative events is replayed on the model. Specificity and precision are measured according to their data mining definition using true positive (TP), false positive (FP) and true negative (TN) counts.

Goedertier et~al.\@ \cite{stijn-JMLR2009} ($prec_M$) defined behavioral specificity precision as \linebreak $prec_M(l,m)=\frac{TN(l,m)}{TN(l,m)+FP(l,m)}$, i.e., the ratio of the induced negative events that were also disallowed by $m$. More recently, De Weerdt et~al.\@ \cite{weerdt-F-measure_ssci_2011} gave an inverse definition, called behavioral precision ($prec_N$), as the ratio of behavior that is allowed by $m$ that does not conflict an induced negative event, i.e. $prec_N(l,m)=\frac{TP(l,m)}{TP(l,m)+FP(l,m)}$.

\subsubsection{Weighted negative event precision ($prec_O$).}
Van den Broucke et~al.\@ \cite{weighted-n-events} proposed an improvement to the approach of Goedertier et~al.\@ \cite{stijn-JMLR2009}, which assigns weights to negative events. These weights indicate the confidence of the negative events actually being negative. To compute the weight, the approach takes the sequence preceding event $e$ and searches for the matching subsequences in the event log. All events that have never followed such a subsequence are identified as negative events for $e$ and their weight is computed based on the length of the matching subsequence. To calculate precision the enhanced log is replayed on the model, similar to the approach introduced in \cite{weerdt-F-measure_ssci_2011}. However, instead of increasing the counters by 1 they are increased by the weight of the negative event. Furthermore, van den Broucke et~al.\@ \cite{weighted-n-events} also introduced a modified trace replay procedure which finds the best fitting firing sequence of transitions, taking force firing of transitions as well as paths enabled by silent transitions into account. 

\subsubsection{Projected precision ($prec_P$).}
Along with projected recall ($rec_E$) Leemans et~al.\@ \cite{sander-scalable-procmin-SOSYM} introduce projected precision. To compute precision, the approach creates a DFA which describes the conjunction of the behavior of the model and the event log. The number of outgoing edges of $DFA(m| _A)$ and  the conjunctive automaton $DFAc(l,m,A)$ are compared.
Precision is calculated for each subset of size $k$ and averaged over the number of subsets.

\subsubsection{Anti-alignment precision ($prec_Q$).}
Van Dongen et~al.\@ \cite{anti-align-bpm2016} propose a conformance checking approach based on anti-alignments. An anti-alignment is a run of a model which differs from all the traces in a log. The principle of the approach assumes that a very precise model only allows for the observed traces and nothing more. If one trace is removed from the log, it becomes the anti-alignment for the remaining log.

Therefore, trace-based precision computes an anti-alignment for each trace in the log. Then the distance $d$ between the anti-alignment and the trace $\sigma$ is computed. This is summed up for each trace and averaged over the number of traces in the log. The more precise a model, the lower the distance. 
However, the anti-alignment used for trace-based precision is limited by the length of the removed trace $\card{\sigma}$. Therefore, log-based precision uses an anti-alignment between the model and the complete log which has a length which is much greater than the traces observed in the log. 
Anti-alignment precision is the weighted combination of trace-based and log-based anti-alignment precision. Note, that the approach allows for duplicate and silent transitions in the process model.

\subsubsection{Eigenvalue precision ($prec_R$).}
Polyvyanyy et~al.\@ \cite{polyvyanyy-conf} also define a precision measure along with the Eigenvalue recall ($rec_G$). For precision, they rely on the relation between the language of a deterministic finite automaton (DFA) that describes the behavior of the model and the language of the log. To overcome the problems arising with infinite languages of the model or log, they compute their corresponding eigenvalues and set them in relation. To compute these eigenvalues, the languages have to be irreducible. Since this is not the case for the language of event logs, Polyvyanyy et~al.\@ \cite{polyvyanyy-conf} introduce a short-circuit measure over languages and proof that it is a deterministic measure over any arbitrary regular language.

\subsection{Evaluation of Existing Recall and Precision Measures}\label{eval}
Several of the existing precision measures are not able to handle non-fitting behavior and remove it by aligning the log to the model. We use a baseline approach for the alignment, which results in a deterministic event log: $l$ is the original event log, which is aligned in a deterministic manner. The resulting event log $l'$ corresponds to unique paths through the model. We use $l'$ to evaluate the propositions.

\subsubsection{Evaluation of Existing Recall Measures}
\label{sec:receval}

\begin{table}[t!]
	\centering \caption{Overview of the recall propositions that hold for the existing measures (under the assumption that $l \neq [~]$, $\traces{m} \neq \emptyset$ and $\la\ra\not\in\traces{m}$):
		\xok\ means that the proposition holds for any log and model and \xnok\ means that the proposition does not always hold.~\\} \label{tabreceval}
	
\begin{tabular}{|c|c||c|c|c|c|c|c|c|}
		\hline
		Proposition & Name & $\mi{rec}_{\mi{A}}$ & $\mi{rec}_{\mi{B}}$ & $\mi{rec}_{\mi{C}}$ & $\mi{rec}_{\mi{D}}$ & $\mi{rec}_{\mi{E}}$ & $\mi{rec}_{\mi{F}}$ & $\mi{rec}_{\mi{G}}$ \\ \hline \hline
		1 & \textbf{DetPro${}^{+}$} & \xok   &  \xnok & \xok  &  \xnok &  \xok & \xnok &  \xok \\ \hline
		2 & \textbf{BehPro${}^{+}$}  & \xok   &  \xnok & \xok  &  \xnok &  \xok & \xok &  \xok  \\ \hline
		3 & \textbf{RecPro1${}^{+}$}  &  \xnok & \xnok  & \xok   & \xnok  & \xok  & \xok &  \xok  \\ \hline
		4 & \textbf{RecPro2${}^{+}$}  &  \xok & \xok  & \xok  &  \xok & \xok  & \xok &  \xok \\ \hline
		5 & \textbf{RecPro3${}^{0}$}  &  \xnok & \xnok  & \xnok  & \xnok  & \xnok  & \xnok &  \xok \\ \hline
		6 & \textbf{RecPro4${}^{0}$}  & \xok  & \xok  & \xok  & \xok  & \xok  & \xok &  \xok \\ \hline
		7 & \textbf{RecPro5${}^{+}$}  & \xnok  & \xok  & \xok  & \xok  & \xok  &  \xnok &  \xok \\ \hline
	\end{tabular}
\end{table}

The previously presented recall measures are evaluated using the corresponding propositions. The results of the evaluation are displayed in Table~\ref{tabreceval}. To ensure the readability of this paper, only the most interesting findings of the evaluation are addressed in the following section. For full details refer to Appendix~\ref{appendixrec}.
\begin{figure}[ht]
	{
		\centering
		\includegraphics[width=0.7\textwidth]{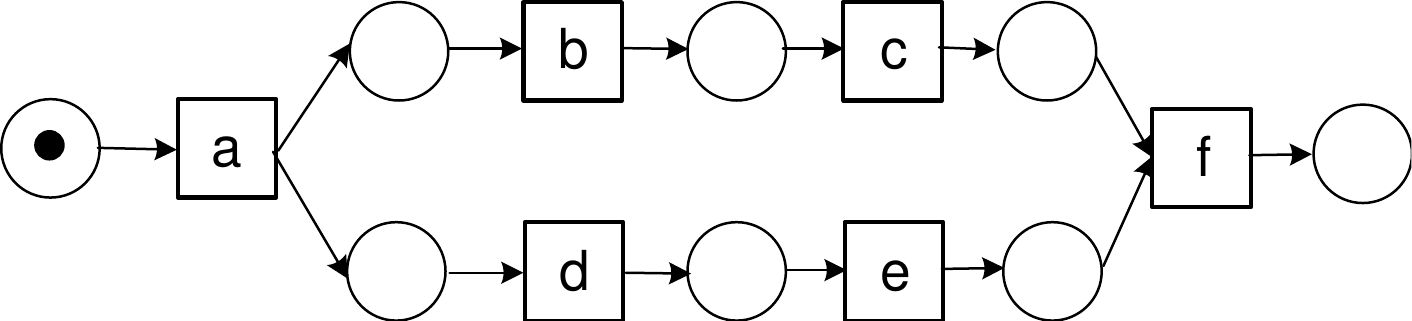}
		\caption{A process model $m4$.
		}\label{f-mod4}
	}
\end{figure}

The evaluation of the \textit{causal footprint recall measure} ($rec_A$) showed that it is deterministic and solely relies on the behavior of the process model. However, the measure violates several propositions such as \textbf{RecPro$1^{+}$}, \textbf{RecPro$3^{0}$}, and \textbf{RecPro$5^{+}$}. These violations are caused by the fact that recall records every difference between the footprint of the log and the model. Behavior that is described by the model but is not observed in the event log has an impact on recall, although Definition~\ref{defr} states otherwise.
To illustrate this, consider $m_4$ in Figure~\ref{f-mod4}, event log  $l_4 = [ \la  a,b,c,d,e,f \ra,  \la  a,b,d,c,e,f \ra ] $ and \textbf{RecPro$5^{+}$}. The traces in $l_4$ perfectly fit process model $m_4$. The footprint of $l_4$ is shown in Table~\ref{tabfootprintmain1} (b). Comparing it to the footprint of $m_4$ in Table~\ref{tabfootprintmain1} (a) shows mismatches although $l_4$ is perfectly fitting. These mismatches are caused by the fact that the log does not show all possible behavior of the model and, therefore, the footprint cannot completely detect the parallelism of the model. Consequently 10 of 36 relations of the footprint represent mismatches: $rec_A(l_4,m_4)=1-\frac{10}{36}= 0.72 \neq 1$. Van der Aalst mentions in \cite{process-mining-book-2016} that checking conformance using causal footprints is only meaningful if the log is complete in term of directly followed relations. Moreover, the measure also includes precision and generalization aspects, next to recall.

\begin{table}[t!]
	\centering \caption{The causal footprints of $m_4$ (a), $l_4$ (b). Mismatching relations are marked in \textcolor{red}{red}.\looseness=-1} \label{tabfootprintmain1}
	\subfloat[][]{\begin{tabular}{ccccccc}
			& \textbf{a} & \textbf{b}      & \textbf{c} & \textbf{d} & \textbf{e} & \textbf{f} \\
			\textbf{a} & \#         & $\rightarrow$ &  \#          & $\rightarrow$           &   \#         &   \#         \\
			\textbf{b} &     $\leftarrow$         & \#              &   $\rightarrow$         &  $||$          &   $||$        &   \#         \\
			\textbf{c} &     \#        &      $\leftarrow$             & \#         &      $||$    &   $||$     &      $\rightarrow$      \\
			\textbf{d} &    $\leftarrow$     &     $||$               &    $||$         & \#         &   $\rightarrow$         &     \#       \\
			\textbf{e} &     \#       &    $||$            &       $||$     &       $\leftarrow$       & \#         &        $\rightarrow$    \\
			\textbf{f} &  \#         &      \#            &     $\leftarrow$         &     \#        &    $\leftarrow$          & \#
	\end{tabular}}
	\quad
	\subfloat[][]{\begin{tabular}{ccccccc}
		& \textbf{a} & \textbf{b}      & \textbf{c} & \textbf{d} & \textbf{e} & \textbf{f} \\
		\textbf{a} & \#         & $\rightarrow$ &  \#          & {\color[HTML]{FE0000}\# }          &   \#          &   \#          \\
		\textbf{b} &     $\leftarrow$         & \#              &   $\rightarrow$         &   {\color[HTML]{FE0000}$\rightarrow$}         &   {\color[HTML]{FE0000}\#  }         &   \#          \\
		\textbf{c} &     \#          &      $\leftarrow$             & \#         &        $||$      & {\color[HTML]{FE0000}$\rightarrow$}         &      {\color[HTML]{FE0000}\#  }        \\
		\textbf{d} &   {\color[HTML]{FE0000}\#  }         &   {\color[HTML]{FE0000}$\leftarrow$}               &    $||$          & \#         &   $\rightarrow$         &     \#       \\
		\textbf{e} &   \#        & {\color[HTML]{FE0000}\#}              &       {\color[HTML]{FE0000}$\leftarrow$}        &       $\leftarrow$       & \#         &        $\rightarrow$    \\
		\textbf{f} &  \#         &      \#            &    {\color[HTML]{FE0000}\#}        &    \#         &    $\leftarrow$          & \#
	\end{tabular}}
\end{table}

In comparison, recall based on \textit{token replay} ($rec_B$) depends on the path taken through the model. Due to duplicate activities and silent transitions, multiple paths through a model can be taken when replaying a single trace. Different paths can lead to different numbers of produced, consumed, missing and remaining tokens. Therefore, the approach is neither deterministic nor independent from the structure of the process model and, consequently, violates \textbf{RecPro$1^{+}$}. The \textit{continued parsing measure} ($rec_F$) builds on a similar replay principle as token-based replay and also violates \textbf{DetPro$^{+}$}. However, the approach translates the process model into a causal matrix and is therefore independent of its structure.

\begin{figure}[t]
	{
		\centering
		\includegraphics[width=0.8\textwidth]{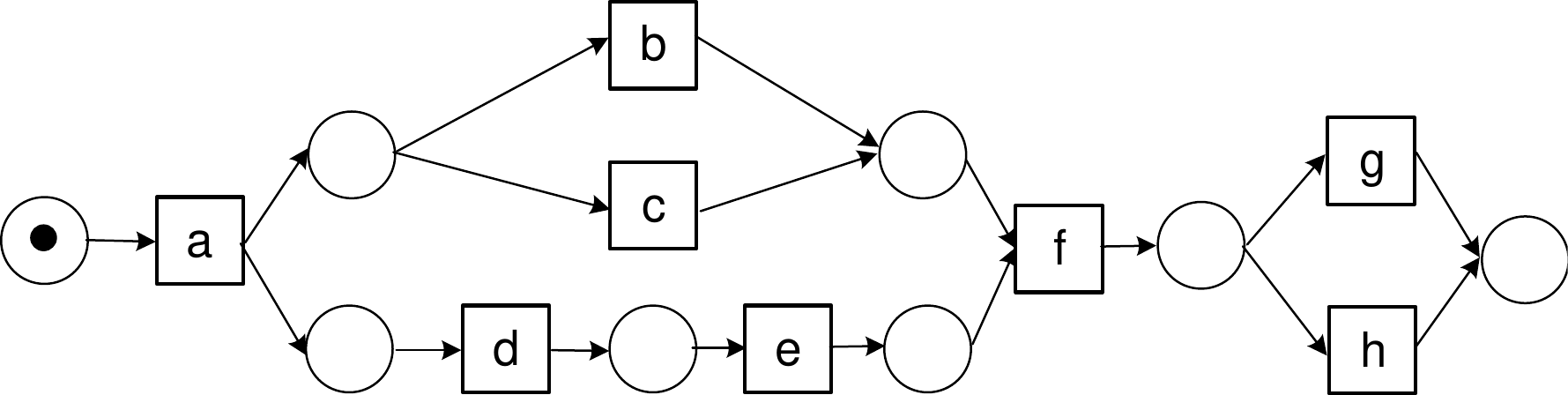}
		\caption{Petri net $m_5$
		}\label{f-mod5}
	}
\end{figure}

Table~\ref{tabreceval} also shows that most measures violate \textbf{RecPro$3^{0}$}.
This is caused by the fact, that we define non-fitting behavior in this paper on a trace level: traces either fit the model or they do not. However, the evaluated approaches measure non-fitting behavior on an event level. A trace consists of fitting and non-fitting events. In cases where the log contains traces with a large number of deviating events, recall can be improved by adding non-fitting traces which contain several fitting and only a few deviating events.
To illustrate this, consider \textit{token replay} ($rec_B$), process model $m_5$ in Figure~\ref{f-mod5},  $l_5 = [ \la a,b,f,g]$ and $l_6 = l_5 \uplus [ \la  a,d,e,f,g \ra]$. The log $l_5$ is not perfectly fitting and replaying it on the model results in 6 produced and 6 consumed tokens, as well as 1 missing and 1 remaining token. $rec_B(l_5, m_5) = \frac{1}{2}(1-\frac{1}{6}) + \frac{1}{2}(1-\frac{1}{6}) = 0.833$. Event log $l_6$ was created by adding non-fitting behavior to $l_5$. Replaying $l_6$ on $m_5$ results in $p = c = 13$, $r = m = 2$ and $rec_B(l_7, m_6) = \frac{1}{2}(1-\frac{2}{13}) + \frac{1}{2}(1-\frac{2}{13}) = 0.846$. Hence, the additional unfitting trace results in proportionally more fitting events than deviating ones which improves recall: $rec_B(l_6, m_6) < rec_B(l_7, m_6)$.

To overcome the problems arising with the differences between trace-based and event-based fitness, one could alter the definition of \textbf{RecPro$3^{0}$} by requiring, that the initial log $l_1$ only contains fitting behavior ($\traces{l_1} \subseteq \traces{m}$). However, to stay within the scope of this paper, we decide to use the propositions as defined in~\cite{propositions-2018} and keep this suggestion for future work.

\subsubsection{Evaluation of Existing Precision Measures }
\label{sec:preceval}

\begin{table}[t!]
	\centering \caption{Overview of the precision propositions that hold for the existing measures (under the assumption that $l \neq [~]$, $\traces{m} \neq \emptyset$ and $\la\ra\not\in\traces{m}$):
		\xok\ means that the proposition holds for any log and model and \xnok\ means that the proposition does not always hold.~\\ } \label{tabpreceval}
	\begin{tabular}{|c|c||c|c|c|c|c|c|c|c|c|c|c|}
		\hline
		Prop. & Name & $\mi{prec}_{\mi{H}}$ & $\mi{prec}_{\mi{I}}$ & $\mi{prec}_{\mi{J}}$ & $\mi{prec}_{\mi{K}}$ & $\mi{prec}_{\mi{L}}$ & $\mi{prec}_{\mi{M}}$ & $\mi{prec}_{\mi{N}}$ & $\mi{prec}_{\mi{O}}$ & $\mi{prec}_{\mi{P}}$ & $\mi{prec}_{\mi{Q}}$ & $\mi{prec}_{\mi{R}}$ \\ \hline \hline
		1 & \textbf{DetPro${}^{+}$} & \xnok   &  \xnok & \xnok  &  \xnok &  \xok & \xnok & \xnok & \xnok & \xok & \xok  & \xok \\ \hline
		2 & \textbf{BehPro${}^{+}$}  & \xok   &  \xnok & \xnok  &  \xnok &  \xnok & \xnok & \xnok & \xnok & \xok & \xok & \xok \\ \hline
		8 & \textbf{PrecPro1${}^{+}$}  & \xok  & \xnok  & \xnok  & \xnok  &  \xnok   & \xnok & \xnok & \xnok & \xnok & \xok & \xok \\ \hline
		9 & \textbf{PrecPro2${}^{+}$}  &  \xok & \xnok  & \xok  & \xnok  &  \xnok   & \xnok & \xnok & \xnok & \xnok & \xnok & \xok \\ \hline
		10 & \textbf{PrecPro3${}^{0}$}  & \xok  & \xnok  & \xnok  & \xnok  &  \xnok   & \xnok & \xnok & \xnok & \xnok &\xnok & \xok \\ \hline
		11 &  \textbf{PrecPro4${}^{0}$}  & \xok  & \xok  & \xok  & \xok  &  \xok  & \xok & \xok & \xok & \xok & \xok & \xok \\ \hline
		12 & \textbf{PrecPro5${}^{+}$}  & \xok & \xnok  & \xok  & \xnok  &  \xnok  & \xok & \xok & \xok & \xok & \xok & \xok \\ \hline
		13 & \textbf{PrecPro6${}^{0}$}  &  \xok & \xnok   & \xok  & \xnok  & \xnok   & \xok & \xok & \xok & \xok & \xok & \xok \\ \hline
	\end{tabular}
\end{table}

The previously presented precision measures are evaluated using the corresponding propositions. The results of the evaluation are displayed in Table~\ref{tabpreceval}. To ensure the readability of this paper, only the most interesting findings of the evaluation are addressed in the following section. For full details, we refer to Appendix~\ref{appendixprec}.

The evaluation showed that several measures violate the determinism \textbf{DetPro${}^{+}$} proposition. For example, the soundness measure ($prec_H$) solely relies on the number of unique paths of the model $\card{\traces{m}}$ and unique traces in the log that comply with the process model $\card{\traces{l} \cap \traces{m}}$. Hence, precision is not defined if the model has infinite possible paths. Additionally to \textbf{DetPro${}^{+}$}, behavioral specificity ($rec_M$) and behavioral precision ($rec_N$) also violate \textbf{BehPro${}^{+}$}. If during the replay of the trace duplicate or silent transitions are encountered, the approach explored which of the available transitions enables the next event in the trace. If no solution is found, one of the transitions is randomly fired, which can lead to different recall values for traces with the same behavior. 

Table~\ref{tabpreceval} shows that simple behavioral appropriateness ($prec_I$) violates all but one of the propositions. One of the reason is that it relies on the average number of enabled transitions during replay. Even when the model allows for all exactly observed behavior (and nothing more), precision is not maximal when the model is not strictly sequential.
Advanced behavioral appropriateness ($prec_J$) overcomes these problems by relying on follow relations. However, it is not deterministic and depends on the structure of the process model.

The results presented in \cite{Niek-IPL2018-imprecision} show that ETC precision ($prec_K$ and  $prec_L$), weighted negative event precision ($prec_O$) and projected precision ($prec_P$) violate \textbf{PrecPro1$^{+}$}. Additionally, all remaining measures aside from anti-alignment precision ($prec_Q$) and eigenvalue precision ($prec_R$) violate the proposition. The proposition states that removing behavior from a model that does not happen in the event log cannot lower precision. Consider, projected precision ($prec_P$) and a model with a length-one-loop. We remove behavior from the model by restricting the model to only execute the looping activity twice. This changes the DFA of the model since future behavior now depends on how often the looping activity was executed: the DFA contains different states for each execution. If these states show a low local precision, overall precision decreases. Furthermore, \cite{Niek-IPL2018-imprecision} showed that ETC precision ($prec_K$ and  $prec_L$), projected precision ($prec_P$) and anti-alignment precision ($prec_Q$) also violate \textbf{PrecPro2$^{+}$}.

\begin{figure}[t]
	{
		\centering
		\subfloat[][]{\includegraphics[width=0.5\textwidth]{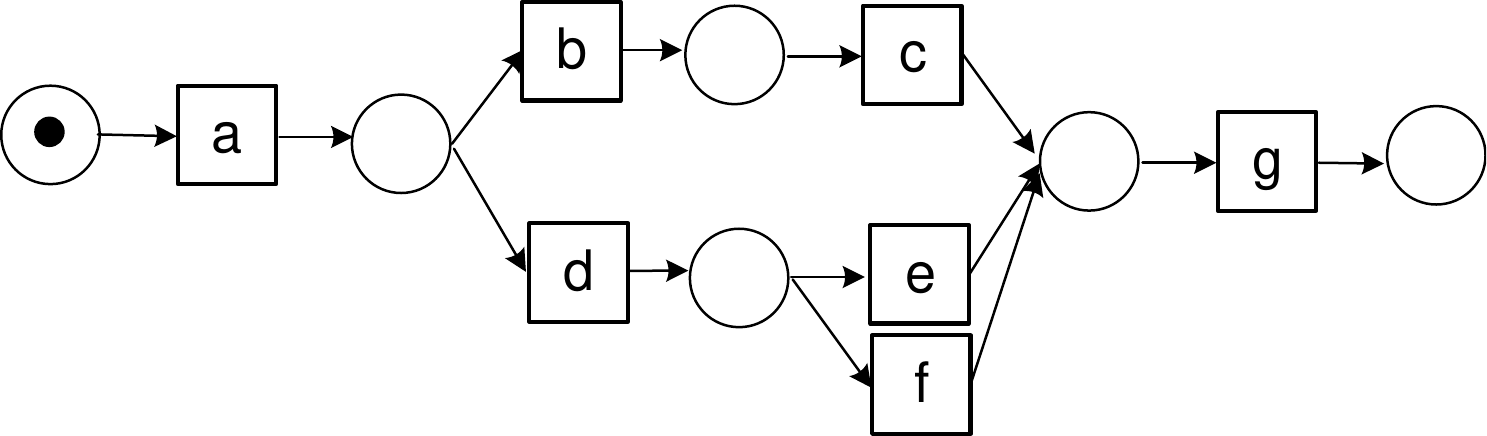}}
		\quad
		\subfloat[][]{\includegraphics[width=0.4\textwidth]{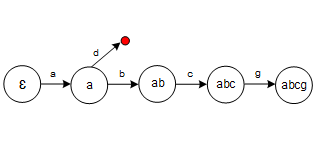}}
		\caption{Petri net $m_6$ (a) and the alignment automaton describing the state space of $\sigma = \la a,b,c,g \ra$ (b)
		}\label{f-mod6}
	}
\end{figure}

In general, looking at Table~\ref{tabpreceval} shows that all precision measures, except for soundness ($prec_H$) and eigenvalue precision ($prec_R$) violate \textbf{PrecPro3$^{0}$}, which states that adding unfitting behavior to the event log should not change precision. However, for example, all variations of the ETC-measure ($prec_K$, $prec_L$) align the log before constructing the alignment automaton. Unfitting behavior can be aligned to a trace that was not seen in the log before and introduce new states to the automaton. Consider process model $m_6$, together with trace $\sigma = \la a,b,c,g \ra$ and its alignment automaton displayed in Figure~\ref{f-mod6}. Adding the unfitting trace $\la a,d,g \ra$ could result in the aligned trace $\la a,d,e,g \ra$ or $\la a,d,f,g \ra$. Both aligned traces introduce new states into the alignment automaton, alter the weights assigned to each state and, consequently, change precision.
Weighted negative precision ($prec_O$) also violates this proposition. The measure accounts for the number of negative events that actually could fire during trace replay ($\text{FP}$). These false positives are caused by behavior that is shown in the model but not observed in the log. As explained in the context of \textbf{RecPro$3^{0}$}, although the trace is not fitting when considered as a whole, certain parts of the trace can fit the model. These parts can possibly represent the previously missing behavior in the event log that leads to the wrong classification of negative events. Adding these traces will, therefore, lead to a decrease in false positives and changes precision. $FP(l_1,m) > FP(l_2,m)$ and $\frac{TP(l_1,m)}{(TP(l_1,m) + FP(l_1,m))} < \frac{TP(l_2,m)}{(TP(l_2,m) + FP(l_2,m))}$.

Table~\ref{tabpreceval} shows that $prec_I$, $prec_K$ and $prec_L$ violate proposition \textbf{PrecPro6$^{0}$}, which states that if all modeled behavior was observed, precision should be maximal and unfitting behavior cannot effect precision. $prec_I$ only reports maximal precision if the model is strictly sequential and both ETC measures ($prec_K$ and $prec_L$) can run into problems with models containing silent or duplicate transitions.\looseness=-1

The ETC ($prec_K$, $prec_L$) and anti-alignment measures ($prec_Q$) form a special group of measures as they are unable to handle unfitting behavior without pre-processing unfitting traces and aligning them to the process model. Accordingly, we evaluate the conformance measure based on this aligned log. The evaluation of \textbf{PrecPro3$^{0}$} and the ETC measure ($prec_K$, $prec_L$) is an example of the alignment of the log resulting in a violation. However, there are also cases where the proposition only holds because of this alignment. Consider, for example, anti-alignment precision ($prec_Q$) and proposition \textbf{PrecPro6$^{0}$}. By definition, an anti-alignment will always fit the model. Consequently, when computing the distance between the unfitting trace and the anti-alignment it will never be minimal. However, after aligning the log, it exactly contains the modeled behavior, precision is maximal and the proposition holds.

\section{Generalization}\label{sec:gen}
Generalization is a challenging  concept to define, in contrast to recall and precision. As a result, there are different viewpoints within the process mining community on what generalization precisely means.
The main reason for this is, that generalization needs to reason about behavior that was not observed in the event log and establish its relation to the model.

The need for a generalization dimension stems from the fact that, given a log, a model can be fitting and precise, but be overfitting.
The algorithm that simply creates a model $m$ such that $\traces{m}=\{t\in l\}$ is useless because it is simply enumerating the event log.
Consider an unknown process.
Assume we observe the first four traces $l_1 = [ \la a,b,c \ra, \allowbreak \la b,a,c \ra, \allowbreak \la a,b,d \ra, \allowbreak \la b,a,d \ra ]$.
Based on this we may construct the model $m_3$ in Figure~\ref{f-mod2} with $\traces{m_3}=\{ \la a,b,c \ra, \allowbreak \la b,a,c \ra, \allowbreak \la a,b,d \ra, \allowbreak \la b,a,d \ra \}$.
This model allows for all the traces in the event log and nothing more. However, because the real underlying process in unknown, this model may be overfitting event log $l_1$.
Based on just four example traces we cannot be confident that the model $m_3$ in Figure~\ref{f-mod2} will be able to explain future behavior of the process.
The next trace may as well be $\la a,c \ra$ or $\la a,b,b,c \ra$.
Now assume that we observe the same process for a longer time and consider the first 100 traces (including the initial four):
$l_2 = [ \la a,b,c \ra^{25}, \allowbreak \la b,a,c \ra^{25}, \allowbreak \la a,b,d \ra^{25}, \allowbreak \la b,a,d \ra^{25} ]$.
After observing 100 traces, we are more confident that model $m_3$ in Figure~\ref{f-mod2} is the right model.
Intuitively, the probability that the next case will have a trace not allowed by $m_3$ gets smaller.
Now assume that we observe the same process for an even longer time and obtain the event log
$l_2 = [ \la a,b,c \ra^{53789}, \allowbreak \la b,a,c \ra^{48976}, \allowbreak \la a,b,d \ra^{64543}, \allowbreak \la b,a,d \ra^{53789}]$.
Although we do not know the underlying process, intuitively, the probability that the next case will have a trace not allowed by $m_3$ is close to 0.
This simple example shows that recall and precision are not enough for conformance checking. We need a generalization notion to address the risk of overfitting example data.

It is difficult to reason about generalization because this refers to unseen cases. Van der Aalst et~al.\@~\cite{wires-replay} was the first to quantify generalization.
In \cite{wires-replay}, each event is seen as an observation of an activity $a$ in some state $s$.
Suppose that state $s$ is visited $n$ times and that $w$ is the number of different activities
observed in this state. Suppose that $n$ is very large and $w$ is very small, then it is
unlikely that a new event visiting this state will correspond to an activity not seen
before in this state. However, if $n$ and $w$ are of the same order of magnitude, then it
is more likely that a new event visiting state $s$ will correspond to an activity not seen
before in this state. This reasoning is used to provide a generalization metric.
This estimate can be derived under the Bayesian assumption that there is an unknown number of possible activities in state $s$ and that probability
distribution over these activities follows a multinomial distribution.

It is not easy to develop an approach that accurately measures generalization.
Therefore, some authors define generalization using the notion of a ``system'' (i.e., a model of the real underlying process).
The system refers to the real behavior of the underlying process that the model tries to capture.
This can also include the context of the process such as the organization or rules.
For example, employees of a company might exceptionally be allowed to deviate from the defined process model in certain situations~\cite{BPM-jans}.
In this view, \emph{system fitness} measures the fraction of the behavior of the system that is captured by the model and \emph{system precision} measures how much of the behavior of the model is part of the system.
Buijs et~al.\@~\cite{joos-4dim-ijcis-2014} link this view to the traditional understanding of generalization. They state that both system fitness and system precision are difficult to obtain under the assumption that the system is unknown. Therefore, state-of-the-art discovery algorithms assume that the process model discovered from an event log does not contain behavior outside of the system. In other words, they assume system precision to be 1. Given this assumption, system fitness can be seen as generalization~\cite{joos-4dim-ijcis-2014}. Janssenswillen et~al.\@~\cite{BPM-jans} agree that in this comparison between the system and the model, especially the system fitness, in fact is what defines generalization. Furthermore, Janssenswillen and Depaire~\cite{Janssenswillen2018} demonstrated the differences between the traditional and the system-based view on conformance checking by showing that state-of-the-art conformance measures cannot reliably assess the similarity between a process model and the underlying system.

Although capturing the unobserved behavior by assuming a model of the system is a theoretically elegant solution,
practical applicability of this solution is hindered by the fact that is often impossible to retrieve full knowledge about the system itself.
Furthermore, \cite{propositions-2018} showed the importance of trace probabilities in process models.
To accurately represent reality, the system would also need to include probabilities for each of its traces.
However, to date, there is only one conformance measure that can actually support probabilistic process models \cite{sander-BPM-forum-2019}.
This approach uses the Earth Movers' distance which measures the effort to transform the distributions of traces of the event log into the
distribution of traces of the model.

Some people would argue that one should use cross-validation (e.g., $k$-fold checking). However, this is a very different setting.
Cross validation aims to estimate the quality of a discovery approach and not the quality of a given model given an event log.
Of course, one could produce multiple process models using fragments of the event log and compare them.
However, such forms of cross-validation evaluate the quality of the discovery technique and are unrelated to generalization.

For these reasons, we define generalization in the traditional sense.

\begin{definition}[Generalization]
A \emph{generalization measure} $\mi{gen} \in \ulogs \times \umodels \rightarrow [0,1]$ aims to quantify the probability that new unseen cases will fit the model.\footnote{Note that the term ``probability'' is used here in an informal manner. Since we only have example observations and no knowledge of the underlying (possibly changing) process, we cannot compute such a probability. Of course, unseen cases can have traces that have been observed before.}
\end{definition}

This definition assumes that a process generates a stream of newly executed cases.
The more traces that are fitting and the more redundancy there is in the event, the more certain one can be that the next case will have a trace that fits the model.
Note that we deliberately do not formalize the notion of probability, since in real-life we cannot know the real process.
Also phenomena like concept drift and contextual factors make it unrealistic to reason about probabilities in a formal sense.

Based on this definition, we present a set of propositions. Note that 
we do not claim our set of propositions to be complete and invite other researchers who represent a different viewpoint on generalization to contribute to the discussion.

\subsection{Generalization Propositions}
\label{sec:geneprop}

Generalization ($\mi{gen} \in \ulogs \times \umodels \rightarrow [0,1]$) aims to quantify the probability that new unseen cases will fit the model.
This conformance dimension is a bit different than the two previously discussed conformance dimensions because it reasons about future \emph{unseen} cases (i.e., not yet in the event log).
If the recall is good and the log is complete with lots of repeating behavior, then future cases will most likely fit the model.
Analogous to recall, model extensions cannot lower generalization (\textbf{GenPro1${}^{+}$}), extending the log with fitting behavior cannot lower generalization (\textbf{GenPro2${}^{+}$}), and
extending the log with non-fitting behavior cannot improve generalization (\textbf{GenPro3${}^{0}$}).

 \begin{proposition}[\textbf{GenPro1${}^{+}$}] \label{GenPro1}
    For any $l \in \ulogs$ and $m_1,m_2 \in \umodels$ such that $\traces{m_1} \subseteq \traces{m_2}$: $\mi{gen}(l,m_1) \leq \mi{gen}(l,m_2)$.
  \end{proposition}

Similar to recall, this proposition implies \textbf{BehPro${}^{+}$}. Generalization measures violating \textbf{BehPro${}^{+}$} also violate \textbf{GenPro1${}^{+}$}.

  \begin{proposition}[\textbf{GenPro2${}^{+}$}] \label{GenPro2}
    For any $l_1,l_2,l_3 \in \ulogs$ and $m \in \umodels$ such that
    $l_2 = l_1 \uplus l_3$ and $\traces{l_3} \subseteq \traces{m}$: $\mi{gen}(l_1,m) \leq \mi{gen}(l_2,m)$.
  \end{proposition}

  \begin{proposition}[\textbf{GenPro3${}^{0}$}] \label{GenPro3}
    For any $l_1,l_2,l_3 \in \ulogs$ and $m \in \umodels$ such that
    $l_2 = l_1 \uplus l_3$ and $\traces{l_3} \subseteq \compltr{m}$: $\mi{gen}(l_1,m) \geq \mi{gen}(l_2,m)$.
  \end{proposition}

Duplicating the event log does not necessarily influence recall and precision. According to propositions \textbf{RecPro4${}^{0}$} and \textbf{PrecPro4${}^{0}$} this should have no effect on recall and precision.
However, making the event log more redundant, should have an effect on generalization.
For fitting logs, adding redundancy without changing the distribution can only improve generalization (\textbf{GenPro4${}^{+}$}).
For non-fitting logs, adding redundancy without changing the distribution can only lower generalization (\textbf{GenPro5${}^{+}$}).
Note that \textbf{GenPro4${}^{+}$} and \textbf{GenPro5${}^{+}$} are special cases of \textbf{GenPro6${}^{0}$} and \textbf{GenPro7${}^{0}$}.
\textbf{GenPro6${}^{0}$} and \textbf{GenPro7${}^{0}$} consider logs where some traces are fitting and others are not.
For a log where more than half of the traces is fitting, duplication can only improve generalization (\textbf{GenPro6${}^{0}$}).
For a log where more than half of the traces is non-fitting, duplication can only lower generalization (\textbf{GenPro7${}^{0}$}).

  \begin{proposition}[\textbf{GenPro4${}^{+}$}] \label{GenPro4}
    For any $l \in \ulogs$, $m \in \umodels$, and $k\geq 1$ such that $\traces{l} \subseteq \traces{m}$: $\mi{gen}(l^k,m) \geq \mi{gen}(l,m)$.
  \end{proposition}

    \begin{proposition}[\textbf{GenPro5${}^{+}$}] \label{GenPro5}
    For any $l \in \ulogs$, $m \in \umodels$, and $k\geq 1$ such that $\traces{l} \subseteq \compltr{m}$: $\mi{gen}(l^k,m) \leq \mi{gen}(l,m)$.
  \end{proposition}

    \begin{proposition}[\textbf{GenPro6${}^{0}$}] \label{GenPro6}
    For any $l \in \ulogs$, $m \in \umodels$, and $k\geq 1$ such that most traces are fitting ($\card{[t\in l \mid t \in  \traces{m}]} \geq \card{[t\in l \mid t \not\in \traces{m}]}$): $\mi{gen}(l^k,m) \geq \mi{gen}(l,m)$.
  \end{proposition}

    \begin{proposition}[\textbf{GenPro7${}^{0}$}] \label{GenPro7}
    For any $l \in \ulogs$, $m \in \umodels$, and $k\geq 1$ such that most traces are non-fitting ($\card{[t\in l \mid t \in \traces{m}]} \leq \card{[t\in l \mid t \not\in \traces{m}]}$): $\mi{gen}(l^k,m) \leq \mi{gen}(l,m)$.
  \end{proposition}

When the model allows for any behavior, clearly the next case will also be fitting (\textbf{GenPro8${}^{0}$}).
Nevertheless, it is marked as controversial because the proposition would also need to hold for an empty event log.

 \begin{proposition}[\textbf{GenPro8${}^{0}$}] \label{GenPro8}
    For any $l \in \ulogs$ and $m \in \umodels$ such that
    $\traces{m} = \utraces$: $\mi{gen}(l,m)=1$.
  \end{proposition}

\subsection{Existing Generalization Measures}
The following sections introduce several state-of-the-art generalization measures, before they will be evaluated using the corresponding propositions.

\subsubsection{Alignment generalization ($gen_S$).}
Van der Aalst et~al.\@ \cite{wires-replay} also introduce a measure for generalization. This approach considers each occurrence of a given event $e$ as observation of an activity in some state $s$. The approach is parameterized by a $state_M$ function that maps events onto states in which they occurred. For each event $e$ that occurred in state $s$ the approach counts how many different activities $w$ were observed in that state. Furthermore, it counts the number of visits $n$ to this state. Generalization is high if $n$ is very large and $w$ is small, since in that case, it is unlikely that a new trace will correspond to unseen behavior in that state.



\subsubsection{Weighted negative event generalization ($gen_T$).}
Aside from improving the approach of Goedertier et~al.\@ \cite{stijn-JMLR2009} van den Broucke et~al.\@ \cite{weighted-n-events} also developed a generalization measure based on weighted negative events.
It defines allowed generalizations $AG$ which represent events, that could be replayed without errors and confirm that the model is general and disallowed generalizations $DG$ which are generalization events, that could not be replayed correctly. If during replay a negative event $e$ is encountered that actually was enabled the $AG$ value is increased by $1 - weight(e)$. Similarly, if a negative event is not enabled the $DG$ value is increased by $1 - weight(e)$. The more disallowed generalizations are encountered during log replay the lower generalization.

\subsubsection{Anti-alignment generalization ($gen_U$).}
Van Dongen et~al.\@ \cite{anti-align-bpm2016} also introduce an anti-alignment generalization and build on the principle that with a generalizing model, newly seen behavior will introduce new paths between the states of the model, however no new states themselves. Therefore, they define a recovery distance $d_{rec}$ which measures the maximum distance between the states visited by the log and the states visited by the anti-alignment $\gamma$. A perfectly generalizing model according to van Dongen et~al.\@ \cite{anti-align-bpm2016} has the maximum distance to the anti-alignment with minimal recovery distance. Similar to recall they define trace-based and log-based generalization. 
Finally, anti-alignment generalization is the weighted combination of trace-based and log-based anti-alignment generalization.

\subsection{Evaluation of Existing Generalization Measures}
\label{sec:geneval}
\begin{table}[t!]
	\centering \caption{An overview of the generalization propositions that hold for the measures: (assuming $l \neq [~]$, $\traces{m} \neq \emptyset$ and $\la\ra\not\in\traces{m}$):
		\xok\ means that the proposition holds for any log and model and \xnok\ means that the proposition does not always hold.~\\ } \label{tabgeneval}
	\begin{tabular}{|c|c||c|c|c|}
		\hline
		Proposition & Name & $\mi{gen}_{\mi{S}}$ & $\mi{gen}_{\mi{T}}$ & $\mi{gen}_{\mi{U}}$  \\ \hline \hline
		1 & \textbf{DetPro${}^{+}$} & \xok   &  \xnok & \xok    \\ \hline
		2 & \textbf{BehPro${}^{+}$}  & \xok   &  \xnok & \xnok   \\ \hline
		14 & \textbf{GenPro1${}^{+}$}  & \xnok  & \xnok  &   \xnok  \\ \hline
		15 & \textbf{GenPro2${}^{+}$}  & \xnok   & \xnok  & \xnok   \\ \hline
		16 & \textbf{GenPro3${}^{0}$}  & \xnok  & \xnok   & \xnok   \\ \hline
		17 & \textbf{GenPro4${}^{+}$}  & \xok  & \xok  &  \xok  \\ \hline
		18 & \textbf{GenPro5${}^{+}$}  &  \xnok  & \xok   &  \xok   \\ \hline
		19 & \textbf{GenPro6${}^{0}$}  & \xok  & \xok  & \xok \\ \hline
		20 & \textbf{GenPro7${}^{0}$}  &  \xnok  & \xok   &  \xok  \\ \hline
		21 & \textbf{GenPro8${}^{0}$}  & \xnok   & \xok   &  \xnok   \\ \hline
	\end{tabular}
\end{table}

The previously presented generalization measures are evaluated using the corresponding propositions. The results of the evaluation are displayed in Table~\ref{tabgeneval}. To improve the readability of this paper, only the most interesting findings of the evaluation are addressed in the following section. For full details refer to Appendix~\ref{appendixgen}.

Table~\ref{tabgeneval} displays that alignment based generalization ($gen_S$) violates several propositions. Generalization is not defined if there are unfitting traces since they cannot be mapped to states of the process model. Therefore, unfitting event logs should be aligned to fit to the model before calculating generalization. Aligning a non-fitting log and duplicating it will result in more visits to each state visited by the log. Therefore, adding non-fitting behavior increases generalization and violates the propositions \textbf{GenPro3$^{0}$}, \textbf{GenPro5$^{+}$} and \textbf{GenPro7$^{0}$}. 

In comparison, weighted negative event generalization ($gen_T$) is robust against the duplication of the event log, even if it contains non-fitting behavior. However, this measure violates \textbf{DetPro$^{+}$}, \textbf{BehPro$^{+}$}, \textbf{GenPro1$^{+}$}, \textbf{GenPro2$^{+}$}  and \textbf{GenPro3$^{0}$}, which states that extending the log with non-fitting behavior cannot improve generalization. However, in this approach, negative events are assigned a weight which indicates how certain the log is about these events being negative ones. Even though the added behavior is non-fitting it might still provide evidence for certain negative events and therefore increase their weight. If these events are then not enabled during log replay the value for disallowed generalizations (DG) decreases $DG(l_1,m) < DG(l_2,m)$ and generalization improves: $\frac{AG(l_1,m)}{AG(l_1,m) + DG(l_1,m)} < \frac{AG(l_2,m)}{AG(l_2,m) + DG(l_2,m)}$.

Table~\ref{tabgeneval} shows that anti-alignment generalization ($gen_U$) violates several propositions. The approach considers markings of the process models as the basis for the generalization computation which violates the behavioral proposition. Furthermore, the measure cannot handle if the model displays behavior that has not been observed in the event log. If the unobserved model behavior and therefore also the anti-alignment introduced a lot of new states which were not visited by the event log, the value of the recovery distance increases and generalization is lowered. This clashes with propositions \textbf{GenPro1$^{+}$} and \textbf{GenPro8$^{+}$}. Finally, the approach also excludes unfitting behavior from its scope. Only after aligning the event log, generalization can be computed. As a result, the measure fulfills \textbf{GenPro5$^{+}$}, \textbf{GenPro6$^{0}$} and \textbf{GenPro7$^{0}$}, but violates \textbf{GenPro3$^{0}$}.

\section{Conclusion}\label{sec:concl}

With the process mining field maturing and more commercial tools becoming available \cite{gartnerpm2018},
there is an urgent need to have a set of agreed-upon measures to determine the quality of discovered processes models. We have revisited the 21 conformance propositions introduced in~\cite{propositions-2018} and illustrated their relevance by applying them to baseline measures. Furthermore, we used the propositions to evaluate currently existing conformance measures. This evaluation uncovers large differences between existing conformance measures and the properties that they possess in relation to the propositions.
It is surprising that seemingly obvious requirements are not met by today's conformance measures. However, there are also measures that do meet all the propositions.

It is important to note that we do not consider the set of propositions to be complete. Instead, we consider them to be an initial step to start the discussion on what properties are to be desired from conformance measures, and we encourage others to contribute to this discussion. Moreover, we motivate researchers to use the conformance propositions as design criteria for the development of novel conformance measures.

One relevant direction of future work is in the area of conformance propositions that have a more fine-grained focus than the trace-level, i.e., that distinguish between \emph{almost fitting} and \emph{completely non-fitting} behavior. Another relevant area of future work is in the direction of probabilistic conformance measures, which take into account branching probabilities in models, and their desired properties.


\subsubsection*{Acknowledgements} We thank the Alexander von Humboldt (AvH) Stiftung for supporting our research.

\bibliographystyle{plain}
\bibliography{lit}

\appendix
\section{Appendix}\label{appendix}

\subsection{Evaluation results of the baseline conformance measures }\label{appendixtab}
\begin{table}[h!]
	\centering \caption{Overview of the conformance propositions that hold for the three baseline measures (under the assumption that $l \neq [~]$, $\traces{m} \neq \emptyset$ and $\la\ra\not\in\traces{m}$):
		\xok\ means that the proposition holds for any log and model and \xnok\ means that the proposition does not always hold.} \label{tabeval}
	\begin{tabular}{|c|c||c|c|c|}
		\hline
		Proposition & Name & $\mi{rec}_{\mi{TB}}$ & $\mi{rec_{\mi{FB}}}$ & $\mi{prec}_{\mi{FB}}$  \\ \hline \hline
		1 & \textbf{DetPro${}^{+}$} & \xok   &  \xok & \xnok     \\ \hline
		2 & \textbf{BehPro${}^{+}$}  & \xok   &  \xok & \xok    \\ \hline
		3 & \textbf{RecPro1${}^{+}$}  &  \xok & \xok  &        \\ \hline
		4 & \textbf{RecPro2${}^{+}$}  &  \xok & \xok  &        \\ \hline
		5 & \textbf{RecPro3${}^{0}$}  &  \xok & \xok  &        \\ \hline
		6 & \textbf{RecPro4${}^{0}$}  & \xok  & \xok  &        \\ \hline
		7 & \textbf{RecPro5${}^{+}$}  & \xok  & \xok  &        \\ \hline
		8 & \textbf{PrecPro1${}^{+}$}  &   &   & \xok    \\ \hline
		9 & \textbf{PrecPro2${}^{+}$}  &   &   & \xok    \\ \hline
		10 & \textbf{PrecPro3${}^{0}$}  &   &   & \xok      \\ \hline
		11 &  \textbf{PrecPro4${}^{0}$}  &   &   & \xok      \\ \hline
		12 & \textbf{PrecPro5${}^{+}$}  &   &   & \xok       \\ \hline
		13 & \textbf{PrecPro6${}^{0}$}  &   &   & \xok      \\ \hline
	\end{tabular}
\end{table}

\pagebreak

\subsection{Detailed Results of the Recall Measure Evaluation }\label{appendixrec}
\subsubsection{Proposition DetPro${}^{+}$}

\paragraph{\textbf{Causal footprint recall ($rec_A$).}}
\textit{Proposition holds.} \\
\textbf{Reasoning.} The causal footprint fully describes the log as well as the model in terms of directly followed relations. By comparing the footprints of the model and the log recall can be determined.

\paragraph{\textbf{Token replay recall ($rec_B$).}}
\textit{Proposition does not hold.} \\
\textbf{Reasoning.} This technique depends on the path taken through the model. Due to duplicate activities and silent transitions, multiple paths through a model can be taken when replaying a single trace. Different paths can lead to different numbers of produced, consumed, missing and remaining tokens and recall is not deterministic.

\paragraph{\textbf{Alignment recall ($rec_C$).}}
\textit{Proposition holds.} \\
\textbf{Reasoning.} When computing alignments the algorithm searches per trace for the optimal alignment: the alignment with the least cost associated to it. There may be multiple alignments, but these all have the same cost. Recall is computed based on this cost. Therefore, given a log, a model and a cost function the recall computation is deterministic.

\paragraph{\textbf{Behavioral recall ($rec_D$).}}
\textit{Proposition does not hold.} \\
\textbf{Reasoning. } If duplicate or silent transitions are encountered during replay of the traces, which were enhanced with negative events, it is explored which of the available transitions enables the next event in the trace. If no solution is found one of the transitions is randomly fired, which can lead to different recall values for traces with the same behavior.

\paragraph{\textbf{Projected recall ($rec_E$).}}
\textit{Proposition holds.} \\
\textbf{Reasoning.} This technique splits log and model into subsets and calculates how many traces in the sub-log can be replayed on the corresponding sub-model, which is represented as a deterministic finite automaton. The sub-logs and models are created by projection on a subset of activities which is a deterministic process. Therefore, the computation of the average recall value over all subsets is also deterministic.

\paragraph{\textbf{Continued parsing measure ($rec_F$).}}
\textit{Proposition does not hold.} \\
\textbf{Reasoning.} The continued parsing measure translates the behavior of the process model into a causal matrix. This translation is not defined if the model contains duplicate or silent transitions. Consequently, the continued parsing measure is not defined for these models, which violates this proposition.

\paragraph{\textbf{Eigenvalue recall ($rec_G$).}}
\textit{Proposition holds.} \\
\textbf{Reasoning.} The measure compares the languages of the model and the language of the process model. These have to be irreducible, to compute their eigenvalue. Since the language of an event log is not irreducible, Polyvyanyy et~al.\@ \cite{polyvyanyy-conf} introduce a short-circuit measure over languages and proof that it is a deterministic measure over any arbitrary regular language.

\subsubsection{Proposition 2 BehPro${}^{+}$}

\paragraph{\textbf{Causal footprint recall ($rec_A$).}}
\textit{Proposition holds.} \\
\textbf{Reasoning.} The causal footprint completely describes the log as well as the model in terms of directly followed relations and therefore does not depend on the representation of the model.

\paragraph{\textbf{Token replay recall ($rec_B$).}}
\textit{Proposition does not hold.} \\
\textbf{Reasoning.} Due to duplicate activities and silent transitions, one can think of models with the same behavior but a different structure. It is also possible to have implicit places that do not change the behavior but do influence the number of produced, consumed, missing and remaining tokens. For example, if a place often has missing tokens, then duplicating this place will lead to even more missing tokens (also relatively). Moreover, nondeterminism during replay can lead to a difference in the replay path and therefore in different numbers of produced, consumed, missing and remaining tokens and shows that token replay-based recall depends on the representation of the model.

\paragraph{\textbf{Alignment recall ($rec_C$).}}
\textit{Proposition holds.} \\
\textbf{Reasoning.} Silent transitions are used for routing behavior of the Petri net and are not linked to the actual behavior of the process. During alignment computation, there is no cost associated with silent transitions. Also, the places do not play a role (e.g., implicit places have no effect). Therefore, the structure of the model itself has no influence on the alignment computation and two different models expressing the same behavior will result in the same recall measures.

\paragraph{\textbf{Behavioral recall ($rec_D$).}}
\textit{Proposition does not hold.} \\
\textbf{Reasoning.} If duplicate or silent transitions are encountered during the replay of the traces, which were enhanced with negative events on the model, it is explored which of the available transitions enables the next event in the trace. If no solution is found, one of the transitions is randomly fired, which can lead to different recall values for a trace on two behaviorally equivalent but structurally different models.

\paragraph{\textbf{Projected recall ($rec_E$).}}
\textit{Proposition holds.} \\
\textbf{Reasoning.} This technique translates the event log as well as the process model into deterministic finite automata before computing recall. Therefore, it is independent of the representation of the model itself.

\paragraph{\textbf{Continued parsing measure ($rec_F$).}}
\textit{Proposition holds.} \\
\textbf{Reasoning.} The continued parsing measure translates the possible behavior into so-called input expressions and output expressions, which describe possible behavior before and after the execution of each activity. Therefore, it abstracts from the structure of the process model.

\paragraph{\textbf{Eigenvalue recall ($rec_G$).}}
\textit{Proposition holds.} \\
\textbf{Reasoning.} The approach computes recall based on the languages of the event log and the language of the process model. This abstracts from the representation of the process model and, therefore, the proposition holds.

\subsubsection{Proposition 3 RecPro1${}^{+}$}

\begin{figure}[t]
{
\centering
\includegraphics[width=\textwidth]{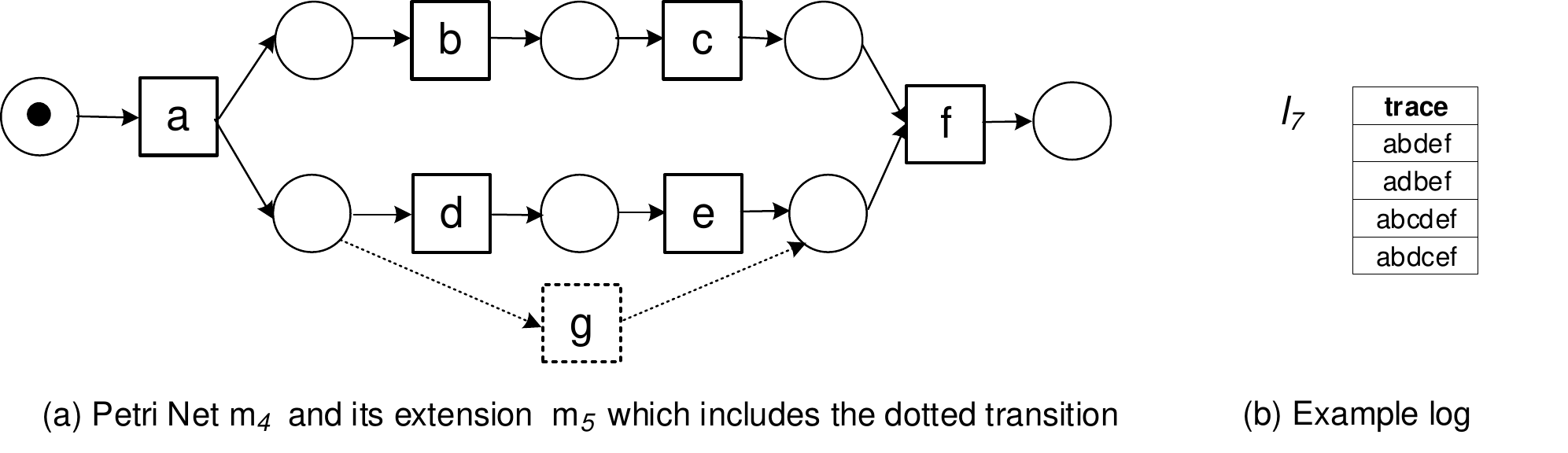}
\caption{Petri net $m_4$ and its extension $m_5$ which includes the dotted transition (b), as well as example log $l_7$.
}\label{ce-mod1}
}
\end{figure}

\paragraph{\textbf{Causal footprint recall ($rec_A$).}}
\textit{Proposition does not hold.} \\
\textbf{Reasoning.} Recall is calculated by dividing the number of relations where log and model differ by the total number of relations. When adding behavior to the model while keeping the log as is, the causal footprint of the model changes while the causal footprint of the log stays the same. This may introduce more differences between both footprints and therefore lowers recall. Process model $m_4$, its extension $m_5$ in Figure~\ref{ce-mod1} (a) and event log $l_7 = [ \la a,d,b,e,f \ra,\la a,b,d,e,f \ra,\la  a,b,c,d,e,f \ra, \la  a,b,d,c,e,f \ra]$ illustrate this. The corresponding causal footprints are displayed in Table~\ref{tabfootprint1}. Since the log $l_7$ does not contain activity $g$, when computing recall between $l_7$ and $m_5$, we assume that all activities show a \#-relation with $g$. Computing recall based on the footprints results in a $rec_A(l_7, m_4) = 1-\frac{6}{36} = 0.83$ and  $rec_A(l_7, m_5) = 1-\frac{14}{49} = 0.71$. The proposition is violated since $rec_A(l_7, m_4) < rec_A(l_7, m_5)$.  Van der Aalst mentions in \cite{process-mining-book-2016} that checking conformance using causal footprints is only meaningful if the log is complete in term of directly followed relations. Furthermore, the approach intended to cover precision, generalization and recall in one conformance value.

\begin{table}[t!]
	\centering \caption{The Causal footprints of $m_4$ (a), $l_7$ (b) and  $m_5$ (c). Mismatching relations are marked in \textcolor{red}{red}.} \label{tabfootprint1}
	\subfloat[][]{\begin{tabular}{ccccccc}
		& \textbf{a} & \textbf{b}      & \textbf{c} & \textbf{d} & \textbf{e} & \textbf{f} \\
		\textbf{a} & \#         & $\rightarrow$ &  \#          & $\rightarrow$           &  \#        &   \#        \\
		\textbf{b} &     $\leftarrow$         & \#              &  $\rightarrow$        &  $||$          &   {\color[HTML]{FE0000}$||$}         &   \#         \\
		\textbf{c} &     \#         &      $\leftarrow$             & \#         &      $||$      &   {\color[HTML]{FE0000}$||$     }    &     {\color[HTML]{FE0000}$\rightarrow$     }       \\
		\textbf{d} &    $\leftarrow$     &     $||$              &      $||$       & \#         &   $\rightarrow$         &     \#       \\
		\textbf{e} &    \#      &     {\color[HTML]{FE0000}$||$ }            &       {\color[HTML]{FE0000}$||$ }        &       $\leftarrow$       & \#         &        $\rightarrow$    \\
		\textbf{f} &  \#          &   \#            &   {\color[HTML]{FE0000}$\leftarrow$      }          &     \#       &    $\leftarrow$          & \#
	\end{tabular}}
\quad
	\subfloat[][]{\begin{tabular}{ccccccc}
		& \textbf{a} & \textbf{b}      & \textbf{c} & \textbf{d} & \textbf{e} & \textbf{f} \\
		\textbf{a} & \#         & $\rightarrow$ &  \#          &     $\rightarrow$      &   \#        &   \#         \\
		\textbf{b} &     $\leftarrow$         & \#              &   $\rightarrow$         &   $||$         &   $\rightarrow$         &   \#         \\
		\textbf{c} &     \#       &      $\leftarrow$             & \#         &       $||$      & $\rightarrow$        &     \#      \\
		\textbf{d} &   $\leftarrow$     &   $||$               &    $||$          & \#         &   $\rightarrow$         &     \#       \\
		\textbf{e} &   \#        & $\leftarrow$              &       $\leftarrow$       &       $\leftarrow$       & \#         &        $\rightarrow$    \\
		\textbf{f} &  \#          &      \#            &     \#         &     \#         &    $\leftarrow$          & \#
	\end{tabular}}
\quad
	\subfloat[][]{\begin{tabular}{cccccccc}
		& \textbf{a} & \textbf{b}      & \textbf{c} & \textbf{d} & \textbf{e} & \textbf{f}& \textbf{g} \\
		\textbf{a} & \#         & $\rightarrow$ &  \#          & $\rightarrow$           &  \#        &   \#   & {\color[HTML]{FE0000}$\rightarrow$     }     \\
		\textbf{b} &     $\leftarrow$         & \#              &  $\rightarrow$        &  $||$          &   {\color[HTML]{FE0000}$||$}         &   \#     & {\color[HTML]{FE0000}$||$     }    \\
		\textbf{c} &     \#         &      $\leftarrow$             & \#         &      $||$      &   {\color[HTML]{FE0000}$||$     }    &     {\color[HTML]{FE0000}$\rightarrow$     }   & {\color[HTML]{FE0000}$||$     }       \\
		\textbf{d} &    $\leftarrow$     &     $||$              &      $||$       & \#         &   $\rightarrow$         &     \#     & \#    \\
		\textbf{e} &    \#      &     {\color[HTML]{FE0000}$||$ }            &       {\color[HTML]{FE0000}$||$ }        &       $\leftarrow$       & \#         &        $\rightarrow$  & \#     \\
		\textbf{f} &  \#          &   \#            &   {\color[HTML]{FE0000}$\leftarrow$      }          &     \#       &    $\leftarrow$          & \#    & {\color[HTML]{FE0000}$\leftarrow$       } \\
		\textbf{g} &  {\color[HTML]{FE0000}$\leftarrow$       }         &   {\color[HTML]{FE0000}$||$     }            &   {\color[HTML]{FE0000}$||$     }         &     \#        &    \#           & {\color[HTML]{FE0000}$\leftarrow$       }    &\#
	\end{tabular}}
\end{table}

\paragraph{\textbf{Token replay recall ($rec_B$).}}
\textit{Proposition does not hold.} \\
\textbf{Reasoning.} \textbf{BehPro$^{+}$} does not hold, which implies that \textbf{RecPro1$^{+}$} does not hold.

\paragraph{\textbf{Alignment recall ($rec_C$).}}
\textit{Proposition holds.} \\
\textbf{Reasoning.} Note that the model extension only adds behavior to the model and does not restrict it further. During alignment computation, this means that either the initial alignments are computed or that the additional behavior resulted in an optimal alignment with even lower cost (i.e. alignments with less log/model moves). Therefore, recall of the extended model cannot be lower than the value calculated for the initial model.

\paragraph{\textbf{Behavioral recall ($rec_D$).}}
\textit{Proposition does not hold.} \\
\textbf{Reasoning.} \textbf{BehPro$^{+}$} does not hold, which implies that \textbf{RecPro1$^{+}$} does not hold.

\paragraph{\textbf{Projected recall ($rec_E$).}}
\textit{Proposition holds.} \\
\textbf{Reasoning.} Based on the definition of projected recall it can only be lowered if fewer traces of the log can be replayed on the model. This is only possible if the model extension also restricts parts of its behavior. Hence, by purely extending the model the number of fitting traces can only be increased: $\card{\left[ t \in l|_A | t \in DFA(m_1|_A)\right]} \leq \linebreak \card{\left[ t \in l|_A | t \in DFA(m_2|_A)\right]}$ if $\traces{m_1} \subseteq \traces{m_2}$. As a result, recall cannot be lowered and the proposition holds.

\paragraph{\textbf{Continued parsing measure ($rec_F$).}}
\textit{Proposition holds.} \\
\textbf{Reasoning.} Note that the model extension only adds behavior to the model and does not restrict it further. When replaying the log on the extended causal matrix the number of missing and remaining activated expressions stays the same or decreases which consequently cannot lower recall.

\paragraph{\textbf{Eigenvalue recall ($rec_G$).}}
\textit{Proposition holds.} \\
\textbf{Reasoning.} Trivially, adding behavior to the model can only increase the intersection between the language of the log and the model.: $\mathcal{L}(l)\cap \mathcal{L}(m_1) \leq \mathcal{L}(l)\cap \mathcal{L}(m_2) $ if $\traces{m_1} \subseteq \traces{m_2}$. Polyvyanyy et~al.\@ \cite{polyvyanyy-conf} proved in lemma 5.6 that the short-circuit measure based on eigenvalue is increasing, i.e. that $\frac{eig(\mathcal{L}(l)\cap \mathcal{L}(m_1))}{eig(\mathcal{L}(l))} \leq \frac{eig(\mathcal{L}(l)\cap \mathcal{L}(m_2))}{eig(\mathcal{L}(l))}$.

\subsubsection{Proposition 4 RecPro2${}^{+}$}
\paragraph{\textbf{Causal footprint recall ($rec_A$).}}
\textit{Proposition holds.} \\
\textbf{Reasoning.} Adding fitting behavior to the event log either does not change the footprint of the log because no new relations were observed or it changes the corresponding causal footprint in a way that more of its relations match the causal footprint of the model. The only three options are, that a \#-relation changes to $\rightarrow$, $\rightarrow$ becomes $||$ or $\leftarrow$ becomes $||$. Hence the differences between the two footprints are minimized and recall improved.

\paragraph{\textbf{Token replay recall ($rec_B$).}}
\textit{Proposition holds.} \\
\textbf{Reasoning.} Adding fitting behavior to the event log means that these traces can be replayed on the model without any problems. Here we assume that if a trace is perfectly replayable it will also be replayed perfectly. In case of duplicate and silent transitions this does not need to be the case. Consider two very long branches in the process model allowing for the same behavior. Only at the end, they have differences. This may lead to the situation where initially the wrong branch was chosen. In this paper, we make the assumption that fitting behavior is replayed correctly. Hence, adding fitting traces results in more produced and consumed tokens without additional missing or remaining tokens ($c_1 < c_2$ and $p_1 < p_2$).  Therefore, recall can only be improved by adding fitting behavior. $\frac{1}{2}(1-\frac{m}{c_1}) + \frac{1}{2}(1-\frac{r}{p_1}) \leq \frac{1}{2}(1-\frac{m}{c_2}) + \frac{1}{2}(1-\frac{r}{p_2})$.

\paragraph{\textbf{Alignment recall ($rec_C$).}}
\textit{Proposition holds.} \\
\textbf{Reasoning.} Fitting behavior results in a perfect alignment which only consists of synchronous moves. Consequently, this alignment has no costs assigned and adding it to the existing log cannot lower recall.

\paragraph{\textbf{Behavioral recall ($rec_D$).}}
\textit{Proposition holds.} \\
\textbf{Reasoning.} For fitting log $l_3$, $FN(l_3,m)=0$ and $TP(l_3,m)$ is proportional to the size of $l_3$. For $l_2=l_1\uplus l_3$, we have $TP(l_2,m)=TP(l_1,m)+TP(l_3,m)$, and $FN(l_2,m)=FN(l_1,m)+FN(l_3,m)=FN(l_1,m)$. Therefore, $rec_D(l_2,m)=\frac{TP(l_1,m)+TP(l_3,m)}{FN(l_1,m)}$ and since $rec_D(l_1,m)=\frac{TP(l_1,m)}{FN(l_1,m)}$, we have $rec_D(l_2,m)\ge rec_D(l_1,m)$.

\paragraph{\textbf{Projected recall ($rec_E$).}}
\textit{Proposition holds.} \\
\textbf{Reasoning.} Based on the definition of projected recall it can only be lowered if fewer traces of the log can be replayed on the model. Fitting traces can be replayed and recall cannot be lowered by adding them to the log. $\card{\left[ t \in l_1|_A | t \in DFA(m|_A)\right]} \leq \linebreak \card{\left[ t \in l_2|_A | t \in DFA(m|_A)\right]}$ if $l_2 = l_1 \uplus l_3$ and $\traces{l_3} \subseteq \traces{m}$. Averaging recall over all subsets of a given length also does not influence this.

\paragraph{\textbf{Continued parsing measure ($rec_F$).}}
\textit{Proposition holds.} \\
\textbf{Reasoning.} Adding fitting behavior to the event log means that these traces can be replayed on the causal matrix without any problems. Here we assume, similar to $rec_B$, that if a trace is perfectly replayable it will also be replayed perfectly. Hence fitting behavior does not yield additional missing or remaining activated expressions. Therefore recall can only be improved.

\paragraph{\textbf{Eigenvalue recall ($rec_G$).}}
\textit{Proposition holds.} \\
\textbf{Reasoning.} Trivially, adding fitting behavior to the event log can only increase the intersection between the language of the log and the model, i.e., $\mathcal{L}(l_1)\cap \mathcal{L}(m) \leq \mathcal{L}(l_2)\cap \mathcal{L}(m) $ if $l_2 = l_1 \uplus l_3$ and $\traces{l_3} \subseteq \traces{m}$. Polyvyanyy et~al.\@ \cite{polyvyanyy-conf} proved in lemma 5.6 that the short-circuit measure based on eigenvalue is increasing, and therefore it also holds that $\frac{eig(\mathcal{L}(l_1)\cap \mathcal{L}(m))}{eig(\mathcal{L}(l_1))} \leq \frac{eig(\mathcal{L}(l_2)\cap \mathcal{L}(m))}{eig(\mathcal{L}(l_2)) }$.

\subsubsection{Proposition 5 RecPro3${}^{0}$}

\paragraph{\textbf{Causal footprint recall ($rec_A$).}}
\textit{Proposition does not hold.} \\
\textbf{Reasoning.} The causal footprint technique defines fitting and non-fitting behavior on an event level, while the proposition states that a trace is either fitting or non-fitting. The added non-fitting behavior could consist of multiple fitting and one non-fitting event. The fitting events could actually decrease the differences between the log and the model while the single non-fitting event introduces an additional difference. The added non-fitting trace in total introduce more similarities than differences and therefore improve recall. Beyond that, it is possible that the other traces in the log already resulted in the appropriate causal relations and the non-fitting event does not change anything.

To illustrate that, consider $m_4$ and $l_7$ in Figure~\ref{ce-mod1}. We extend $l4$ with four clearly unfitting traces: $l_8 = l_7 \uplus [ \la a,d,b,f \ra,\la  a,e,b,f \ra,\la  a,e,c \ra, \la  a,b,c,e \ra]$. The footprint of $l_8$ is displayed in Table~\ref{tabfootprint2}. It shows that adding the unfitting traces reveals the parallelism between the activities $d, e$ and $b$, and decreases the differences to the footprint of $m_4$ in Figure~\ref{ce-mod1} (a). Consequently, $rec_A(l_8,m_4) = 1 - \frac{4}{36} = 0.88$ and $rec_A(l_7,m_4) < rec_A(l_8,m_4)$.

\begin{table}[t!]
	\centering \caption{The causal footprints of $l_8$. The differences to the footprint of $m_4$ in Figure~\ref{ce-mod1} are marked in {\color[HTML]{FE0000}red}} \label{tabfootprint2}
	{\begin{tabular}{ccccccc}
		& \textbf{a} & \textbf{b}      & \textbf{c} & \textbf{d} & \textbf{e} & \textbf{f} \\
		\textbf{a} & \#         & $\rightarrow$ &  \#          &     $\rightarrow$      &   {\color[HTML]{FE0000}$\rightarrow$}         &   \#         \\
		\textbf{b} &     $\leftarrow$         & \#              &   $\rightarrow$         &   $||$         &   $||$          &   {\color[HTML]{FE0000}$\rightarrow$}          \\
		\textbf{c} &     \#       &      $\leftarrow$             & \#         &       $||$      & $||$      &     \#      \\
		\textbf{d} &   $\leftarrow$     &   $||$               &    $||$          & \#         &   $\rightarrow$         &     \#       \\
		\textbf{e} &  {\color[HTML]{FE0000}$\leftarrow$}       & $||$             &       $||$       &       $\leftarrow$       & \#         &        $\rightarrow$    \\
		\textbf{f} &  \#          &     {\color[HTML]{FE0000}$\leftarrow$}              &     \#         &     \#         &    $\leftarrow$          & \#
	\end{tabular}}
\end{table}

\paragraph{\textbf{Token replay recall ($rec_B$).}}
\textit{Proposition does not hold.} \\
\textbf{Reasoning.} In the recall formula used during token replay, the number of produced and consumed tokens is in the denominator, while the number of missing and remaining token is in the nominator. If we add a very long non-fitting trace to the event log, which yields a lot of produced and consumed token but only a few missing and remaining token, recall is improved. For long cases with only a few deviations, the approach gives too much weight to the fitting part of the trace.

\begin{figure}[t]
	{
		\centering
		\includegraphics[width=0.8\textwidth]{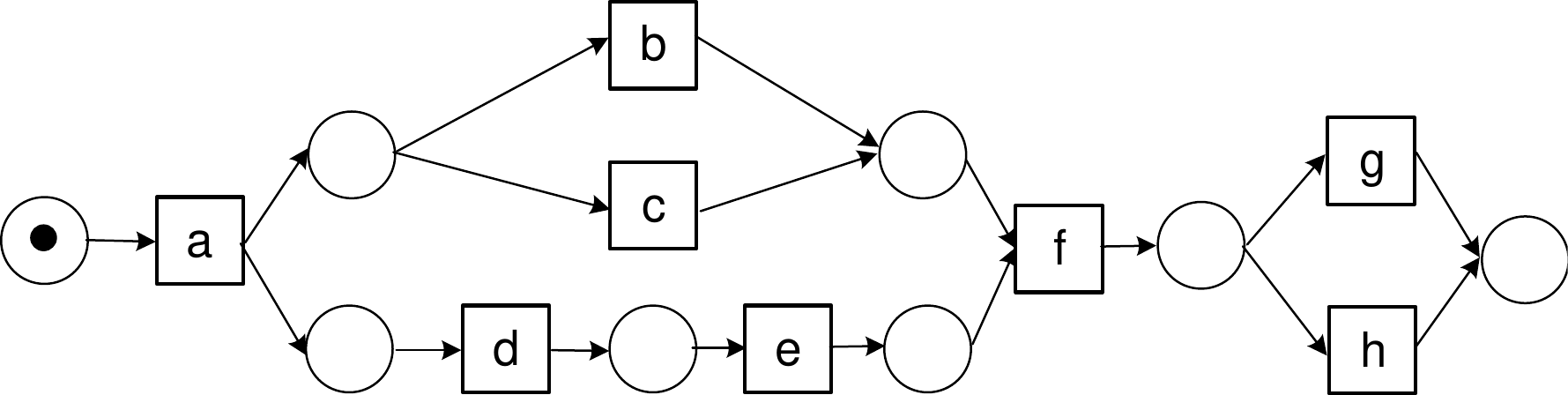}
		\caption{Petri net $m_6$
		}\label{ce-mod2}
	}
\end{figure}

To illustrate this consider process model $m_6$ of Figure~\ref{ce-mod2} and log $l_9 =  [ \la a,b,f,g]$. This log is not perfectly fitting and therefore results in 6 produced and 6 consumed tokens, as well as 1 missing and 1 remaining token. $Rec_B(l_9, m_6) = \frac{1}{2}(1-\frac{1}{6}) + \frac{1}{2}(1-\frac{1}{6}) = 0.833$. We extend the log $l_9$ with non-fitting behavior: $l_{10} = l_9 \uplus [ \la  a,d,e,f,g \ra]$. Replaying $l_{10}$ on $m_6$ results in $p = c = 13$, $r = m = 2$ and $rec_B(l_{10}, m_6) = \frac{1}{2}(1-\frac{2}{13}) + \frac{1}{2}(1-\frac{2}{13}) = 0.846$. Hence, $rec_B(l_9, m_6) < rec_B(l_{10}, m_6)$.

\paragraph{\textbf{Alignment recall ($rec_C$).}}
\textit{Proposition does not hold.} \\
\textbf{Reasoning.} Consider model $m_4$ in Figure~\ref{ce-mod1} and event log $l_{11} = [ \la f,a \ra]$, which result in costs $fcost(L,M) = 6$ and worst-case cost $move_L(L) = 2$ and $move_M(M) = 6$. Consequently, recall is $rec_C(l_{11},m_4)= 1-\frac{6}{2+6} = 0.25$. We add a non-fitting trace to the log $l_{12} = l_{11} \uplus \la a,b,c,d,e \ra$, which shows less deviations than $l_{11}$. This results in an additional cost of 1 and additional worst-case costs of $5+6$. This leads to $rec_C(l_{12},m_4)=1- \frac{7}{(2+5)+ 2 \times 6} = 0.64$. Hence, adding a non-fitting trace with fewer deviations improves recall $rec_C(l_{11},m_4) < rec_C(l_{12},m_4)$ and  violates the proposition.

\paragraph{\textbf{Behavioral recall ($rec_D$).}}
\textit{Proposition does not hold.} \\
\textbf{Reasoning.} In the recall formula used during token replay, the number correctly replayed events (TP) is in the denominator as well as in the nominator, while the number of transitions that were forced to fire although they were not enabled (FN) is only in the denominator. If we now add a very long non-fitting trace to the event log, which consists of a large number of correctly replayed events but only a few force firings, recall is improved. Furthermore, remaining tokens during replay are not considered in the formula and cannot lower recall, although these tokens are clear indications for unfitting traces.

Consider $m_6$ of Figure~\ref{ce-mod2} and log $l_9 =  [ \la a,b,f,g]$. After replaying the log on the model there are 3 recorded true positive events and 1 recorded false negative events. This results in $rec_D(l_9, m_6) = \frac{3}{3 + 1} = 0.75$. We extend the log $l_9$ with non-fitting behavior: $l_{10} = l_9 \uplus [ \la  a,d,e,f,g \ra]$. Replaying $l_{10}$ on $m_6$ results in 7 recorded true positive events and 2 recorded false negative events. Consequently $rec_D(l_{10}, m_6) = \frac{7}{7 + 2} = 0.77$ and $rec_D(l_9, m_6) < rec_D(l_{10}, m_6)$.

\paragraph{\textbf{Projected recall ($rec_E$).}}
\textit{Proposition does not hold.} \\
\textbf{Reasoning.} According to this approach, there can be traces that are fitting most of the automata and traces that are fitting only a few automata. The counter-example of $rec_C$ illustrates this and shows that the proposition does not hold. Projecting the model and both logs on $\{\la a,b \ra\}$ results in recall values for both logs of $ rec_E(l_9|_{\la a, b \ra}, m_6|_{\la a, b \ra})=\frac{0}{1}$ and $ rec_E(l_{10}|_{\la a, b \ra}, m_6|_{\la a, b \ra})=\frac{1}{2}$. The additional non-fitting trace in $l_{10}$ similarly fits the other projected DFAs of size 2, except for the ones containing $f$. However, event log $l_9$ fits none of the projected DFAs of size 2.
Therefore adding non-fitting traces that fit most projected automata can increase the aggregated recall.

\paragraph{\textbf{Continued parsing measure ($rec_F$).}}
\textit{Proposition does not hold.} \\
\textbf{Reasoning.} In the recall formula, the number of events $e$ in the log  is present in the denominator as well as in the nominator while the number of missing $m$ and remaining activated expressions $r$ is subtracted from the nominator. Similar to token replay recall (B), adding a very long non-fitting trace to the event log which introduces a large number of events but only a few missing and remaining activated expressions, improves recall.


\paragraph{\textbf{Eigenvalue recall ($rec_G$).}}
\textit{Proposition holds.} \\
\textbf{Reasoning.}
Trivially, unfitting behavior to the event log cannot change  the intersection between the language of the log and the model and consequently also not their eigenvalue. However, the language of the log might increase which lowers recall. $\mathcal{L}(l_1)\cap \mathcal{L}(m) = \mathcal{L}(l_2)\cap \mathcal{L}(m) $ if $l_2 = l_1 \uplus l_3$ and $\traces{l_3} \subseteq \compltr{m}$. Polyvyanyy et~al.\@ \cite{polyvyanyy-conf} proved in lemma 5.6 that the short-circuit measure based on eigenvalue is increasing, and therefore it also holds that $\frac{eig(\mathcal{L}(l_1)\cap \mathcal{L}(m))}{eig(\mathcal{L}(l_1))} \geq \frac{eig(\mathcal{L}(l_2)\cap \mathcal{L}(m))}{eig(\mathcal{L}(l_2))} $.

\subsubsection{Proposition 6 RecPro4${}^{0}$}
\paragraph{\textbf{Causal footprint recall ($rec_A$).}}
\textit{Proposition holds.} \\
\textbf{Reasoning.} The causal footprint does not account for frequencies of traces and events. Therefore, multiplying the log has no influence on the causal footprint and therefore recall does not change.

\paragraph{\textbf{Token replay recall ($rec_B$).}}
\textit{Proposition holds.} \\
\textbf{Reasoning.} Multiplying the log $k$ times will equally increase the number of produced, consumed, missing and remaining token. Their ratios stay the same and recall does not change. $\frac{1}{2}(1 - \frac{k \cdot m}{k \cdot c}) + \frac{1}{2}(1 - \frac{k \cdot r}{k \cdot p}) = \frac{1}{2}(1 - \frac{m}{c}) + \frac{1}{2}(1 - \frac{r}{p})$, $rec_B(l^k,m) = rec_B(l,m)$.

\paragraph{\textbf{Alignment recall ($rec_C$).}}
\textit{Proposition holds.} \\
\textbf{Reasoning.} Multiplying the event log $k$ times has no influence on recall since the formula accounts for trace frequency in denominator and nominator. The ratio of replay cost and cost of the worst case scenario stays the same and recall does not change. \linebreak $1 - \frac{k \times fcost(L,M) }{k \times move_L(L)+ k \times \card{L}\times move_M(M)} =  1 - \frac{fcost(L,M)}{move_L(L)+ \card{L}\times move_M(M)}$, $rec_C(l^k,m) = rec_C(l,m)$.

\paragraph{\textbf{Behavioral recall ($rec_D$).}}
\textit{Proposition holds.} \\
\textbf{Reasoning.} Multiplying the log $k$ times will equally increase the number of correctly replayed events and force fired transitions. \\ For any $l\in\mathcal{L}$, $rec_D(l^k,m)=\frac{k \times TP(l,m)}{k \times TP(l,m) + k \times FN(l,m)}=\frac{k \times TP(l,m)}{k \times (TP(l,m) + FN(l,m))}=\frac{TP(l,m)}{(TP(l,m) + FN(l,m))}=rec_D(l,m)$.

\paragraph{\textbf{Projected recall ($rec_E$).}}
\textit{Proposition holds.} \\
\textbf{Reasoning.} Multiplying the log will equally increase the total number of traces and fitting traces. Their ratio stays the same and recall does not change. \linebreak $ \frac{k \cdot \card{\left[ t \in l|_A | t \in DFA(m|_A)\right] } }{(k \cdot \card{l|_A})}  = \frac{\card{\left[ t \in l|_A | t \in DFA(m|_A)\right] }}{\card{l|_A}}$, $rec_E(l^k,m) = rec_E(l,m)$. Averaging over all subsets of a given length also does not influence this.

\paragraph{\textbf{Continued parsing measure ($rec_F$).}}
\textit{Proposition holds.} \\
\textbf{Reasoning.} Multiplying the log will equally increase the number of parsed events, missing and remaining activated expressions. Their ratio stays the same and recall does not change. $\frac{1}{2} \frac{k \cdot (e-m)}{k \cdot e} + \frac{1}{2} \frac{k \cdot (e-r)}{k \cdot e} = \frac{1}{2} \frac{(e-m)}{e} + \frac{1}{2} \frac{(e-r)}{e}$, $rec_F(l^k,m) = rec_F(l,m)$.

\paragraph{\textbf{Eigenvalue recall (G.)}}
\textit{Proposition holds.} \\
\textbf{Reasoning.} Eigenvalue recall is defined purely on the language of the log and the model and it does not take into account trace frequencies in the log, therefore, this proposition holds.

\subsubsection{Proposition 7 RecPro5${}^{+}$}

\paragraph{\textbf{Causal footprint recall ($rec_A$).}}
\textit{Proposition does not hold.} \\
\textbf{Reasoning.} The recall measure based on causal footprints compares the behavior in both directions. If the model has additional behavior that is not present in the log, even in the case where all traces in the log fit the model, the footprint comparison will show the difference and recall will not be maximal.

To illustrate this, consider $m_4$ in Figure~\ref{ce-mod1}. We compute recall for $m_4$ and $l_{13} = [ \la  a,b,c,d,e,f \ra, \la  a,b,d,c,e,f \ra ] $. The traces in $l_{13}$ perfectly fit process model $m_4$. The footprint of $l_{13}$ is shown in Table~\ref{tabfootprint3}. Comparing it to the footprint of $m_4$ in Table~\ref{tabfootprint1} (a) shows mismatches although $l_{13}$ is perfectly fitting. These mismatches are caused by the fact that the log does not show all possible paths of the model and therefore the footprint cannot completely detect the parallelism of the model. Consequently, $rec_A(l_{13},m_4)=1-\frac{10}{36}= 0.72 \neq 1$ even though $\traces{l} \subseteq \traces{m}$.

Van der Aalst mentions in \cite{process-mining-book-2016} that checking conformance using causal footprints is only meaningful if the log is complete in term of directly followed relations.

\begin{table}[t!]
	\centering \caption{The causal footprints of $l_{13}$. Mismatches with the footprint of $m_4$ are marked in red.} \label{tabfootprint3}
	\begin{tabular}{ccccccc}
		& \textbf{a} & \textbf{b}      & \textbf{c} & \textbf{d} & \textbf{e} & \textbf{f} \\
		\textbf{a} & \#         & $\rightarrow$ &  \#          & {\color[HTML]{FE0000}\# }          &   \#          &   \#          \\
		\textbf{b} &     $\leftarrow$         & \#              &   $\rightarrow$         &   {\color[HTML]{FE0000}$\rightarrow$}         &   {\color[HTML]{FE0000}\#  }         &   \#          \\
		\textbf{c} &     \#          &      $\leftarrow$             & \#         &        $||$      & {\color[HTML]{FE0000}$\rightarrow$}         &      {\color[HTML]{FE0000}\#  }        \\
		\textbf{d} &   {\color[HTML]{FE0000}\#  }         &   {\color[HTML]{FE0000}$\leftarrow$}               &    $||$          & \#         &   $\rightarrow$         &     \#       \\
		\textbf{e} &   \#        & {\color[HTML]{FE0000}\#}              &       {\color[HTML]{FE0000}$\leftarrow$}        &       $\leftarrow$       & \#         &        $\rightarrow$    \\
		\textbf{f} &  \#         &      \#            &    {\color[HTML]{FE0000}\#}        &    \#         &    $\leftarrow$          & \#
	\end{tabular}
\end{table}

\paragraph{\textbf{Token replay recall ($rec_B$).}}
\textit{Proposition holds.} \\
\textbf{Reasoning.} There will be no missing and remaining tokens if all traces in the log fit the model. Hence, recall is maximal, if $\traces{l} \subseteq \traces{m}$. $\frac{1}{2}(1-\frac{0}{p}) + \frac{1}{2}(1-\frac{0}{c} = 1 $. Note, that again we make the assumption that perfectly fitting behavior is replayed perfectly. Due to the nondeterministic nature of replay in the presence of silent and duplicate transition, this is not guaranteed.

\paragraph{\textbf{Alignment recall ($rec_C$).}}
\textit{Proposition holds.} \\
\textbf{Reasoning.} The alignments only consist of synchronous moves if all traces in the log fit the model. Consequently, the alignment costs $fcost(L,M)$  are 0 and recall is maximal. $rec_C = 1 - \frac{fcost(L,M)}{(move_L(L)+\card{L}\times move_M(M))} = 1 - \frac{0}{(move_L(L)+\card{L}\times move_M(M))}= 1$, if $\traces{l} \subseteq \traces{m}$.

\paragraph{\textbf{Behavioral recall ($rec_D$).}}
\textit{Proposition holds.} \\
\textbf{Reasoning.} If all traces in log $l$ fit model $m$, then $FN(l,m)=0$. As a result, $rec_D(l,m)=\frac{TP(l,m)}{TP(l,m)+FN(l,m)}=\frac{TP(l,m)}{TP(l,m)}=1$.

\paragraph{\textbf{Projected recall ($rec_E$).}}
\textit{Proposition holds.} \\
\textbf{Reasoning.} If all traces in the log fit the model, the number of correctly replayed traces equals the number of traces in the log  $\card{\left[ t \in l|_A | t \in DFA(m|_A)\right] } = \card{l|_A}$, if $\traces{l} \subseteq \traces{m}$ and recall is maximal. The approach also defines that recall is maximal if the log is empty. \cite{sander-scalable-procmin-SOSYM}.

\paragraph{\textbf{Continued parsing measure ($rec_F$).}}
\textit{Proposition does not hold.} \\
\textbf{Reasoning.} Flower models consisting of one place that connects to all transitions do not have a final place. Translating this model into a causal matrix will cause that there is no activity with an empty output expression. Hence, after replaying the fitting log, there will always be remaining activated output expressions and recall is not maximal.

\paragraph{\textbf{Eigenvalue recall (G.)}}
\textit{Proposition holds.} \\
\textbf{Reasoning.}
Proven in Corollary 5.15 of~\cite{polyvyanyy-conf}.

\subsection{Detailed Results of the Precision Measure Evaluation}\label{appendixprec}
\subsubsection{Proposition 1 DetPro${}^{+}$}
\paragraph{\textbf{Soundness ($prec_H$).}}
\textit{Proposition does not hold.} \\
\textbf{Reasoning.} The formula divides the unique traces observed in the log by the unique paths through the model. If the model contains loops there are infinitely many unique paths and precision is not defined.

\paragraph{\textbf{Simple behavioral appropriateness ($prec_I$).}}
\textit{Proposition does not hold.} \\
\textbf{Reasoning.} Shown to be non-deterministic in~\cite{Niek-IPL2018-imprecision} and was already stressed in the original paper~\cite{anne_confcheck_is} that introduced the measure.

\paragraph{\textbf{Advanced behavioral appropriateness ($prec_J$).}}
\textit{Proposition does not hold.} \\
\textbf{Reasoning.} Shown to be undefined for some combinations of logs and models in~\cite{Niek-IPL2018-imprecision}. Note, that implementation of the approach in the process mining tool ProM\footnote{http://www.promtools.org} defines precision for these combinations and is, therefore, deterministic. However, in this paper, we only consider the approach as formally defined in the paper~\cite{anne_confcheck_is}.

\paragraph{\textbf{ETC-one/ETC-rep ($prec_K$).}}
\textit{Proposition does not hold.} \\
\textbf{Reasoning.} For the construction of the state space and its escaping edges, the aligned log is used. In the case of multiple optimal alignments, one (ETC-one) or a set of representative alignments (ETC-rep) is used to construct the state space. During regular conformance checking based on alignments, all optimal alignments are equal. However different alignments can lead to different escaping edges and therefore to different precision measures.

\begin{figure}[t]
	{
		\centering
		\includegraphics[width=0.75\textwidth]{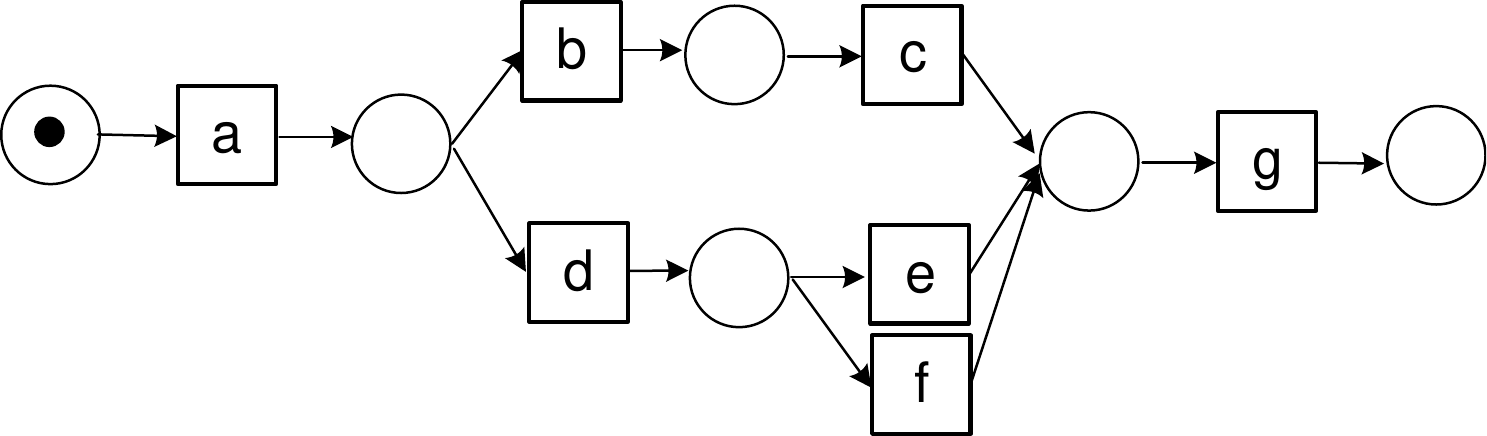}
		\caption{Petri net $m_7$
		}\label{ce-mod3}
	}
\end{figure}

\begin{figure}[t]
	{
		\centering
		\subfloat[][]{\includegraphics[width=0.4\textwidth]{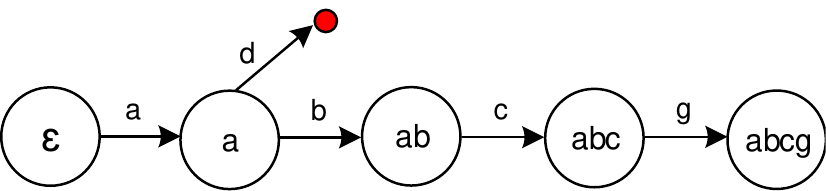}}
		\quad
		\subfloat[][]{\includegraphics[width=0.4\textwidth]{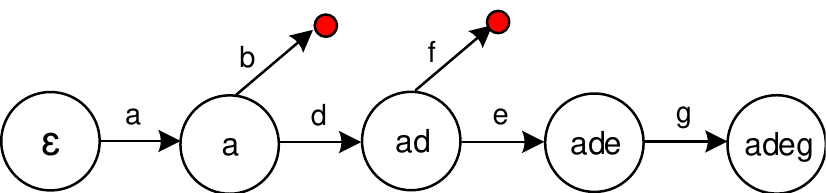}}
		\subfloat[][]{\includegraphics[width=0.4\textwidth]{./figures/autom-1}}
		\quad
		\subfloat[][]{\includegraphics[width=0.4\textwidth]{./figures/autom-2}}
		\caption{Two alignment automata describing the state space of $\sigma_1 = \la a,b,c,g \ra$ (a) and $\sigma_2 = \la a,d,e,g \ra$ (b).
		}\label{autom-1}
	}
\end{figure}

Consider process $m_7$ in Figure~\ref{ce-mod3} along with event log $l_{14} = [\la  a,g \ra]$. It is clear that the log does not fit the process model and after aligning log and model there are three possible aligned traces: $\sigma_1 = \la a,b,c,g \ra$, $\sigma_2 = \la a,d,e,g \ra$ and $\sigma_3 = \la a,d,f,g \ra$. The ETC-one approach randomly picks one of the traces and construct the corresponding alignment automaton. The automata of $\sigma_1$ and $\sigma_2$ in Figure~\ref{autom-1} show the different escaping edges that result from both traces. As a result, precision is different for these two aligned traces: $prec_K(\sigma_1,m_7)= \frac{4}{5} = 0.8$ and $prec_K(\sigma_2,m_7)= \frac{4}{6} = 0.67$.

\paragraph{\textbf{ETC-all ($prec_L$).}}
\textit{Proposition holds.} \\
\textbf{Reasoning.} For ETC-all all optimal alignments are used. Which leads to a complete state space and a deterministic precision measure.

\paragraph{\textbf{Behavioral specificity ($prec_M$).}}
\textit{Proposition does not hold.} \\
\textbf{Reasoning.} If during the replay of the trace duplicate or silent transitions are encountered, the approach explored which of the available transitions enables the next event in the trace. If no solution is found, one of the transitions is randomly fired, which can lead to different recall values for traces with the same behavior.

Furthermore, to balance the proportion of negative and positive events in the log, the algorithm induces the log with the calculated negative events based on a probabilistic parameter. Only if this parameter is set to 1 all negative events are added to the log. Hence the recall measure is non-deterministic for parameter settings smaller than 1.

Finally, it is possible that the negative event induction algorithm does not induce any negative events. This, for example, happens when the algorithm assesses that all activity types in the log are in parallel. When there are no negative events found, it follows from the definition that precision is $\frac{0}{0}$ and thus undefined.
\\
\paragraph{\textbf{Behavioral precision ($prec_N$).}}
\textit{Proposition does not hold.} \\
\textbf{Reasoning.} $prec_N$ uses the same non-deterministic replay procedure and the same negative event induction approach (possibly also non-deterministic, depending on parameter settings) as $prec_M$.

$prec_N$ does not have the same problem as $prec_M$ with regards to being undefined when there are no negative events, as this measure additionally has the number of true positives in the formula.

\paragraph{\textbf{Weighted negative event precision ($prec_O$).}}
\textit{Proposition does not hold.} \\
\textbf{Reasoning.} $prec_O$ uses a non-deterministic replay procedure, which is detailed in~\cite{Broucke2013}. Therefore, the precision calculation is non-deterministic.

\paragraph{\textbf{Projected precision ($prec_P$).}}
\textit{Proposition holds.} \\
\textbf{Reasoning.} This technique projects the behaviors of the log and the model onto subsets of activities and compares their deterministic finite automata to calculate precision. The sub-logs and models are created by projection on a subset of activities which is a deterministic process. Moreover, the process of creating a deterministic automaton is also deterministic. There is a unique DFA which has the minimum number of states, called the minimal automaton. Therefore, the computation of the average precision value over all subsets is also deterministic.

\paragraph{\textbf{Anti-alignment precision ($prec_Q$).}}
\textit{Proposition holds.} \\
\textbf{Reasoning.} Precision is computed based on the maximal anti-alignment. Even if there are multiple maximal anti-alignments, the distance will always be maximal and, therefore, precision is deterministic. Note, that we assume in case of non-fitting behavior that the log is first aligned before evaluating this proposition.

\paragraph{\textbf{Eigenvalue precision ($prec_R$).}}
\textit{Proposition holds.} \\
\textbf{Reasoning.}  The measure compares the languages of the model and the language of the process model. These have to be irreducible, to compute their eigenvalue. Since the language of an event log is not irreducible, Polyvyanyy et~al.\@ \cite{polyvyanyy-conf} introduce a short-circuit measure over languages and proof that it is a deterministic measure over any arbitrary regular language.

\subsubsection{Proposition 2 BehPro${}^{+}$}
\paragraph{\textbf{Soundness ($prec_H$).}}
\textit{Proposition holds.} \\
\textbf{Reasoning.} The behavior of the model is defined as sets of traces $\traces{m}$, which abstracts from the representation of the process model itself.

\paragraph{\textbf{Simple behavioral appropriateness ($prec_I$).}}
\textit{Proposition does not hold.} \\
\textbf{Reasoning.} A counter-example to Axiom 4 (as introduced in~\cite{Niek-IPL2018-imprecision}), which is equivalent to\textbf{ BehPro$^+$}, was shown in~\cite{Niek-IPL2018-imprecision}.

\paragraph{\textbf{Advanced behavioral appropriateness ($prec_J$).}}
\textit{Proposition does not hold.} \\
\textbf{Reasoning.} A counter-example to Axiom 4 (as introduced in~\cite{Niek-IPL2018-imprecision}), which is equivalent to \textbf{BehPro$^+$}, was shown to hold in~\cite{Niek-IPL2018-imprecision}.

\paragraph{\textbf{ETC-one/ETC-rep ($prec_K$).}}
\textit{Proposition does not hold.} \\
\textbf{Reasoning.} A counter-example to Axiom 4 (as introduced in~\cite{Niek-IPL2018-imprecision}), which is equivalent to B\textbf{ehPro$^+$}, was shown in~\cite{Niek-IPL2018-imprecision}.

\begin{figure}[t]
	{
		\centering
		\subfloat[][]{\includegraphics[width=0.4\textwidth]{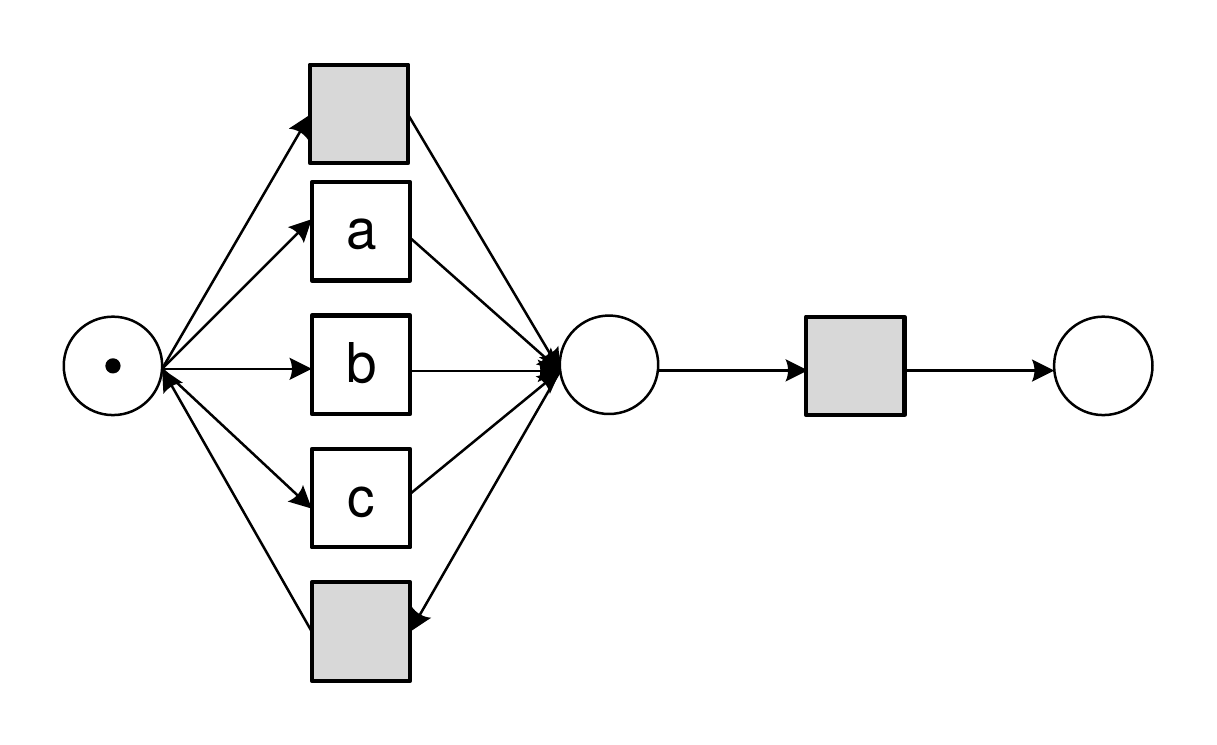}}
		\quad
		\subfloat[][]{\includegraphics[width=0.55\textwidth]{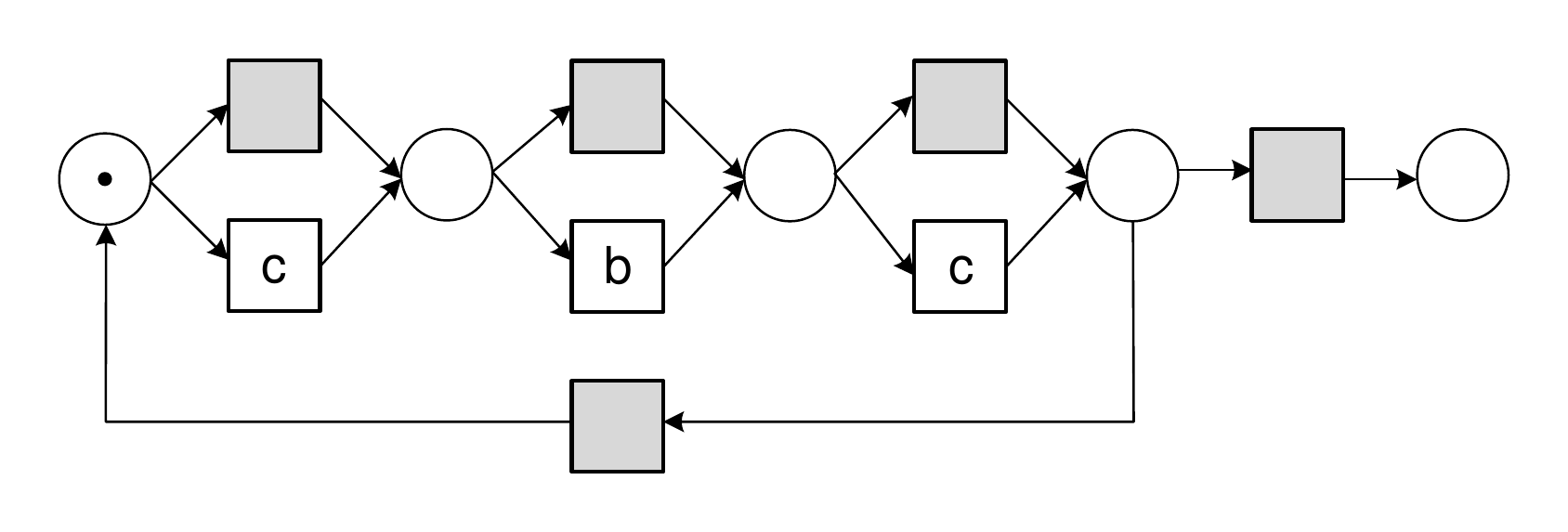}}
		\caption{Two process models $m_8$ (a) and $m_9$ (b) that show the same behavior but different representations.
		}\label{ce-mod4}
	}
\end{figure}

\paragraph{\textbf{ETC-all ($prec_L$).}}
\textit{Proposition does not hold.} \\
\textbf{Reasoning.} This technique depends on the path taken through the model to examine the visited states and its escaping edges. One can think of Petri nets with the same behavior but described by different paths through the net which then also results in different escaping edges and hence different precision. The two models shown in Figure~\ref{ce-mod4} prove this case.

\paragraph{\textbf{Behavioral specificity ($prec_M$).}}
\textit{Proposition does not hold.} \\
\textbf{Reasoning.}  If duplicate or silent transitions are encountered while replaying a trace, the approach checks if one of the available transitions enables the next event in the trace. Whether this is the case can depend on the structure of the model.

\paragraph{\textbf{Behavioral precision ($prec_N$).}}
\textit{Proposition does not hold.} \\
\textbf{Reasoning.} For $prec_M$, \textbf{BehPro$^{+}$ }did not hold because of its replay procedure. $prec_N$ uses the same replay procedure as $prec_M$.

\paragraph{\textbf{Weighted negative event precision ($prec_O$).}}
\textit{Proposition does not hold.} \\
\textbf{Reasoning.} Like with $prec_M$ and $prec_N$ the outcome of the replay procedure can be impacted by duplicate transitions and by silent transitions. Therefore, this proposition does not hold.

\paragraph{\textbf{Projected precision ($prec_P$).}}
\textit{Proposition holds.} \\
\textbf{Reasoning.} This technique translates the event log as well as the process model into deterministic finite automata before computing recall (recall that the minimal deterministic automaton is unique due to the Myhill--Nerode theorem). Therefore, it is independent of the representation of the model itself.

\paragraph{\textbf{Anti-alignment precision ($prec_Q$).}}
\textit{Proposition holds.} \\
\textbf{Reasoning.} The authors define an anti-alignment as a run of a model which differs sufficiently from the observed traces in a log. This anti-alignment is solely constructed based on the possible behavior of the process model and the observed behavior of the log. It is independent of the structure of the net.

\paragraph{\textbf{Eigenvalue precision ($prec_R$).}}
\textit{Proposition holds.} \\
\textbf{Reasoning.} The approach calculates precision based on the languages of the model and the language of the process model. This abstracts from the representation of the process model and, consequently, the proposition holds.

\subsubsection{Proposition 8 PrecPro1${}^{+}$}
\paragraph{\textbf{Soundness ($prec_H$).}}
\textit{Proposition holds.} \\
\textbf{Reasoning.} This proposition holds since, removing behavior from the model that does not happen in the event log decreases the set of traces allowed by the model $\card{\traces{m_1}} \geq \card{\traces{m_2}}$, while the set of traces of the event log complying with the model stays the same $\card{\traces{l} \cap  \traces{m_1}} = \card{\traces{l} \cap  \traces{m_2}}$.

\paragraph{\textbf{Simple behavioral appropriateness ($prec_I$).}}
\textit{Proposition does not hold.} \\
\textbf{Reasoning.} \textbf{BehPro$^{+}$} does not hold, which implies that \textbf{PrecPro1$^{+}$} does not hold.

\paragraph{\textbf{Advanced behavioral appropriateness ($prec_J$).}}
\textit{Proposition does not hold.} \\
\textbf{Reasoning.} \textbf{BehPro$^{+}$} does not hold, which implies that \textbf{PrecPro1$^{+}$} does not hold.

\paragraph{\textbf{ETC-one/ETC-rep ($prec_K$).}}
\textit{Proposition does not hold.} \\
\textbf{Reasoning.} A counter-example to Axiom 2 (as introduced in~\cite{Niek-IPL2018-imprecision}) was presented in~\cite{Niek-IPL2018-imprecision}. Since \textbf{PrecPro1$^+$} is a generalization of Axiom 2, the same counter-example shows that \textbf{PrecPro1$^+$} does not hold. Furthermore, \textbf{BehPro$^{+}$} does not hold, which implies that \textbf{PrecPro1$^{+}$} does not hold.

\paragraph{\textbf{ETC-all ($prec_L$).}}
\textit{Proposition does not hold.} \\
\textbf{Reasoning.} A counter-example to Axiom 2 (as introduced in~\cite{Niek-IPL2018-imprecision}) was presented in~\cite{Niek-IPL2018-imprecision}. Since \textbf{PrecPro1$^+$} is a generalization of Axiom 2, this implies that \textbf{PrecPro1$^+$} does not hold. Furthermore, \textbf{BehPro$^{+}$} does not hold, which implies that \textbf{PrecPro1$^{+}$} does not hold.

\paragraph{\textbf{Behavioral specificity ($prec_M$).}}
\textit{Proposition does not hold.} \\
\textbf{Reasoning.} \textbf{BehPro$^{+}$} does not hold, which implies that \textbf{PrecPro1$^{+}$} does not hold.

\paragraph{\textbf{Behavioral precision ($prec_N$).}}
\textit{Proposition does not hold.} \\
\textbf{Reasoning.} \textbf{BehPro$^{+}$} does not hold, which implies that \textbf{PrecPro1$^{+}$} does not hold.

\paragraph{\textbf{Weighted negative event precision ($prec_O$).}}
\textit{Proposition does not hold.} \\
\textbf{Reasoning.} A counter-example to Axiom 2 (as introduced in~\cite{Niek-IPL2018-imprecision}) was presented in~\cite{Niek-IPL2018-imprecision}. Since \textbf{PrecPro1$^+$} is a generalization of Axiom 2, this implies that \textbf{PrecPro1$^+$} does not hold. Furthermore, \textbf{BehPro$^{+}$} does not hold, which implies that \textbf{PrecPro1$^{+}$} does not hold.


\paragraph{\textbf{Projected precision ($prec_P$).}}
\textit{Proposition does not hold.} \\
\textbf{Reasoning.}  A counter-example to Axiom 2 (as introduced in~\cite{Niek-IPL2018-imprecision}) was presented in \cite{Niek-IPL2018-imprecision}. To illustrate this, the paper considers a model with a length-one-loop and its more precise corresponding model that unrolled the loop up to two executions. The DFA of the unrolled model will contain more states since the future allowed behavior depends on the number of executions of the looping activity, while the DFA of the initial model will contain only one state for this activity. This can cause that the unrolled model is considered less precise which violates the proposition.

\paragraph{\textbf{Anti-alignment precision ($prec_Q$).}}
\textit{Proposition holds.} \\
\textbf{Reasoning.} The behavior of the model that is not observed in the log will become the anti-alignment between the log and the model. The distance between the log and the anti-alignment is big which leads to low precision. If this behavior is removed from the model an anti-alignment closer to the log is found which leads to a higher precision.

\paragraph{\textbf{Eigenvalue precision ($prec_R$).}}
\textit{Proposition holds.} \\
\textbf{Reasoning.} Proven in Lemma 5.6 of~\cite{polyvyanyy-conf}.

\subsubsection{Proposition 9 PrecPro2${}^{+}$}
\paragraph{\textbf{Soundness ($prec_H$).}}
\textit{Proposition holds.} \\
\textbf{Reasoning.} Adding fitting behavior to the event log can lead to additional unique process executions that comply with the process model: $\card{\traces{l_1} \cap \traces{m}} \leq \card{\traces{m}} \leq \card{\traces{l_2}}$ . This cannot lower precision according to the definition of soundness. \linebreak $\frac{ \card{\traces{l_1} \cap  \traces{m}} }{\card{\traces{m}}}  \leq \frac{ \card{\traces{l_2} \cap  \traces{m}} }{\card{\traces{m}}}$, if $l_2 = l_1 \uplus l_3$.

\paragraph{\textbf{Simple behavioral appropriateness ($prec_I$).}}
\textit{Proposition does not hold.} \\
\textbf{Reasoning.} This approach does not consider whether the behavior of the log fits the model or not, but it focuses on the average number of enabled transitions during log replay. It is possible that the additional behavior enables a large number of transitions, this increases the average count and thereby lowers precision.

\paragraph{\textbf{Advanced behavioral appropriateness ($prec_J$).}}
\textit{Proposition holds.} \\
\textbf{Reasoning.} Adding fitting behavior to the log can only increase the intersection between the follow relations of the log and the model. $\card{ S^{l1}_F \cap S^{m}_F} \leq \card{ S^{l2}_F \cap S^{m}_F}$ and $\card{ S^{l1}_P \cap S^{m}_P} \leq \card{ S^{l2}_P \cap S^{m}_P}$, if $l_2 = l_1 \uplus l_3$ and $\traces{l_3} \subseteq \traces{m}$. Hence by adding fitting behavior to the log precision can only be increased.

\paragraph{\textbf{ETC-one/ETC-rep ($prec_K$).}}
\textit{Proposition does not hold.} \\
\textbf{Reasoning.} A counter-example to Axiom 5 (as introduced in~\cite{Niek-IPL2018-imprecision}) was presented in~\cite{Niek-IPL2018-imprecision}. Since \textbf{PrecPro2$^+$} is a generalization of Axiom 5, this implies that \textbf{PrecPro2$^+$} does not hold.

\begin{figure}[t]
	{
		\centering
		\subfloat[][]{\includegraphics[width=0.4\textwidth]{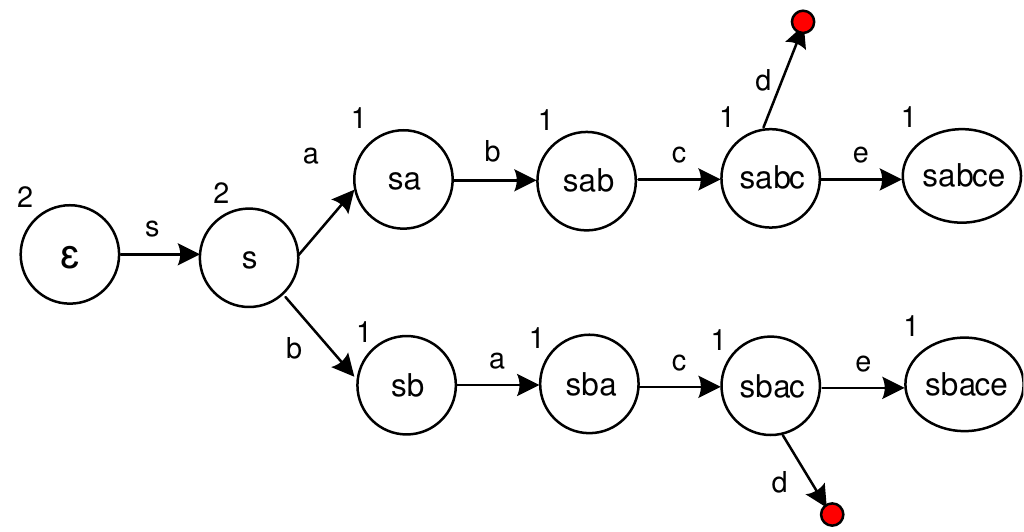}}
		\quad
		\subfloat[][]{\includegraphics[width=\textwidth]{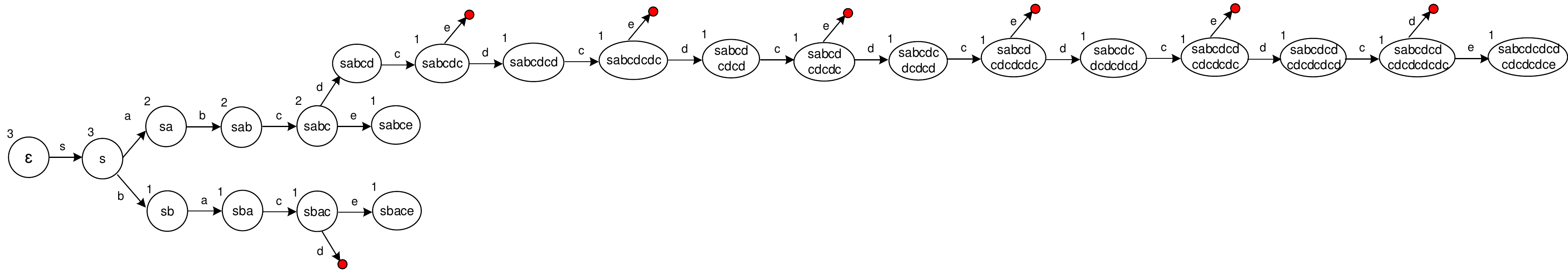}}
		\caption{Two alignment automata describing the state space of $m_1$ and $l_{15}$ (a) and the state space of $m_1$ and $l_{10}$ (b).
		}\label{autom-2}
	}
\end{figure}

\paragraph{\textbf{ETC-all ($prec_L$).}}
\textit{Proposition does not hold.} \\
\textbf{Reasoning.} The conclusions drawn in \cite{Niek-IPL2018-imprecision} for ETC-one can also be transferred to ETC-all. When adding fitting behavior it is possible that the new trace visits states that introduce a lot of new escaping edges. The increase in escaping edges is bigger than the increase in non-escaping ones which lowers precision. Consider process model $m_1$ in Figure~\ref{f-procmod-loop} (a) and the event log $l_{15} = [\la s,a,b,c,e \ra, \la s,b,a,c,e \ra ]$ and its extension with a fitting trace $l_{16} = l_{15} \uplus [\la s,a,b,c,d,c,d,c,d,c,d,c,d,c,e\ra ]$. Note, that the ``start" and ``end" activities of $m_1$ are abbreviated to ``s" and ``e". The corresponding automata in Figure~\ref{autom-2} show that the additional fitting trace adds additional states and escaping edges. This decreases precision: $prec_L(l_{15},m_1) = \frac{12}{14} = 0.857$ and $prec_L(l_{16},m_1) = \frac{31}{37} = 0.838$.

\begin{figure}[t]
	{
		\centering
		\includegraphics[width=0.5\textwidth]{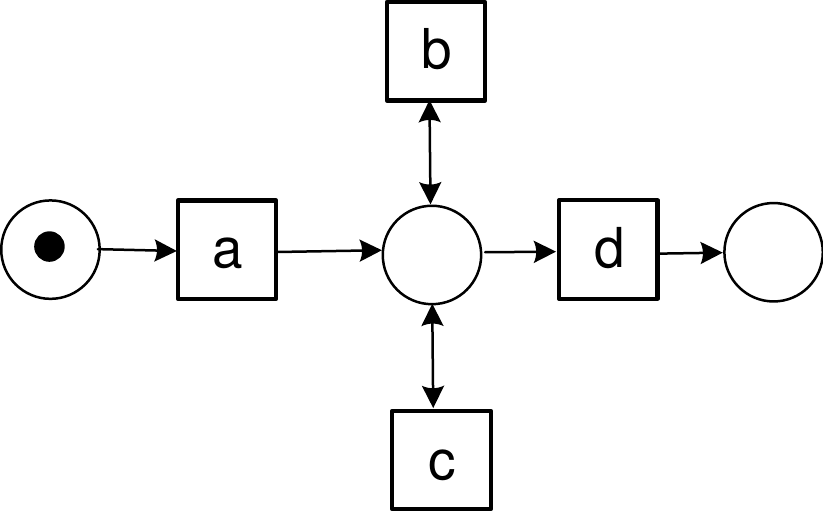}
		\caption{Process model $m_{10}$
		}\label{ce-neg}
	}
\end{figure}

\paragraph{\textbf{Behavioral specificity ($prec_M$).}}
\textit{Proposition does not hold.} \\
\textbf{Reasoning.} This proposition does not hold when the additional fitting trace introduces proportionally more negative events that could actually fire (FP) than correctly identified negative events (TN). To illustrate this, consider process model $m_{10}$ in Figure~\ref{ce-neg} and event log $l_{16} = [\la a,b,b,d \ra,\la a,b,c,d \ra]$. Table~\ref{tabneg1} shows the negative events calculated for the log. We assume a window size that equals the longest trace in the event log and we generate the negative events with probability 1. After replaying the log on the process model we record $FP(l_{16},m_{10})=10$ and $TN(l_{16},m_{10})=8$. Hence, $\frac{TN(l_{16},m_{10})}{TN(l_{16},m_{10})+FP(l_{16},m_{10})}= \frac{12}{22} = 0.545$.
We extend the log with fitting trace $l_10$, i.e., $l_{17} = l_{16} \uplus [\la a,b,c,b,b,b,b,b,d \ra ]$. The negative events calculated for $l_{17}$ are displayed in Table~\ref{tabneg2} and replaying it on $m_{10}$ results in $FP(l_{17},m_{10})=31$ and $TN(l_{17},m_{10})=23$. Consequently, $\frac{TN(l_{17},m_{10})}{TN(l_{17},m_{10})+FP(l_{17},m_{10})}= \frac{23}{54} = 0.426$. Although $l_{11}$ is fitting, it introduces more negative events that are actually enabled during replay. Therefore, $prec_M(l_{16},m_{10}) > prec_M(l_{17},m_{10})$, which violates the proposition.

\begin{table}[t] \caption{The traces of $l_{16}$ with the corresponding negative events.}
\label{tabneg1}
\centering
\begin{tabular}{c|c|c|c|}
\textbf{a} & \textbf{b}  & \textbf{b} & \textbf{d}   \\ \hline
$\neg$ b & $\neg$ a & $\neg$ a & $\neg$ a \\
$\neg$ c & $\neg$ c & $\neg$ d & $\neg$ b \\
$\neg$ d & $\neg$ d &          & $\neg$ c
\end{tabular}
\quad
\begin{tabular}{c|c|c|c}
\textbf{a} & \textbf{b} & \textbf{c} & \textbf{d}   \\ \hline
$\neg$ b & $\neg$ a & $\neg$ a & $\neg$ a \\
$\neg$ c & $\neg$ c & $\neg$ d & $\neg$ b \\
$\neg$ d & $\neg$ d &          & $\neg$ c
\end{tabular}
\end{table}

\begin{table}[t] \caption{The traces of $l_{17}$ with the corresponding negative events.}
\label{tabneg2}
\centering
\begin{tabular}{c|c|c|c|}
\textbf{a} & \textbf{b}  & \textbf{b} & \textbf{d}   \\ \hline
$\neg$ b & $\neg$ a & $\neg$ a & $\neg$ a \\
$\neg$ c & $\neg$ c & $\neg$ d & $\neg$ b \\
$\neg$ d & $\neg$ d &          & $\neg$ c
\end{tabular}
\quad
\begin{tabular}{c|c|c|c}
\textbf{a} & \textbf{b} & \textbf{c} & \textbf{d}   \\ \hline
$\neg$ b & $\neg$ a & $\neg$ a & $\neg$ a \\
$\neg$ c & $\neg$ c & $\neg$ d & $\neg$ b \\
$\neg$ d & $\neg$ d &          & $\neg$ c
\end{tabular}
\quad
\begin{tabular}{l|l|l|l|l|l|l|l|l}
\textbf{a} & \textbf{b}  & \textbf{c}   & \textbf{b}  & \textbf{b}    & \textbf{b}   & \textbf{b} & \textbf{b}   & \textbf{d} \\ \hline
$\neg$ b & $\neg$ a & $\neg$ a & $\neg$ a & $\neg$ a & $\neg$ a & $\neg$ a & $\neg$ a & $\neg$ a \\
$\neg$ c & $\neg$ c & $\neg$ d & $\neg$ c & $\neg$ c & $\neg$ c & $\neg$ c & $\neg$ c & $\neg$ b \\
$\neg$ d & $\neg$ d &          & $\neg$ d & $\neg$ d & $\neg$ d & $\neg$ d & $\neg$ d & $\neg$ c
\end{tabular}
\end{table}

\paragraph{\textbf{Behavioral precision ($prec_N$).}}
\textit{Proposition does not hold.} \\
\textbf{Reasoning.} The counter-example of $prec_M$ and can also be used to show that this proposition is violated. During replay of $l_{16}$ and $l_{17}$ we also count the positive events that can be correctly replayed (TP). This results in $TP(l_{16},m_{10}) = 8$ and \linebreak $TP(l_{16},m_{10}) = 17$. When we calculate precision, we obtain $prec_N(l_{16},m_{10}) = \frac{TP(l_{16},m_{10})}{ TP(l_{16},m_{10}) + FP(l_{16},m_{10})} = \frac{8}{8+10} = 0.44$ and $prec_N(l_{17},m_{10}) = \frac{17}{17+31} = 0.35$. The additional fitting trace lowers precision: $prec_N(l_{16},m_{10}) > prec_N(l_{17},m_{10})$.

\paragraph{\textbf{Weighted negative event precision ($prec_O$).}}
\textit{Proposition does not hold.} \\
\textbf{Reasoning.} Consider the same counter-example as that we provided for $prec_M$. Negative events are weighted by the size of the longest matching window of events. For the long repetition of $b$-events in $\la a,b,c,b,b,b,b,b,d\ra$, the negative events for $a$, $c$ and $d$ have weight $2$ due to the 2 consecutive b's in trace $\la a,b,b,d\ra$. Since the negative events in $l_3$ that caused the precision to go up when $l_3$ was added to $l_1$ have above average weight, the weighting does not invalidate the counter-example.


\paragraph{\textbf{Projected precision ($prec_P$).}}
\textit{Proposition does not hold.} \\
\textbf{Reasoning.} A counter-example to Axiom 5 (as introduced in~\cite{Niek-IPL2018-imprecision}) was presented in~\cite{Niek-IPL2018-imprecision}. Since \textbf{PrecPro2$^+$} is a generalization of Axiom 5, this implies that \textbf{PrecPro2$^+$} does not hold.

\paragraph{\textbf{Anti-alignment precision ($prec_Q$).}}
\textit{Proposition does not hold.} \\
\textbf{Reasoning.} A counter-example to Axiom 5 (as introduced in~\cite{Niek-IPL2018-imprecision}) was presented in~\cite{Niek-IPL2018-imprecision}. Since \textbf{PrecPro2$^+$} is a generalization of Axiom 5, this implies that \textbf{PrecPro2$^+$} does not hold.


\paragraph{\textbf{Eigenvalue precision ($prec_R$).}}
\textit{Proposition holds.} \\
\textbf{Reasoning.} Proven in Lemma 5.6 of~\cite{polyvyanyy-conf}.

\subsubsection{Proposition 10 PrecPro3${}^{0}$}
\paragraph{\textbf{Soundness ($prec_H$).}}
\textit{Proposition holds.} \\
\textbf{Reasoning.} Adding non-fitting behavior to the event log cannot lead to additional process executions, which comply with the process model. Hence, it cannot lower precision according to the definition of soundness. $\card{\traces{l_1} \cap \traces{m}} = \card{\traces{l_2} \cap \traces{m}}$, if $l_2 = l_1 \uplus l_3$ and $\traces{l_3} \subseteq \compltr{m}$.

\paragraph{\textbf{Simple behavioral appropriateness ($prec_I$).}}
\textit{Proposition does not hold.} \\
\textbf{Reasoning.} Adding non-fitting behavior to the event log might change the average number of enabled transitions during log replay and therefore precision. Note that this scenario was not considered by the approach since the authors assume a fitting log \cite{anne_confcheck_is}.

\paragraph{\textbf{Advanced behavioral appropriateness ($prec_J$).}}
\textit{Proposition does not hold.} \\
\textbf{Reasoning.} This approach records relations between activities and compares these relations between the model and the log. Hence, it does not consider entire traces to be fitting or non-fitting but refines it to an activity level. Therefore, it is possible that non-fitting traces contain fitting events that improve precision. For example, the non-fitting traces change a \textit{never} follows relation of the event log to a \textit{sometimes} follows relation that matches the process model. Consequently, precision increases and violates this proposition.

\paragraph{\textbf{ETC-one/ETC-rep ($prec_K$).}}
\textit{Proposition does not hold.} \\
\textbf{Reasoning.} Before the alignment automaton is constructed the log is aligned to ensure that the traces fit the model. Adding non-fitting behavior can possibly lead to new alignments that lead to new escaping edges and change precision.

Consider model $m_7$ in Figure~\ref{ce-mod3} and alignment automata in Figure~\ref{autom-1} (a) corresponding to trace $\la a,b,c,g \ra$. Adding the unfitting trace $\la a,d,g \ra$ results in the aligned trace $\la a,d,e,g \ra$ or $\la a,d,f,g \ra$. Either of the two aligned traces introduces new states into the alignment automaton. Additionally, the new trace alters the weights of each state. Therefore, even though both automata contain one escaping edge, precision changes.

\paragraph{\textbf{ETC-all ($prec_L$).}}
\textit{Proposition does not hold.} \\
\textbf{Reasoning.} The counter-example presented for $prec_K$ also shows that this proposition does not hold for this approach. The unfitting trace $\la a,d,g \ra$ results in the aligned trace $\la a,d,e,g \ra$ or $\la a,d,f,g \ra$. This variant of ETC precision uses both of the two aligned traces to construct the alignment automaton. This introduces new states, alters the weights of the states, removes the escaping edge and changes precision.

\paragraph{\textbf{Behavioral specificity ($prec_M$).}}
\textit{Proposition does not hold.} \\
\textbf{Reasoning.}  Negative events describe behavior that was not allowed during process execution. They are constructed based on the behavior observed in the event log. From this point of view, non-fitting behavior is behavior that should have been described by negative events, but since it was observed in the event log, the algorithm does not define it as negative events anymore. Hence adding non-fitting behavior $l_3$ to event log $l_1$ decreases the number of correctly identified negative events (TN) in the traces of $l_1$.

Furthermore, this measure accounts for the number of negative events that actually could fire during trace replay (FP). These false positives are caused by the fact that behavior is shown in the model but not observed in the log. Although the trace is not fitting when considered as a whole, certain parts of the trace can fit the model, and these parts can represent the previously missing behavior in the event log that leads to the wrong classification of negative events. Adding these non-fitting traces $l_3$ can, therefore, lead to a decrease in false positives in the traces of $l_1$ and changes precision.

\paragraph{\textbf{Behavioral precision ($prec_N$).}}
\textit{Proposition does not hold.} \\
\textbf{Reasoning.} As shown in the reasoning for $prec_M$, adding non-fitting traces $l_3$ to a fitting log $l_1$ can decrease the number of false positives FP in the negatives events that were generated for the traces of $l_1$.

\paragraph{\textbf{Weighted negative event precision ($prec_O$).}}
\textit{Proposition does not hold.} \\
\textbf{Reasoning.} As shown in the reasoning for $prec_M$, adding non-fitting traces $l_3$ to a fitting log $l_1$ can decrease the number of false positives FP in the negatives events that were generated for the traces of $l_1$. Weighing the negative events does not change this.

\paragraph{\textbf{Projected precision ($prec_P$).}}
\textit{Proposition does not hold.} \\
\textbf{Reasoning.} Since projected precision calculates precision based on several projected sub-models and sub-logs, it is possible that unfitting behavior fits some of these sub-models locally. To illustrate this consider a model with the language $\traces{m_{11}} =\{\la a,b\ra, \linebreak \la c,d \ra\}$ and $l_{18}=[\la a,b \ra]$. It is clear, that the model is not perfectly precise since trace $\la c,d \ra$ is not observed in the event log. Hence, if we project our model and log on $\{c,d\}$, precision will be 0 for this projection.

We extend the log with an unfitting trace $l_{19}= l_{18} \uplus [\la c,d,a \ra]$. Projecting $l_{19}$ on $\{c,d\}$ results in a precision value of 1. Since this approach aggregates the precision over several projections, it is clear, that unfitting behavior can improve precision.

\paragraph{\textbf{Anti-alignment precision ($prec_Q$).}}
\textit{Proposition does not hold.} \\
\textbf{Reasoning.} By definition anti-alignments always fit the model. Consequently, there will always be a distance between a non-fitting trace and an anti-alignment. Adding non-fitting behavior to the event log will, therefore, change precision. The proposition does not require $l_1$ to be fitting. Therefore it could be the case that $l_1$ has a trace that has a higher distance to behavior that is allowed by $m$ than what can be found amongst the traces of $l_3$. Note that this scenario is not considered by the approach since the authors assume a fitting log \cite{anti-align-bpm2016}. However, also after aligning the log, it might still change precision by resulting in alignments that were not contained in the initial log.

\paragraph{\textbf{Eigenvalue precision ($prec_R$).}}
\textit{Proposition holds.} \\
\textbf{Reasoning.} The measure is defined as $\frac{eig(\traces{m}\cap\traces{l})}{eig(\traces{m})}$. As $\traces{m}\cap\traces{l}$ does not change when adding non-fitting traces to $l$, neither does the measure change.

\subsubsection{Proposition 11 PrecPro4${}^{0}$}
\paragraph{\textbf{Soundness ($prec_H$).}}
\textit{Proposition holds.} \\
\textbf{Reasoning.} Since the approach only considers unique model executions, duplicating the log has no effect on precision.

\paragraph{\textbf{Simple behavioral appropriateness ($prec_I$).}}
\textit{Proposition holds.} \\
\textbf{Reasoning.} Since the number of traces $n_i$ is present in the denominator as well as the nominator of the formula duplication has no effect on precision. $\frac{\sum_{i=1}^{k}n_i (\card{T_V}-x_i)}{(\card{T_V} - 1) \cdot \sum_{i=1}^{k}n_i}$. Here we assume that if a trace is perfectly replayable it will also be replayed perfectly.

\paragraph{\textbf{Advanced behavioral appropriateness ($prec_J$).}}
\textit{Proposition holds.} \\
\textbf{Reasoning.} The sometimes follows relations will not change by duplicating the event log. Hence, the result is unaffected.

\paragraph{\textbf{ETC-one/ETC-rep ($prec_K$).}}
\textit{Proposition holds.} \\
\textbf{Reasoning.} The weight of escaping and non-escaping edges is calculated based on the trace frequency. However, since the distribution of the traces does not change the weight of both edge types grows proportionally and precision does not change.

\paragraph{\textbf{ETC-all ($prec_L$).}}
\textit{Proposition holds.} \\
\textbf{Reasoning.} The weight of escaping and non-escaping edges is calculated based on the trace frequency. However, since the distribution of the traces does not change the weight of both edge types grows proportionally and precision does not change.

\paragraph{\textbf{Behavioral specificity ($prec_M$).}}
\textit{Proposition holds.} \\
\textbf{Reasoning.}  Multiplying the event log $k$ times leads to a proportional increase in true negatives and false positives. Consequently, precision does not change. \linebreak $\frac{k \times TN(l,m)}{k \times (TN(l,m) + FP(l,m)} = \frac{TN(l,m)}{TN(l,m) + FP(l,m)}$, hence $prec_M(l^k,m) = prec_M(l,m)$.

\paragraph{\textbf{Behavioral precision ($prec_N$).}}
\textit{Proposition holds.} \\
\textbf{Reasoning.} Multiplying the event log $k$ times leads to a proportional increase in true positives and false positives. Consequently, precision does not change. \linebreak $\frac{k \times TP(l,m)}{k \times (TP(l,m) + FP(l,m))} = \frac{TP(l,m)}{TP(l,m) + FP(l,m)}$, hence $prec_N(l^k,m) = prec_N(l,m)$.

\paragraph{\textbf{Weighted negative event precision ($prec_O$).}}
\textit{Proposition holds.} \\
\textbf{Reasoning.} Multiplying the event log $k$ times leads to a proportional increase in true negatives and false positives. Consequently precision does not change. \linebreak $\frac{k \times TN(l,m)}{k \times (TN(l,m) + FP(l,m)} = \frac{TN(l,m)}{TN(l,m) + FP(l,m)}$, hence $prec_M(l^k,m) = prec_M(l,m)$.

\paragraph{\textbf{Projected precision ($prec_P$).}}
\textit{Proposition holds.} \\
\textbf{Reasoning.} The precision measure does not consider trace frequency and therefore will not be changed by duplicating the event log.

\paragraph{\textbf{Anti-alignment precision ($prec_Q$).}}
\textit{Proposition holds.} \\
\textbf{Reasoning.} This approach sums for each trace in the log the distance between anti-alignment and trace. This sum is averaged over the number of traces in the log and consequently, duplication of the log will not change precision.

\paragraph{\textbf{Eigenvalue precision (R.)}}
\textit{Proposition holds.} \\
\textbf{Reasoning.} Eigenvalue precision is defined purely on the language of the log and the model and it does not take into account trace frequencies in the log, therefore, this proposition holds.

\subsubsection{Proposition 12 PrecPro5${}^{+}$}
\paragraph{\textbf{Soundness ($prec_H$).}}
\textit{Proposition holds.} \\
\textbf{Reasoning.} If the model allows for the behavior observed and nothing more each, unique process execution corresponds to a unique path through the model $\traces{l} = \traces{m}$. Therefore precision is maximal: $\card{\traces{l}} / \card{\traces{m}} = 1$.

\paragraph{\textbf{Simple behavioral appropriateness ($prec_I$).}}
\textit{Proposition does not hold.} \\
\textbf{Reasoning.} This approach only considers strictly sequential models to be perfectly precise. If the model has choices, loops or concurrency, then, multiple transitions might be enabled during replay even if the model only allows only for the observed behavior. As a result, precision is not maximal.

\paragraph{\textbf{Advanced behavioral appropriateness ($prec_J$).}}
\textit{Proposition holds.} \\
\textbf{Reasoning.} If the model allows for only the behavior observed and nothing more, the set of sometimes follows/precedes relations of the model are equal to the ones of the event log. $ S^{l}_{F} \cap S^{m}_{F} = S^{m}_{F}$ and  $S^{l}_{P} \cap S^{m}_{P}= S^{m}_{P}$, if $\traces{l} = \traces{m}$. Consequently precision is maximal.

\paragraph{\textbf{ETC-one/ETC-rep ($prec_K$).}}
\textit{Proposition does not hold.} \\
\textbf{Reasoning.} Consider a model with a choice between two a-labeled transitions and a trace $[\la a \ra]$. When constructing the alignment automaton, there will be an escaping edge for the other a-labeled transition. Similar problems may arise with silent transitions.

\paragraph{\textbf{ETC-all ($prec_L$).}}
\textit{Proposition does not hold.} \\
\textbf{Reasoning.} The counter-example for $prec_K$ shows that also $prec_L$ violates this proposition.

\paragraph{\textbf{Behavioral specificity ($prec_M$).}}
\textit{Proposition holds.} \\
\textbf{Reasoning.} When the model allows for only observed behavior (i.e., in $l$), then $\text{FP}(l,m)=0$, as false positives are caused by the fact that behavior is shown in the model but not observed in the log. Therefore, $\traces{l} = \traces{m}\implies prec_M(l,m)=\frac{\text{TN}(l,m)}{\text{TN}(l,m) + \text{FP}(l,m) } = \frac{\text{TN}(l,m)}{\text{TN}(l,m) + 0}  = 1$.

\paragraph{\textbf{Behavioral precision ($prec_N$).}}
\textit{Proposition holds.} \\
\textbf{Reasoning.} When the model allows for only observed behavior (i.e., in $l$), then $\text{FP}(l,m)=0$, as false positives are caused by the fact that behavior is shown in the model but not observed in the log. Therefore, $\traces{l} = \traces{m}\implies prec_N(l,m)=\frac{\text{TP}(l,m)}{\text{TP}(l,m) + \text{FP}(l,m) } = \frac{\text{TP}(l,m)}{\text{TP}(l,m) + 0}  = 1$.

\paragraph{\textbf{Weighted negative event precision ($prec_O$).}}
\textit{Proposition holds.} \\
\textbf{Reasoning.} See the reasoning for $prec_N$ above. The weighing of negative events doesn't change the fact that false positives cannot occur when  $\traces{l} = \traces{m}$, therefore the same reasoning applies to $prec_O$.

\paragraph{\textbf{Projected precision ($prec_P$).}}
\textit{Proposition holds.} \\
\textbf{Reasoning.} If the model allows for only the behavior observed and nothing more, the two automata describing the behavior of the log and the model are exactly the same: $\text{DFA}(m|_A) = \text{DFA}(l|_A) = \text{DFAc}(l,m,A)$, if $\traces{l} = \traces{m}$. Hence, precision is maximal.

\paragraph{\textbf{Anti-alignment precision ($prec_Q$).}}
\textit{Proposition holds.} \\
\textbf{Reasoning.} If the model allows for the behavior observed and nothing more, each anti-alignment will exactly match its corresponding trace. Consequently, the distance between the log and the anti-alignment is minimal and precision maximal.

\paragraph{\textbf{Eigenvalue precision ($prec_R$).}}
\textit{Proposition holds.} \\
\textbf{Reasoning.} Proven in Corollary 5.15 of~\cite{polyvyanyy-conf}.

\subsubsection{Proposition 13 PrecPro6${}^{0}$}
\paragraph{\textbf{Soundness ($prec_H$).}}
\textit{Proposition holds.} \\
\textbf{Reasoning.} If the log contains non-fitting behavior and all modeled behavior was observed, the set of traces in the event log complying with the process model equals the paths through the model. $\card{\traces{l} \cap \traces{m}} = \card{\traces{m}}$ and precision is maximal.

\paragraph{\textbf{Simple behavioral appropriateness ($prec_I$).}}
\textit{Proposition does not hold.} \\
\textbf{Reasoning.} This approach only considers strictly sequential models to be perfectly precise. If the model contains choices, concurrency, loops, etc., multiple transitions may be enabled, lowering precision. Moreover, the approach assumes all behavior to be fitting. Hence, behavior that is observed and not modeled is likely to lead to problems.

\paragraph{\textbf{Advanced behavioral appropriateness ($prec_J$).}}
\textit{Proposition holds.} \\
\textbf{Reasoning.} Non-fitting behavior cannot affect the follow relations of the process model. Furthermore, it does not influence the \textit{sometimes} follows/precedes relations of the event log if all the modeled behavior was observed. Hence, the sets of \textit{sometimes} follows/precedes relations of the log and the model are equal to each other. $ S^{l}_{F} \cap S^{m}_{F} = S^{m}_{F}$ and  $S^{l}_{P} \cap S^{m}_{P}= S^{m}_{P}$ and, therefore, precision is maximal.

\paragraph{\textbf{ETC-one/ETC-rep ($prec_K$).}}
\textit{Proposition does not hold.} \\
\textbf{Reasoning.} The counter-example from \textbf{PrecPro5$^{+}$} also shows that \textbf{PrecPro6$^{0}$} is violated.

\paragraph{\textbf{ETC-all ($prec_L$).}}
\textit{Proposition does not hold.} \\
\textbf{Reasoning.} The counter-example from \textbf{PrecPro5$^{+}$} also shows that \textbf{PrecPro6$^{0}$} is violated.

\paragraph{\textbf{Behavioral specificity ($prec_M$).}}
\textit{Proposition holds.} \\
\textbf{Reasoning.} When the model allows for only observed behavior (i.e., in $l$), then $\text{FP}(l,m)=0$, as false positives are caused by the fact that behavior is shown in the model but not observed in the log. Therefore, $\traces{m} \subseteq \traces{l} \implies prec_M(l,m)=\frac{\text{TN}(l,m)}{\text{TN}(l,m) + \text{FP}(l,m) } = \frac{\text{TN}(l,m)}{\text{TN}(l,m) + 0}  = 1$.

\paragraph{\textbf{Behavioral precision ($prec_N$).}}
\textit{Proposition holds.} \\
\textbf{Reasoning.} When the model allows for only observed behavior (i.e., in $l$), then $\text{FP}(l,m)=0$, as false positives are caused by the fact that behavior is shown in the model but not observed in the log. Therefore, $\traces{m} \subseteq \traces{l} \implies prec_M(l,m)=\frac{\text{TP}(l,m)}{\text{TP}(l,m) + \text{FP}(l,m) } = \frac{\text{TN}(l,m)}{\text{TN}(l,m) + 0}  = 1$.

\paragraph{\textbf{Weighted negative event precision ($prec_O$).}}
\textit{Proposition holds.} \\
\textbf{Reasoning.} See the reasoning for $prec_N$ above. The weighing of negative events doesn't change the fact that false positives cannot occur when  $\traces{m}\subseteq\traces{l}$, therefore the same reasoning applies to $prec_O$.

\paragraph{\textbf{Projected precision ($prec_P$).}}
\textit{Proposition holds.} \\
\textbf{Reasoning.} If the all modeled behavior is observed, the automaton describing the model and the conjunctive automaton of the model and the log are exactly the same. $\text{DFAc}(S,M,A) \setminus \text{DFA}(M|_A) = \emptyset$ , if $\traces{m} \subseteq \traces{l}$. Hence, precision is maximal. Furthermore, the authors define that precision is 1 if the model is empty \cite{sander-scalable-procmin-SOSYM}.

\paragraph{\textbf{Anti-alignment precision ($prec_Q$).}}
\textit{Proposition holds.} \\
\textbf{Reasoning.} By definition, an anti-alignment will always fit the model. Consequently, when computing the distance between the unfitting trace and the anti-alignment, it will never be minimal. However note, that scenarios with unfitting behavior were not considered by the approach since the authors assume a fitting log. After aligning the log, it contains exactly the modeled behavior $\traces{l} = \traces{m}$ and precision is maximal.

\paragraph{\textbf{Eigenvalue precision ($prec_R$).}}
\textit{Proposition holds.} \\
\textbf{Reasoning.} Corollary 5.15 of~\cite{polyvyanyy-conf} proves that $prec_R(l,m)=1$ when $\traces{m}=\traces{l}$. From the definition of the precision measure (i.e., $prec_R(l,m)=\frac{eig(\traces{m}\cap\traces{l})}{eig(\traces{m})}$) it follows that the proposition holds, as the numerator $eig(\traces{m}\cap\traces{l})$ is equal to $eig(\traces{m})$ when $\traces{m}\subseteq\traces{l}$.

\subsection{Generalization}\label{appendixgen}
\subsubsection{Proposition 1 DetPro${}^{+}$}
\paragraph{\textbf{Alignment generalization ($gen_S$).}}
\textit{Proposition holds.} \\
\textbf{Reasoning.} Generalization is calculated based on the states visited by the process. The approach counts how often each state is visited ($n$) and how many different activities were observed in this state ($w$). These two numbers can be obtained from the model and the log at all times. Hence generalization is deterministic. Note, that we assume in case of non-fitting behavior that the log is first aligned before evaluating this proposition.

\paragraph{\textbf{Weighted negative event generalization ($gen_T$).}}
\textit{Proposition does not hold.} \\
\textbf{Reasoning.} If duplicate or silent transitions are encountered during the replay of the trace, which was enhanced with negative events, the approach explores which of the available transitions enables the next event in the trace. If no solution is found', one of the transitions is randomly fired. Hence precision also depends on the representation of the model.

\paragraph{\textbf{Anti-alignment generalization ($gen_U$).}}
\textit{Proposition holds.} \\
\textbf{Reasoning.} Precision is computed based on the maximal anti-alignment. Even if there are multiple maximal anti-alignments the distance will always be maximal and, therefore, precision is deterministic. Note, that we assume in case of non-fitting behavior that the log is first aligned before evaluating this proposition.

\subsubsection{Proposition 2 BehPro${}^{+}$}
\paragraph{\textbf{Alignment generalization ($gen_S$).}}
\textit{Proposition holds.} \\
\textbf{Reasoning.} The approach abstracts from the concrete representation of the process models. A key element is the function $state_M$ which is a parameter of the approach and maps each event onto the state in which the event occurred. This function only uses behavioral properties. Hence, the proposition holds.

\paragraph{\textbf{Weighted negative event generalization ($gen_T$).}}
\textit{Proposition does not hold.} \\
\textbf{Reasoning.} If duplicate or silent transitions are encountered while replaying a trace, the approach checks if one of the available transitions enables the next event in the trace. Whether this is the case can depend on the structure of the model.

\paragraph{\textbf{Anti-alignment generalization ($gen_U$).}}
\textit{Proposition does not hold.} \\
\textbf{Reasoning.} This approach defines a so-called recovery distance which measures the distance between the states of the anti-alignment and the states visited by the log. It defines a state as a marking of the Petri net. One can think of two Petri nets with the same behavior but different markings based on their structure. The two process models presented in Figure~\ref{ce-mod4} can be used as examples. Therefore generalization depends on the representation of the process model.

\subsubsection{Proposition 14 GenPro1${}^{+}$}
\paragraph{\textbf{Alignment generalization ($gen_S$).}}
\textit{Proposition does not hold.} \\
\textbf{Reasoning.} The approach does not allow for unfitting behavior and therefore aligns the log with the process model. These aligned traces might visit states that have already been observed by the fitting behavior, which increases the number of visits $n$ to these states and improves generalization. The extension of the model such that $\traces{m_1} \subseteq \traces{m_2}$ might cause this previously unfitting behavior to fit $m_2$ and aligning the log is not necessary anymore. However, these ``missing" aligned traces cause a decrease in the number of visits $n$ to each state of the previously aligned trace and generalization decreases.

\paragraph{\textbf{Weighted negative event generalization ($gen_T$).}}
\textit{Proposition does not hold.} \\
\textbf{Reasoning.} \textbf{BehPro$^{+}$} does not hold, which implies that \textbf{GenPro1$^{+}$} does not hold

\paragraph{\textbf{Anti-alignment generalization ($gen_U$).}}
\textit{Proposition does not hold.} \\
\textbf{Reasoning.} \textbf{BehPro$^{+}$} does not hold, which implies that \textbf{GenPro1$^{+}$} does not hold.

\subsubsection{Proposition 15 GenPro2${}^{+}$}
\paragraph{\textbf{Alignment generalization ($gen_S$).}}
\textit{Proposition does not hold.} \\
\textbf{Reasoning.} According to the definition of generalization, it is possible that additional fitting behavior in the event log decreases generalization if the additional traces introduce new unique events to the log and $pnew(w,n) = 1$. Hence, the new traces raise the number of unique activities ($w$) in state $s$  while the number of times $s$ was visited by the event log stays low ($n$).

\paragraph{\textbf{Weighted negative event generalization ($gen_T$).}}
\textit{Proposition does not hold.} \\
\textbf{Reasoning.} The fitting behavior can lead to the generation of additional negative events. If these negative events are correctly identified, they increase the value of disallowed generalizations (DG).  $AG(l_1,m) = AG(l_2,m)$ and $DG(l_1,m) < DG(l_2,m)$ which decreases generalization $\frac{AG(l_1,m)}{AG(l_1,m) + DG(l_1,m)} > \frac{AG(l_2,m)}{AG(l_2,m) + DG(l_2,m)}$.

\paragraph{\textbf{Anti-alignment generalization ($gen_U$).}}
\textit{Proposition does not hold.} \\
\textbf{Reasoning.} The approach defines the perfectly generalizing model as a model with a maximal anti-alignment distance $d$ and minimal recovery distance $d_{rec}$. The newly observed behavior of the general model should introduce new paths between states but no new states \cite{anti-align-bpm2016}.
However, if the model is very imprecise and with a lot of different states, it is possible that the added traces visit very different states than the anti-alignment, generalization will be low for these traces. Consequently, the average generalization over all traces decreases.

\subsubsection{Proposition 16 GenPro3${}^{0}$}
\paragraph{\textbf{Alignment generalization ($gen_S$).}}
\textit{Proposition does not hold.} \\
\textbf{Reasoning.} According to this approach, generalization is not defined if there are unfitting traces, since unfitting behavior cannot be mapped to states of the process model. Note that unfitting behavior was intentionally excluded from the approach and the authors state that unfitting event logs should be preprocessed to fit to the model. However, after the log is aligned, the added traces might improve generalization by increasing $n$ the times how often certain states are visited while executing events that have already been observed by the fitting traces in the log.

\paragraph{\textbf{Weighted negative event generalization (S).}}
\textit{Proposition does not hold.} \\
\textbf{Reasoning.} In this approach, negative events are assigned a weight which indicates how certain the log is about these events being negative ones. Even though the added behavior is non-fitting it might still provide evidence for certain negative events and therefore increase their weight. If these events are not enabled during log replay the value for disallowed generalizations (DG) decreases $DG(l_1,m) > DG(l_2,m)$ and generalization improves: $\frac{AG(l,m)}{AG(l,m) + DG(l_1,m)} < \frac{AG(l,m)}{AG(l,m) + DG(l_2,m)}$.

\paragraph{\textbf{Anti-alignment generalization (R).}}
\textit{Proposition does not hold.} \\
\textbf{Reasoning.} According to this approach, generalization is not defined if there are unfitting traces since unfitting behavior cannot be mapped to states of the process model. Note that the authors exclude unfitting behavior from this approach and state that unfitting event logs should be preprocessed to fit to the model. But after aligning the event log, it might be the case that the added and aligned traces report a big distance to the anti-alignment without introducing new states, which increases generalization.

\subsubsection{Proposition 17 GenPro4${}^{+}$}
\paragraph{\textbf{Alignment generalization ($gen_S$).}}
\textit{Proposition holds.} \\
\textbf{Reasoning.} Multiplying a fitting log $k$ times will result in more visits $n$ to each state while the number of different activities observed $w$ stays the same and generalization increases.

\paragraph{\textbf{Weighted negative event generalization ($gen_T$).}}
\textit{Proposition holds.} \\
\textbf{Reasoning.} Multiplying the log $k$ times will proportionally increase the number of allowed and disallowed generalizations and therefore not change generalization: \linebreak $\frac{k \times AG(l,m)}{k \times (AG(l,m) + DG(l,m))} = \frac{AG(l,m)}{AG(l,m) + DG(l,m) }$, $gen_T(l^k,m) = gen_T(l,m)$.

\paragraph{\textbf{Anti-alignment generalization ($gen_U$).}}
\textit{Proposition holds.} \\
\textbf{Reasoning.} This approach sums for each trace in the log the trace-based generalization. This sum is averaged over the number of traces in the log and consequently, duplication of the log will not change precision.


\subsubsection{Proposition 18 GenPro5${}^{+}$}
\paragraph{\textbf{Alignment generalization ($gen_S$).}}
\textit{Proposition does not hold.} \\
\textbf{Reasoning.} According to this approach, generalization is not defined if there are unfitting traces, since they cannot be mapped to states of the process model. Note that unfitting behavior was intentionally excluded from the approach and the authors state that unfitting event logs should be preprocessed to fit the model. However, after the log is aligned the added traces might improve generalization by increasing $n$ the times how often certain states are visited while being aligned to traces that have already been observed by the other traces in the log. Hence, generalization increases.

\paragraph{\textbf{Weighted negative event generalization ($gen_T$).}}
\textit{Proposition holds.} \\
\textbf{Reasoning.} Multiplying the log $k$ times will proportionally increase the number of allowed and disallowed generalizations and therefore not change generalization: \linebreak $\frac{k \times AG(l,m)}{k \times (AG(l,m) + DG(l,m))} = \frac{AG(l,m)}{AG(l,m) + DG(l,m) }$, $gen_T(l^k,m) = gen_T(l,m)$.

\paragraph{\textbf{Anti-alignment generalization ($gen_U$).}}
\textit{Proposition holds.} \\
\textbf{Reasoning.} According to this approach, generalization is not defined if there are unfitting traces since unfitting behavior cannot be mapped to states of the process model. Note, that the authors exclude unfitting behavior from this approach and state that unfitting event logs should be preprocessed to fit the model. Therefore we evaluate this proposition after the event log was aligned. Duplicating the aligned log will not change generalization since the sum of trace-generalization is averaged over the number of traces in the log.

\subsubsection{Proposition 19 GenPro6${}^{0}$}
\paragraph{\textbf{Alignment generalization ($gen_S$).}}
\textit{Proposition holds.} \\
\textbf{Reasoning.} According to this approach, generalization is not defined if there are unfitting traces since they cannot be mapped to states of the process model. Note that unfitting behavior was intentionally excluded from the approach and the authors state that unfitting event logs should be preprocessed to fit the model. Duplicating the aligned log will result in more visits to each state visited by the log. Generalization increases.

\paragraph{\textbf{Weighted negative event generalization ($gen_T$).}}
\textit{Proposition holds.} \\
\textbf{Reasoning.} Multiplying the log $k$ times will proportionally increase the number of allowed and disallowed generalizations and therefore not change generalization: \linebreak $\frac{k \times AG(l,m)}{k \times (AG(l,m) + DG(l,m))} = \frac{AG(l,m)}{AG(l,m) + DG(l,m) }$, $gen_T(l^k,m) = gen_T(l,m)$.

\paragraph{\textbf{Anti-alignment generalization ($gen_U$).}}
\textit{Proposition holds.} \\
\textbf{Reasoning.} According to this approach, generalization is not defined if there are unfitting traces since unfitting behavior cannot be mapped to states of the process model. Note that the authors exclude unfitting behavior from this approach and state that unfitting event logs should be preprocessed to fit the model. Duplicating the aligned log will not change generalization since the sum of trace-generalization is averaged over the number of traces in the log.

\subsubsection{Proposition 20 GenPro7${}^{0}$}
\paragraph{\textbf{Alignment generalization ($gen_S$).}}
\textit{Proposition does not hold.} \\
\textbf{Reasoning.} According to this approach, generalization is not defined if there are unfitting traces since they cannot be mapped to states of the process model. Note that unfitting behavior was intentionally excluded from the approach and the authors state that unfitting event logs should be preprocessed to fit the model. Duplicating an aligned log will result in more visits to each state. Hence, generalization increases and violates the proposition.

\paragraph{\textbf{Weighted negative event generalization ($gen_T$).}}
\textit{Proposition holds.} \\
\textbf{Reasoning.} Multiplying the log $k$ times will proportionally increase the number of allowed and disallowed generalizations and therefore not change generalization: \linebreak $\frac{k \times AG(l,m)}{k \times (AG(l,m) + DG(l,m))} = \frac{AG(l,m)}{AG(l,m) + DG(l,m) }$, $gen_T(l^k,m) = gen_T(l,m)$.

\paragraph{\textbf{Anti-alignment generalization ($gen_U$).}}
\textit{Proposition holds.} \\
\textbf{Reasoning.} According to this approach, generalization is not defined if there are unfitting traces since unfitting behavior cannot be mapped to states of the process model. Note that the authors exclude unfitting behavior from this approach and state that unfitting event logs should be preprocessed to fit the model. Duplicating the aligned log will not change generalization since the sum of trace-generalization is averaged over the number of traces in the log.

\subsubsection{Proposition 21 GenPro8${}^{0}$}

\paragraph{\textbf{Alignment generalization ($gen_S$).}}
\textit{Proposition does not hold.} \\
\textbf{Reasoning.} According to the definition, generalization can never become 1. It only approaches 1. Consider a model allowing for just $\traces{m}={\la a \ra}$ and the log $l = [\la a ra ^k] $. The log visits the state k-times and observes one activity $w =1$ in this state. The function $pnew = \frac{1(1+1)}{k(k-1})$  will approach 0 as $k$ increases but never actually be 0. Hence, $gen_S(l,m) = 1 - pnew (1,k) $ approaches 1, but will never be precisely 1.

\paragraph{\textbf{Weighted negative event generalization ($gen_T$).}}
\textit{Proposition holds.} \\
\textbf{Reasoning.} If the model allows for any behavior, it does not contain any negative behavior which is not allowed. Hence the algorithm cannot find negative events, which are not enabled during replay (DG) and generalization will be maximal. \linebreak $gen_R = AG(l,m) / (AG(l,m) + DG(l,m))  =  AG(l,m) / (AG(l,m) + 0) = 1$.

\begin{figure}[t!]
	{
		\centering
		\includegraphics[width=0.6\textwidth]{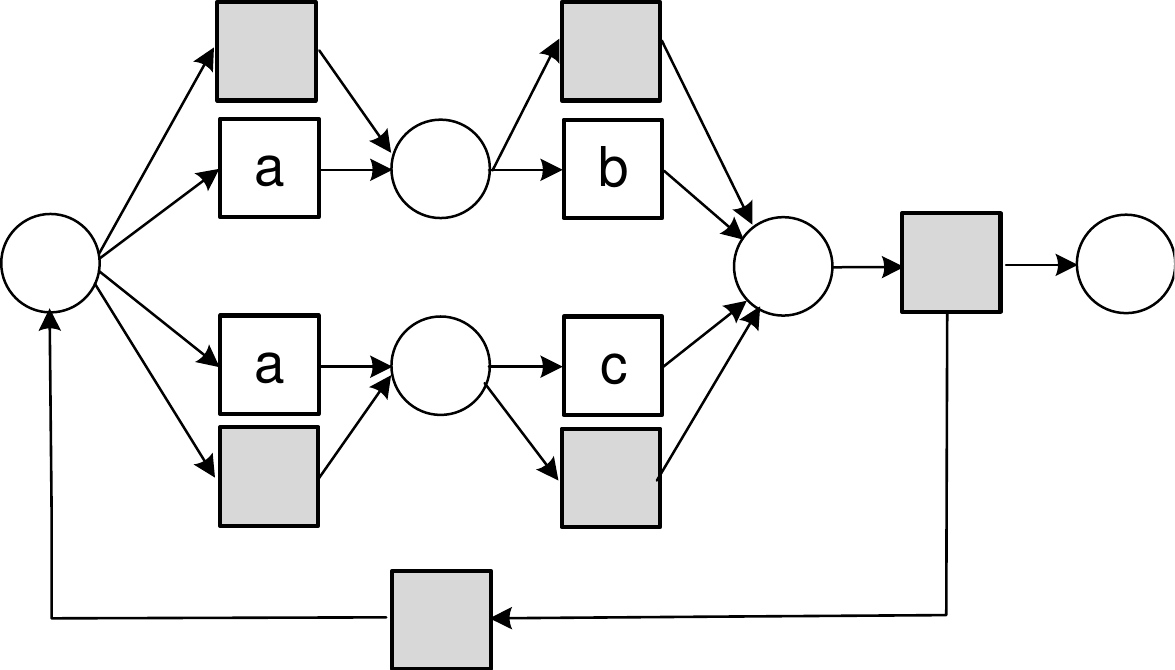}
		\caption{A process model that allows for any behavior while displaying different states.
		}\label{ce-mod8}
	}
\end{figure}

\paragraph{\textbf{Anti-alignment generalization ($gen_U$).}}
\textit{Proposition does not hold.} \\
\textbf{Reasoning.} Assume a model that allows for any behavior because of silent transition, loops and duplicate transitions. The distance between the log and the anti-alignment is maximal. However, due to the duplicate transitions which are connected to separate places the recovery distance is not minimal. Consequently, generalization would not be maximal which violates the proposition. Figure~\ref{ce-mod8} is an example of such a process model.

\end{document}